\documentclass{article}

\usepackage[nonatbib, preprint]{neurips_2026}

\usepackage[numbers,sort&compress]{natbib}

\usepackage[utf8]{inputenc} 
\usepackage[T1]{fontenc}    
\usepackage{hyperref}       
\usepackage{url}            
\usepackage{booktabs}       
\usepackage{amsfonts}       
\usepackage{nicefrac}       
\usepackage{microtype}      
\usepackage{xcolor}         

\usepackage[accsupp]{axessibility}
\usepackage{graphicx}
\usepackage{ragged2e} 
\usepackage{soul} 
\usepackage{xcolor}

\usepackage{graphicx}
\usepackage{booktabs}
\usepackage{threeparttable}
\usepackage{xspace}
\usepackage[dvipsnames]{xcolor}
\usepackage{comment}
\usepackage{adjustbox}
\usepackage{array}
\usepackage{makecell}
\usepackage{pifont}
\usepackage{multirow}
\usepackage{flushend}
\usepackage{colortbl}
\usepackage{enumitem}
\usepackage{algorithm}
\usepackage{algorithmic}
\usepackage{pgf}
\usepackage{import}
\usepackage{comment}
\usepackage[normalem]{ulem}
\usepackage{sidecap}
\usepackage{multirow}
\usepackage{makecell}
\newcolumntype{R}[2]{%
    >{\adjustbox{angle=#1,lap=1.3\width-(#2)}\bgroup}%
    l%
    <{\egroup}%
}
\usepackage[]{pgfplots,pgfplotstable}
\usepackage[]{tikz}
\usepackage{tabularx}
\usetikzlibrary{shapes.geometric, fit, positioning, calc, arrows.meta}
\usepackage{nicematrix}
\usepackage{siunitx}
\usepackage{etoolbox} 
\robustify\bfseries
\usepackage{xcolor}
\definecolor{ourscolor}{RGB}{142, 69, 133}
\usepackage{xfp} 
\newcommand{\round}[1]{\fpeval{round(#1, 2)}}

\makeatletter
\newcommand\myparagraph{\@startsection{paragraph}{4}{\z@}%
    {-6\p@ \@plus -3\p@ \@minus -3\p@}%
    {-0.5em \@plus -0.22em \@minus -0.1em}%
    {\normalfont\normalsize\itshape}}
\renewcommand\paragraph{\@startsection{paragraph}{4}{\z@}%
    {1.25ex \@plus 1ex \@minus .2ex}%
    {-1em}%
    {\normalfont \normalsize \bfseries}}
\makeatother

\definecolor{myblue}{rgb}{0.19, 0.55, 0.91}
\definecolor{myred}{rgb}{0.82, 0.1, 0.26}
\definecolor{MyGreen}{RGB}{0, 104, 55}
\definecolor{MyRed}{RGB}{248, 3, 7}
\definecolor{MyYellow}{RGB}{180, 180, 0}

\newbool{showcolors}
\booltrue{showcolors} 
\ifbool{showcolors}{

}{

}

\newenvironment{manualproof}[1][Proof]
  {\par\noindent\textbf{#1.}\ }
  {\hfill$\square$\par}

\def\ours{REPA-G\xspace} 
\def\ourslong{Representation-Aligned Guidance\xspace} 

\newcommand{\cX}{\mathcal{X}}
\newcommand{\cZ}{\mathcal{Z}}
\newcommand{\bR}{\mathbb{R}}

\usepackage{amsmath}

\usepackage{mathrsfs}

\usepackage{amsthm}

\newcommand{\fdino}{\ensuremath{\text{F}_{\text{DINO}}}} 
\newcommand{\fbase}{\ensuremath{\text{F}_{\text{SiT}}}} 
\newcommand{\falign}{\ensuremath{\text{F}_{\text{SiT}}}} 
\newcommand{\fproj}{\ensuremath{\text{F}_{\text{SiT}}^{*}}} 
\definecolor{turquoise}{RGB}{64, 224, 208}

\newcommand{\orangebox}{\textcolor{orange}{\rule{1.2ex}{1.2ex}}}

\newcommand{\turqoise}{\textcolor{turquoise}{\rule{1.2ex}{1.2ex}}}

\usepackage[abs]{overpic} 

\def\App{App.}
\newcommand{\appref}[1]{\App~\ref{#1}}
\theoremstyle{plain}
\newtheorem{theorem}{Theorem}[section]
\newtheorem{proposition}[theorem]{Proposition}
\newtheorem{lemma}[theorem]{Lemma}

\theoremstyle{definition}

\newtheorem{assumption}[theorem]{Assumption}
\theoremstyle{remark}

\title{\ours: Test-Time Conditioning with Representation-Aligned Visual Features}

\author{
  Nicolas Sereyjol-Garros$^{\,1,\,2}$\\
  \And
  Ellington Kirby$^{1}$
  \And
  Victor Letzelter$^{1,\,3}$
  \And
  Victor Besnier$^{1}$
  \And
  Nermin Samet$^{1}$
  \AND
  \normalfont $^1$ Valeo.ai, Paris, France \\
  \normalfont $^2$ LIGM, CNRS, Univ Gustave Eiffel, ENPC, Institut Polytechnique de Paris, France \\
  \normalfont $^3$ LTCI, Télécom Paris, Institut Polytechnique de Paris, France
}

\begin{document}

\maketitle

\begin{abstract}
While representation alignment with self-supervised models has been shown to improve diffusion model training, its potential for enhancing inference-time conditioning remains largely unexplored. 
We introduce \ourslong (\ours), a framework that leverages these aligned representations, with rich semantic properties, to enable test-time conditioning from features, in generation. By optimizing a similarity objective (the \textit{potential}) at inference, we steer the denoising process toward a conditioned representation extracted from a pre-trained feature extractor. Our method provides versatile control at multiple levels of granularity, ranging from patch level matching via single patches to broad semantic guidance using global image feature tokens. We further extend this to multi-concept composition, allowing for the faithful combination of distinct concepts. 
\ours operates entirely at inference time with no additional training required, offering a flexible and precise alternative to often ambiguous text prompts or coarse class labels.
Our approach achieves high-quality, diverse generations on ImageNet and COCO.
Code is available at https://github.com/valeoai/REPA-G
\end{abstract}    
\section{Introduction}
\label{sec:intro}
\begin{figure}[t]
    \centering
    \resizebox{0.85\linewidth}{!}{
        \begin{tikzpicture}[
            column2/.style={anchor=center, font=\small, align=justify, text width=4.5cm},
            column3/.style={anchor=center},
            header/.style={font=\Large\bfseries, align=center, anchor=south},
            subhead/.style={font=\large\bfseries, text=black!90, anchor=north}, 
            rect/.style={fill=turquoise, draw=none, line width=0pt, draw opacity=0, minimum width=0.3cm, minimum height=0.3cm, inner sep=0pt}
        ]
            
            \draw[line width=1.pt, black!90] (-0.6, 0.45) -- (13.6, 0.45); 

            \draw[line width=1.pt, black!90] (-0.6,-0.4) -- (13.6,-0.4); 

            \draw[line width=1pt, black!90] (-0.6,-0.95) -- (6.5,-0.95); 
            \draw[line width=1pt, black!90] (6.8,-0.95) -- (13.6,-0.95); 

            \draw[black!90] (-0.6,-4.24) -- (6.7,-4.24);
            
            \draw[line width=1.pt, black!90] (-0.6,-7.5) -- (13.6,-7.5); 

            \draw[line width=1.pt, black!90] (6.7,-0.5) -- (6.7,-7.5);

            \node[header] at (3.05, -0.35) {Standard Conditionings};  
            \node[header] at (10.25, -0.35) {\ours (ours)}; 

            \node[subhead] at (1.5, -0.4) {Condition};
            \node[subhead] at (4.9, -0.4) {Output};
            \node[subhead] at (8.5, -0.4) {Condition};    
            \node[subhead] at (12., -0.4) {Output}; 

            \node[column2, align=center] at (1.5, -2.6) {\textbf{Class Condition} \\ $\langle$Angora Rabbit$\rangle$};
            \node[column3] at (4.9, -2.6) {\includegraphics[width=3cm, height=3cm]{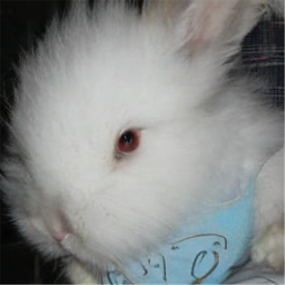}};
            \node[column3] at (12., -2.6) {\includegraphics[width=3cm, height=3cm]{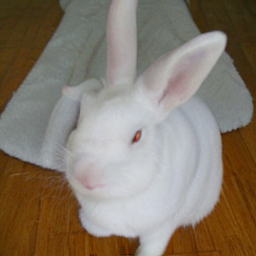}};

            \node (orgim) at (8.1, -2.1) {\includegraphics[width=2.1cm, height=2.1cm]{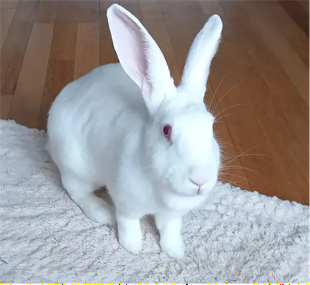}};
            \node (volcano) at (8.1, -6.4) {\includegraphics[width=2.1cm, height=2.1cm]{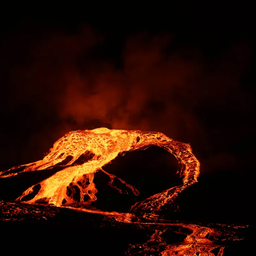}};
            \node (featmap) at (8.1, -3.7) {\includegraphics[width=0.8cm, height=0.8cm]{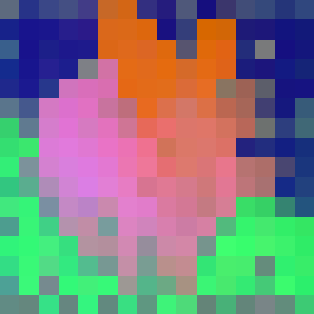}}; 
            \node[left=-0.15cm of featmap, font=\scriptsize, align=center, rotate=90, anchor=south] {Image \\ Feats.};
            \node (targetfeat) at (8.1, -4.8) {\includegraphics[width=0.8cm, height=0.8cm]{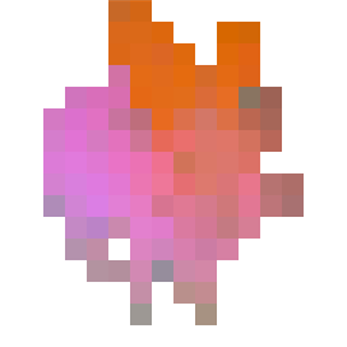}};
            \node[left=-0.15cm of targetfeat, font=\scriptsize, align=center, rotate=90, anchor=south] {Masked \\ Feats.};

            \draw[->, black, thick] (8.55,-3.7) -- (9.35, -3.7)
                node[midway, above, font=\scriptsize\bfseries] {Avg.}; 
            \node[rect, fill=orange] (dot2) at ([xshift=1.45cm,yshift=-0.0cm]featmap.center) {};
            \draw[->, black, thick] (9.75,-3.7) -- (10.45, -3.7)
                node[midway, above, font=\scriptsize\bfseries] {};
            \draw[->, black, thick] (8.55,-4.8) -- (10.45, -4.8)
                node[midway, above, font=\scriptsize\bfseries] {};

            \node[column2] at (1.5, -5.9) {
            \begin{minipage}{4cm}
                \centering \textbf{Text Prompt} \\ \smallskip
                \scriptsize A hyper-realistic, close-up of a fluffy white albino rabbit sitting on active volcanic terrain. Ground is cracked, blackened earth with fissures of bright, molten lava bubbling underneath. The light of lava casts an orange glow onto the rabbit's paws and lower fur. Rabbit has tall, upright ears. Background is dark, blurry volcanic landscape. High contrast, surreal, photorealistic.
            \end{minipage}
};
            \node[column3] at (4.9, -5.9) {\includegraphics[width=3cm, height=3cm]{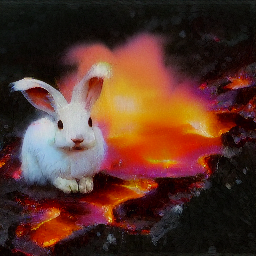}};

            \node[rect] (dot) at ([xshift=1.45cm,yshift=-0.85cm]volcano.center) {};
            \draw[->,thick, black] (8.70, -7.26) -- (9.35, -7.26);

            \node[column3] at (12., -5.9) {\includegraphics[width=3cm, height=3cm]{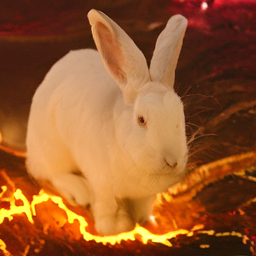}};
            \draw[->, black, thick] (9.75,-7.26) -- (10.45, -7.26)
                node[midway, above, font=\scriptsize\bfseries] {};

            \draw[
                draw=yellow,
                line width=1pt,
                opacity=1.0,    
                rounded corners=0pt
            ] 
            ($(volcano.north east) + (-0.55cm, -2.cm)$) 
            rectangle 
            ($(volcano.north east) + (-0.65cm, -2.1cm)$);

        \end{tikzpicture}
    }
    \caption{\textbf{Comparing class-label, text-prompt, and \ours conditioning.} All models are trained on ImageNet. \textit{(Top)} We average extracted features (\orangebox) from a anchor image to generate a generic ``rabbit'' image. \textit{(Bottom)} We combine a masked feature map with a specific ``lava'' patch (\turqoise) to synthesize a rabbit on a volcano. While text prompts require lengthy descriptions and often lack precision, our feature-based conditioning offers better compositional control and
    provides more precise generation.}
    \label{fig:teaser_cond_comparison}
\end{figure}

In recent years, generative models have achieved impressive results in image synthesis~\citep{flux2024,esser2024scaling}, reaching near photo-realistic quality. This progress has been driven in particular by diffusion models~\citep{song_score-based_2020,ho_denoising_2020} and flow matching~\citep{lipmanflow,ma_sit_2024} approaches, which iteratively remove Gaussian noise from images during generation. Recently, a line of work has shown that, beyond the standard denoising objective, diffusion models can be trained to align their intermediate representations~\citep{wang_repa_2025, leng_repa-e_2025, tian_u-repa_2025, wu2025geometryforcing, sereyjol2026r3dpa} using pre-trained self-supervised learning models such as DINOv2~\citep{oquab2024dinov2}. This representation alignment has been shown to accelerate training and improve sample quality. However, it remains unclear how these aligned representations can be effectively leveraged at inference time.

Beyond improving sample quality, controllability has emerged as a central challenge in image synthesis. Current conditioning strategies typically rely on class labels~\citep{peebles2023scalable} or textual descriptions~\citep{rombach_high-resolution_2022}. Class-based conditioning provides only coarse control and is limited to predefined categories. Using text is more flexible but often imprecise and ambiguous (see the left panel of Figure~\ref{fig:teaser_cond_comparison}). 

In this work, we leverage representation alignment training in diffusion models for conditional image generation. Our method (\ours) enables test-time visual conditioning via inference-time optimization by exploiting the alignment between a diffusion model's internal representations and those of a vision encoder. By steering the diffusion process toward target feature tokens, we achieve control over generation without relying on a fixed set of class labels or potentially ambiguous textual prompts. This improves controllability while preserving flexibility and generality.

Specifically, we extract visual tokens from a real image using the same self-supervised learning network used during representation alignment training, 
for example DINOv2~\cite{oquab2024dinov2}. We then guide generation using a single token or a set of tokens that capture the concept at varying levels of granularity .
The model can be conditioned with multiple tokens extracted from a mask to preserve shape or pose. Finally, conditioning with the average of all image tokens provides a broad concept signal such as \emph{rabbit}, \emph{car} (see the top-right of Figure~\ref{fig:teaser_cond_comparison}). 

We achieve this by computing the gradient of a \textit{potential} that quantifies the alignment between representations of generated sample and conditioning data points, corresponding to semantic \textit{concepts} tokens. In addition to predicted scores and the velocities, the sampling procedure is guided by this gradient. We then generalize the idea to multiple potentials by aligning with several concept tokens simultaneously. Our proposed method pulls relevant features toward the specified concepts enabling faithful generation, and makes it possible to generate flexible compositions (Figure \ref{fig:teaser_cond_comparison}, bottom right).

In summary, we introduce the first test-time visual conditioning framework for diffusion models trained with representation alignment. Our contributions are as follows:

\textbf{We introduce \ourslong (\ours)}. \ours leverages representation alignment during sampling to align internal diffusion features with targets from a vision encoder. Our method operates entirely at inference time without requiring additional parameters or finetuning.

\textbf{We analyze the properties of the representation space for feature conditioning.} We justify that the embedding space provided by self-supervised feature alignment is well suited for this task.

\textbf{We conduct extensive experiments, showing control over concrete and abstract visual concepts}. Our framework handles objects and background semantics, enabling both spatial and concept-level guidance as well as concept composition.
\section{Related works}
\label{sec:rw} 

\paragraph{Representation alignment for generation.}

Image synthesis is dominated by diffusion and flow matching models~\citep{ho_denoising_2020,lipmanflow}. These methods enable photorealistic image generation at the cost of a slow generation speed and a long training time~\citep{song2021denoising}. Recently, REPA \cite{yu2025representation} argues that the training speed of latent diffusion models is largely constrained by how quickly they learn a meaningful internal image representation \cite{wang_repa_2025}. To mitigate this, they guide the learning by aligning intermediate model projections with feature maps from a pre-trained encoder, which accelerates training and improves generation quality.
iREPA \cite{singh2025irepa} shows that spatial structure in representations correlates more strongly with generation quality than global information. They use a convolutional projection instead of a MLP and a spatial normalization layer. 
VA-VAE \cite{yao_reconstruction_2025} applies a latent alignment loss solely in the VAE latent space prior to diffusion training. Finally, REPA-E \cite{leng_repa-e_2025} enables end-to-end training by backpropagating the REPA loss through both the diffusion model and the VAE encoder. This jointly shapes their internal representations, yielding better alignment and higher-quality generations. These representation-alignment methodologies have since been extended to various modalities beyond static images, including video~\cite{wu2025geometryforcing}, LiDAR point clouds~\cite{sereyjol2026r3dpa}, and audio synthesis~\cite{ton2025taro}.

\paragraph{Conditional generation in diffusion models.}
Beyond sample quality, recent research has increasingly emphasized controllability as a central aspect of image synthesis, often considered as important as visual fidelity or diversity~\citep{hertz2023prompttoprompt,couairondiffedit}. Early adaptations of diffusion models to image generation were largely unconditional~\citep{ho_denoising_2020}. Subsequent works introduced conditioning through auxiliary classifiers~\citep{dhariwal2021diffusion} or classifier-free guidance~\citep{ho2022classifier}. Nowadays, diffusion models are commonly conditioned during training on various modalities, including class labels~\citep{dhariwal2021diffusion}, textual descriptions~\citep{rombach_high-resolution_2022,saharia2022photorealistic}, or reference images for image editing~\citep{labs2025flux1kontextflowmatching}. These mechanisms typically require task-specific training and architectural modifications that remain fixed at inference. An alternative line of work explores post-training strategies~\citep{controlnet,ye2023ip-adapter,stracke2024ctrloralter}, which enable conditioning on structural signals like depth or edge maps. While effective, these approaches introduce additional parameters and require extra post-training steps, increasing both model complexity and computational cost, making them not directly comparable and orthogonal to our approach.

Our method enables test-time control in trained flow models, without additional training or parameters, provided that representation alignment holds. This is orthogonal to image inpainting methods \citep{xie2023smartbrush, yang2023paint}, which typically require explicit training on masked data and text prompts, often using Stable Diffusion models \citep{rombach_high-resolution_2022} pre-trained on large-scale, scene-centric datasets. In contrast, \ours targets general guidance in generation, allowing conditioning on arbitrary features rather than text.

To the best of our knowledge, we are the first to explore such test-time conditioning  with visual features in this context. TFG~\citep{ye_tfg_2024} provides a unified framework for various training-free methods, but it does not explore visual feature guidance.

\cite{mallat_ttc} study the interpretability of intermediate features in an unconditional diffusion model, rather than proposing a test-time conditioning method. 

\section{Preliminaries}\label{sec:preliminary}

Let $p_{0}$ be a (clean) data distribution on $\cX$, which may correspond to pixel space in image generation.
We describe hereafter the training of a flow matching model to be able to produce new samples from $p_{1}\sim\mathcal{N}(\mathbf{0},\mathbf{I})$ to $p_{0}$ while being ``representation-aligned'' with a pretrained backbone $\phi$.
For more details on the training and sampling of flow models see \appref{app:proof_prelim}.

The flow model $v_\theta$ can be expressed as $v_\theta = g_{\theta} \circ f_{\theta}$ where $f_{\theta}: \cX \times [0,1] \rightarrow \cZ$, projects input into a representation $f_{\theta}(x_{t}, t)$ at a given layer, and $g_{\theta}: \cZ \rightarrow \cX$ process this representation to construct the prediction. REPA \cite{yu2025representation} suggests optimizing the flow matching denoising loss $\mathcal{L}^{\mathrm{diff}}(\theta)$ along with an alignment loss using an additional projection layer $h_{\theta} : \mathcal{Z} \rightarrow \mathcal{Z}'$:
\begin{equation}
\begin{aligned}
\mathcal{L}^{\mathrm{align}}(\theta)
&= - \mathbb{E}_{\substack{
t \sim \mathcal{U}(0,1) \\
(x_0, x_1) \sim p_{0} \times p_{1}
}}
\Big[
\mathscr{V}\big( (h_{\theta} \circ f_{\theta})(x_{t}, t),\, \phi(x_0) \big)
\Big] .
\end{aligned}
\label{eq:alignment}
\end{equation}

in a compound loss $\mathcal{L}^{\mathrm{diff}}(\theta) + \beta \mathcal{L}^{\mathrm{align}}(\theta)$, with $\beta > 0$.

In \eqref{eq:alignment}, $\mathscr{V}: \cZ' \times \cZ' \rightarrow \bR$ is a \textit{potential} that measures similarity between the features, predicted by $f_{\theta}$ and projected with $h_{\theta}$, with those of a pretrained (frozen) backbone $\phi: \cX \rightarrow \cZ'$ evaluated on the \textit{clean} image. When $\cZ' = \mathbb{R}^{N \times d}$, $N$ is the number of features and $d$ is the features dimension, $\mathscr{V}$ is typically expressed as an average patch-wise similarity $\mathscr{V}(h_{t}, h^{\star}) = \frac{1}{N} \sum_{n=1}^{N} \langle [h_{t}]_{n}, [h^{\star}]_{n} \rangle$ where $\langle \cdot, \cdot \rangle$ denotes the euclidean dot product on $\bR^{d}$ and $[h]_{n}$ is the $n$-th row of $h \in \mathbb{R}^{N \times d}$. Importantly, $[h_{t}]_{n}$ and $[h^{\star}]_{n}$ are set to have unit norm.

\section{Guiding generation at inference}\label{sec:guidance} 

We perform inference by leveraging flow models trained with representation alignment, enabling generation conditioned on features $\phi(x_{c})$ from a reference image $x_{c}$. This approach is analogous to classifier guidance \citep{dhariwal2021diffusion, song_score-based_2020}, but extends the paradigm to continuous features rather than discrete classes.

We formulate the time dependent \textit{posterior} distribution with respect to the condition $h^{\star}$ as $p_t(h^{\star}|x;\lambda)\propto \exp \left( \lambda \mathscr{V}\left((h_{\theta} \circ f_{\theta})(x,t), h^{\star} \right)\right) $ with $\lambda > 0$ controlling the strength of the conditioning.
To generate samples $x$ conditioned on $\phi(x_{c})$ in accordance with classifier guidance literature \citep{dhariwal2021diffusion,song2021denoising}, we introduce a guidance term that modifies the score function $\nabla_x \log p_t(x_t)$ used during sampling as:

\begin{equation}
\nabla_x \log p_t(x_t) + \lambda \nabla_x \mathscr{V}\left((h_{\theta} \circ f_{\theta})(x_{t},t),\phi(x_{c})\right)\;,\label{eq:ttvc}
\end{equation}

The guidance term allows to steer the samples toward regions of low energy while the unconditional score transport them toward the image manifold. Under assumptions provided in \appref{apx:proof_lemma}, sampling with this modified score allow to recover the tilted distribution $p_0(x|h^{\star};\lambda)$. It results in an effective way to condition the generation as shown in the 2D toy example in \appref{apx:toy}. We also show in Section~\ref{sec:experiments} that it allows to generate high quality images with high semantic alignment with the condition. Details of the algorithm are provided in \appref{app:algorithm}.
\section{Properties of representation space}
\label{sec:properties}

\newcommand{\img}[1]{\includegraphics[height=0.8cm]{#1}}

\newcommand{\blockdino}{
{%
\setlength{\tabcolsep}{0pt}%
\renewcommand{\arraystretch}{0}%
\setlength{\fboxsep}{0pt}    
\begin{tabular}{@{}cccc@{}}
{\setlength{\fboxsep}{0.0pt}\fcolorbox{red}{white}{\img{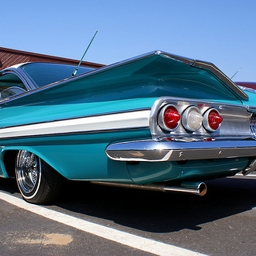}}} &
\img{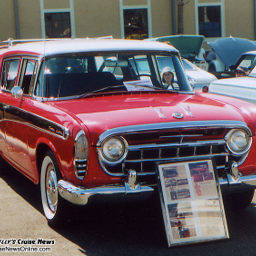} &
\img{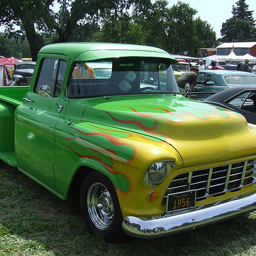} &
\img{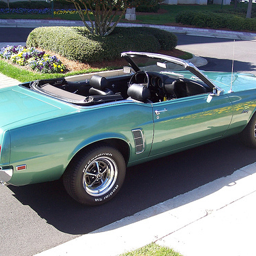} \\

\img{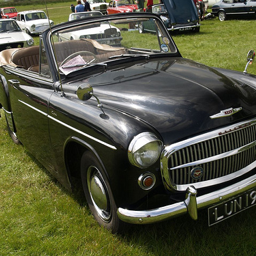} &
\img{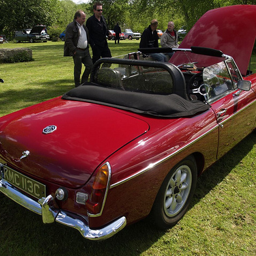} &
\img{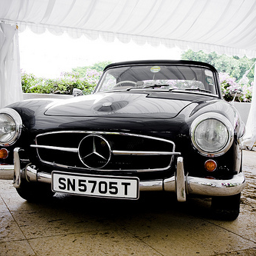} &
\img{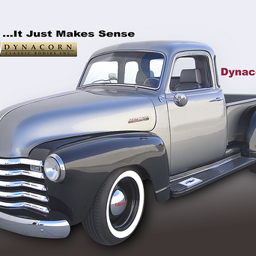} \\

\end{tabular}
}%
}

\newcommand{\blocksit}{
{%
\setlength{\tabcolsep}{0pt}%
\renewcommand{\arraystretch}{0}%
\begin{tabular}{@{}cccc@{}}
{\setlength{\fboxsep}{0.0pt}\fcolorbox{red}{white}{\img{fig/clusters/ref_car.jpg}}} &

\img{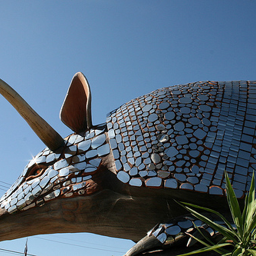} &
\img{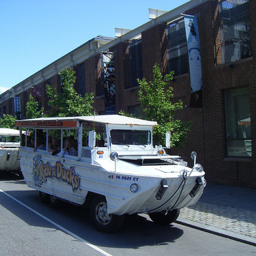} &
\img{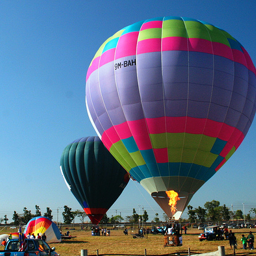} \\

\img{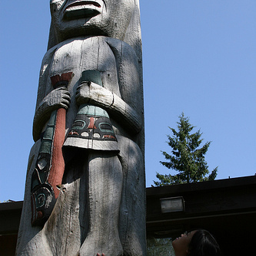} &
\img{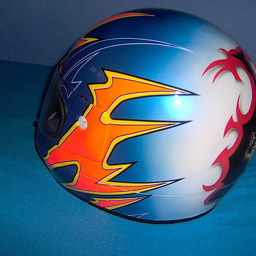} &
\img{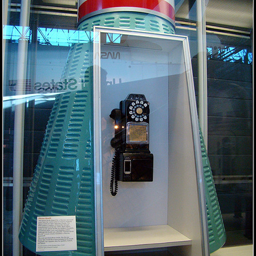} &
\img{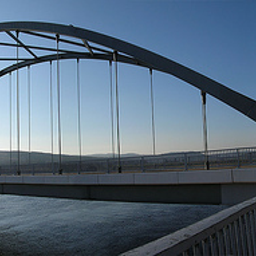} \\

\end{tabular}
}%
}

\newcommand{\blocksitrepa}{
{%
\setlength{\tabcolsep}{0pt}%
\renewcommand{\arraystretch}{0}%
\begin{tabular}{@{}cccc@{}}
{\setlength{\fboxsep}{0.0pt}\fcolorbox{red}{white}{\img{fig/clusters/ref_car.jpg}}} &
\img{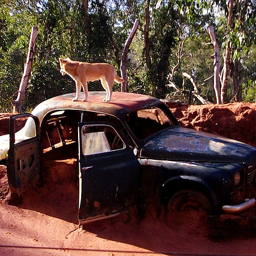} &
\img{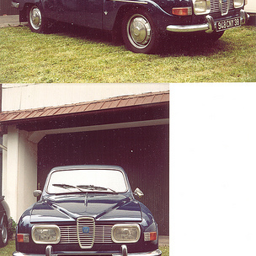} &
\img{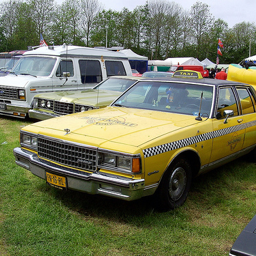} \\

\img{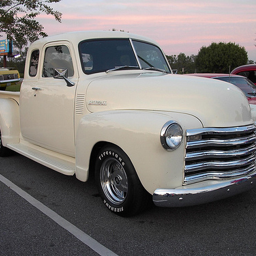} &
\img{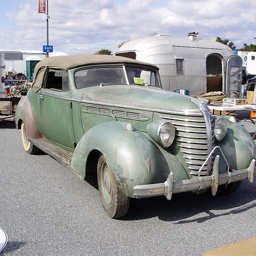} &
\img{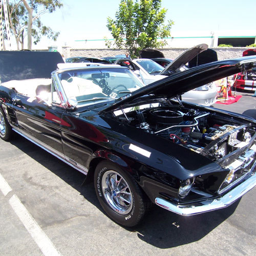} &
\img{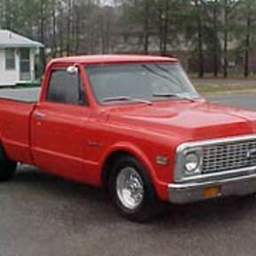} \\

\end{tabular}
}%
}

\newcommand{\blocksitrepaproj}{
{%
\setlength{\tabcolsep}{0pt}%
\renewcommand{\arraystretch}{0}%
\begin{tabular}{@{}cccc@{}}
{\setlength{\fboxsep}{0.0pt}\fcolorbox{red}{white}{\img{fig/clusters/ref_car.jpg}}} &
\img{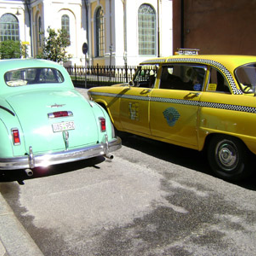} &
\img{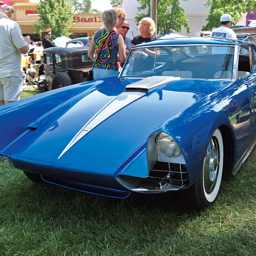} &
\img{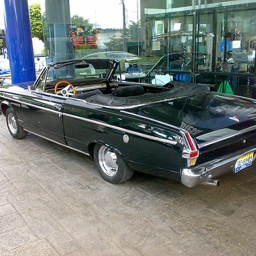} \\

\img{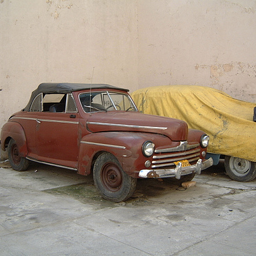} &
\img{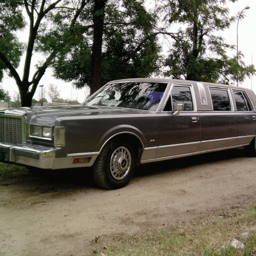} &
\img{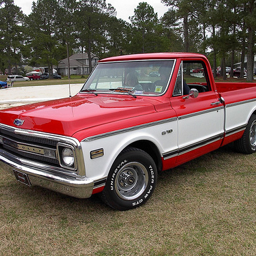} &
\img{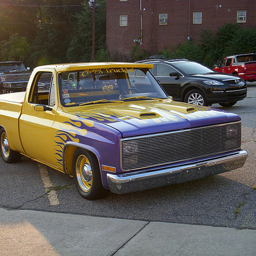} \\

\end{tabular}
}%
}




\begin{figure}[t]
\centering
\setlength{\tabcolsep}{1.5pt} 
\begin{tabular}{cccc}
{\textbf{\small DINOv2}} & {\textbf{\small SiT}} & {\textbf{\small SiT (Align)}} & {\textbf{\small SiT (Align + Proj)}} \\
\blockdino & \blocksit & \blocksitrepa & \blocksitrepaproj \\
\end{tabular}

\caption{\textbf{Impact of representation alignment on feature space.} 
We perform $k$-means clustering ($k=1{,}000$) on four feature spaces across ImageNet (see~\appref{app:clusteringimagenet}). For a reference image (with red frame), we visualize others from its assigned cluster. Without alignment, SiT fails to form semantic groupings, making its latent space unsuitable for conditioning. In contrast, the aligned model successfully replicates the teacher's semantic structure both before and after the projection layer, resulting in semantically consistent clusters.
}

\label{fig:clusters}
\end{figure}

We study how representation alignment \citep{yu2025representation} shapes the internal feature space of diffusion transformers and how this impacts feature conditioning. To this end, we use SiT models \citep{ma_sit_2024} trained on ImageNet \citep{deng2009imagenet}. In this setting, the generation is done in the latent space of an autoencoder, but the visual backbone $\phi$ takes images as input.

We argue that effective conditioning requires: $(i)$ a semantically meaningful embedding space, where nearby embeddings correspond to similar concepts; and $(ii)$ a smooth mapping from conditioning features to conditional distributions, so similar embeddings yield similar conditional densities.

\paragraph{Representation alignment enables semantic features in diffusion transformers.}

Self-supervised vision models learn rich representations from large-scale unlabeled data by enforcing invariance to augmentations while preserving semantic content.
Among them, DINOv2 provides particularly strong semantic features and is widely used for representation alignment in diffusion models \citep{yu2025representation, wang_repa_2025, leng_repa-e_2025, tian_u-repa_2025}.

In contrast, the diffusion denoising objective alone rarely produces semantic features.
Figure \ref{fig:clusters} highlights this difference: clustering ImageNet with DINOv2 features yields coherent, concept-level groups, whereas clustering SiT's internal features produces clusters non-semantic clusters. 
Training SiT with a representation alignment loss restores semantic information in its internal feature space.
\begin{figure}[ht]
    \centering
    \begin{minipage}{0.48\textwidth}
        \centering
        \includegraphics[width=0.85\linewidth]{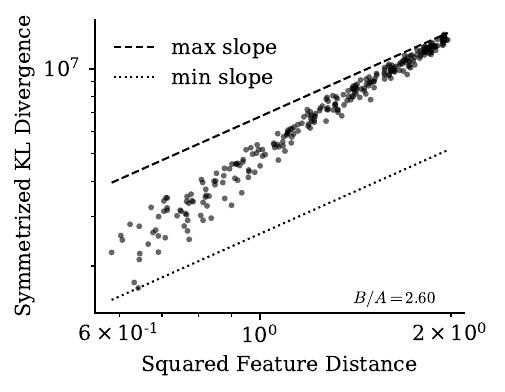}
        \caption{\textbf{Correlation between embedding and density distances.} The narrow range of the $B/A$ ratio indicates a well-conditioned space where distances in any direction behave consistently.}
        \label{fig:kl}
    \end{minipage}\hfill
    \begin{minipage}{0.48\textwidth}
        \centering
        \includegraphics[width=\linewidth]{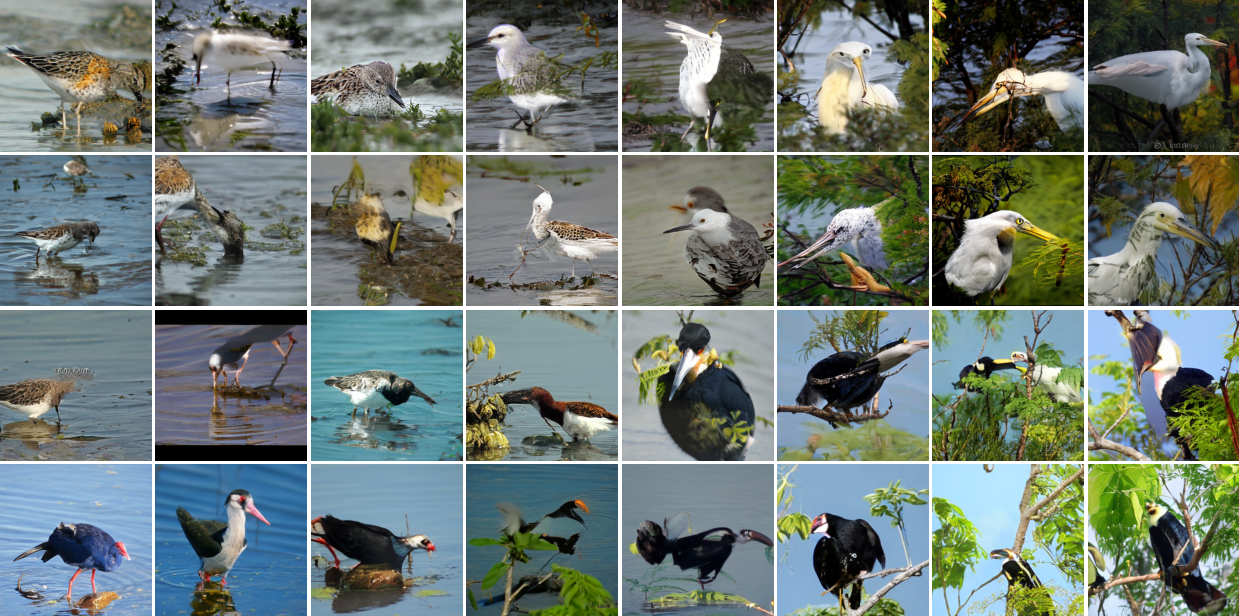}
        \vspace{-3mm}
        \caption{\textbf{Semantic interpolation in the feature space.} Samples are generated by bilinearly interpolating the global conditioning features between four anchor images. The smooth transitions show the semantic coherence and stability of the mapping between the embedding space and conditional densities.}
        \label{fig:interpolation}
    \end{minipage}
    \vspace{-10pt}
\end{figure}
\paragraph{Representation alignment ensures a well-conditioned mapping from features to image densities.}

Following \cite{mallat_ttc}, we evaluate the Euclidean embedding property to ensure the conditioning space structure transfers to the conditional densities. This requires the distance $d^2$ between the conditional densities $p_1 = p(\cdot|\phi(x_1))$ and $p_2 = p(\cdot|\phi(x_2))$ to scale with the squared distance between $\phi(x_1)$, and $\phi(x_{2})$. Specifically, there must exist $0 < A \le B$ and $B/A$ not too large such that for all $x_1$, $x_2$:
\begin{equation}
A \lVert \phi(x_1) - \phi(x_2) \rVert^2
\le d^2(p_1, p_2)
\le B \lVert \phi(x_1) - \phi(x_2) \rVert^2\;.
\end{equation}
We define the density dissimilarity as:
\begin{equation}
    d^2(p_1,p_2) = \frac{\lambda}{N} \sum_{n=1}^N\langle  [\mathbb{E}_{p_1}[\phi(x)] - \mathbb{E}_{p_2}[\phi(x)]]_n, [\phi_1-\phi_2]_n\rangle\;,
\end{equation}

with more details in \appref{apx:eep}. Figure~\ref{fig:kl} shows a strong relation between embedding and density distances, with a tight ratio $B/A=2.60$, confirming the stability of the conditional space.

We further validate the semantic coherence of the embedding space and the smoothness of the density mapping in Figure~\ref{fig:interpolation} by interpolating between conditioning features. The smooth transitions confirm the meaningful structure of the embedding space and its stable mapping to conditional densities.

\section{Design choices for the Potential $\mathscr{V}$}
\label{sec:method}

We have presented a guidance method for test-time visual conditioning within the representatively aligned class of diffusion models. While Section \ref{sec:preliminary} outlined the required properties of the feature extractor $\phi$, the model is specifically trained to align with the potential:
\begin{equation}\mathscr{V}(h, h^{\star}) = \frac{1}{N} \sum_{n=1}^{N} \langle [h]_{n}, [h^{\star}]_{n} \rangle \; ,\label{eq:potential}
\end{equation}
where $h, h^{\star} \in \mathbb{R}^{N \times d}$. In this section, we discuss how \eqref{eq:potential} can be adapted during inference to achieve specific characteristics in the generated images.

\subsection{Guidance via Independent Patch Alignment (IPA)}\label{sec:weighting}

Eq. \eqref{eq:potential} can be generalized to allow for spatial flexibility:
\begin{equation}
\mathscr{V}_P\left(h, h^{\star}\right)=\sum_{n=1}^N \sum_{m=1}^N P_{n,m}\left\langle[h]_n,\left[h^{\star}\right]_m\right\rangle\;,
\label{eq:generalization}
\end{equation}
where $P \in \mathbb{R}^{N \times N}$ is a weight matrix such that $P_{n,m}$ defines the interaction strength between predicted patch $n$ and conditioning patch $m$, normalized such that $\sum_{n} \sum_{m} P_{n,m} = 1$. By manipulating $P$, we achieve fine-grained control over the generation process. Because $P$ is determined independently of the feature alignments, we refer to this class of conditioning as \textit{Independent Patch Alignment} (IPA). We detail three specific configurations of $P$ below.

\paragraph{Alignment with full feature map.} Setting $P = \frac{1}{N} I$ recovers the original alignment objective \eqref{eq:potential}. This configuration enforces a dense, spatial constraint, resulting in a stochastic reconstruction of the conditioning image.  

\paragraph{Alignment with feature mask.} When only specific regions of the conditioning image are relevant, we define a binary mask $m \in \{0,1\}^{N}$ and its corresponding index set $\mathscr{S} = \{n \mid m_{n} = 1\}$. The weights are defined as $P_{n,m} = \frac{1}{|\mathscr{S}|} \mathbf{1}[m = n] \mathbf{1}[n \in \mathscr{S}]$. This \textit{masked-conditioning} potential preserves information within the masked region while allowing for variation in the surrounding areas.

\paragraph{Alignment with an average concept.} To steer generation toward the global semantic content of a reference image without enforcing pixel-wise spatial fidelity, we utilize an \textit{average-concept} potential. Here, we set $P_{n,m} = \frac{1}{N^{2}\|\bar{h}\|\|\bar{h^*}\|}$, which is equivalent to aligning the spatial average of the predicted features with the spatial average of the conditioning features, $\bar{h}^{\star}$. This preserves the general semantic category while removing all spatial constraints.

\paragraph{Alignment with a single concept.} Alternatively, we consider \textit{single-concept} alignment, where the full generated feature map is compared against a single target patch $i$. This corresponds to $P_{n,m} = \frac{1}{N} \mathbf{1}[m = i]$, focusing the guidance signal on a local feature from the conditioning image.

\subsection{Guidance via Selective Patch Alignment (SPA)} 
\label{sec:spa}

While IPA uses constant weighting, \textit{Selective Patch Alignment} dynamically weights predicted patches based on their similarity to a target concept $h^{\star}_i$, as follows:
\begin{equation}
\mathscr{V}^{\mathrm{SPA}}(h, h^{\star}) = T\log \left[ \sum_{n=1}^{N}\exp \left( \frac{\langle [h]_{n}, [h^{\star}]_{i}\rangle}{T}\right) \right]\;,
\label{eq:free_energy}
\end{equation}
where the temperature $T$ modulates selection sparsity. As $T \rightarrow 0$, the potential recovers a hard-maximum over patches; as $T \rightarrow \infty$, it converges to the uniform average-concept baseline. This soft-maximum mechanism allows the model to adaptively localize concepts, bypassing the rigid spatial constraints of the conditioning image. More intuition and justifications are given in~\appref{apx:spa}.

\section{Experiments}
\label{sec:experiments}

\textbf{Datasets.}
We evaluate on ImageNet~\cite{deng2009imagenet}, which contains 1.2M images across 1,000 classes, and on COCO~\cite{coco}, which includes 123K images.
Both datasets are preprocessed to a resolution of 256×256.

\textbf{Implementation details.}
We evaluate the performance of our proposed test-time visual feature guidance mechanisms by analyzing their generated outputs. 
For our experiments, we use a pretrained flow matching models with representation alignment on the ImageNet.
We first evaluate the conditional generation capabilities of \ours across varying granularities of feature types. We then explore the ability of our method to compose features from multiple images.
We follow~\cite{ma_sit_2024} by using a Stochastic Differential Equation integrated via the Euler-Maruyama solver with 250 steps.

\subsection{Test-time guidance with single source}

In this series of experiments, we use a single "anchor image" whose features guide the generation process. We generate images using guidance mechanisms derived from Independent Patch Alignment (IPA), including: (i) Full Feature Map, utilizing all patch features; (ii) Masked Feature Map, utilizing a  subset of patch features; and (iii) Average Feature Map, utilizing the global average of all patch features. To generate the masked feature maps, we extract DINOv2 \cite{oquab2024dinov2} patch features, compute the first principal component, and apply a threshold set to $0.5$ to isolate the foreground~\cite{10.1109/TIP.2024.3380243}.

We evaluate three flow matching models: a standard SiT-XL \cite{ma_sit_2024} (trained without representation alignment), REPA \cite{wang_repa_2025} and REPA-E \cite{leng_repa-e_2025}, both trained with representation alignment. In Table \ref{tab:ipa_imagenet_fids} and Table \ref{tab:ipa_imagenet_sim}, we remove class conditioning to focus solely on visual conditioning, while \appref{app:more_cond} reports results with both class and visual conditioning. For visual conditioning, we extract internal features (\fbase), at $t=0$, from the same backbone layer across all models, specifically the layer before the projection layer used in the REPA variants, i.e. $f_{\theta}(x_{t}, t)$. For the REPA-based models, we also evaluated features extracted after the projection layer (i.e. $h_{\theta}(f_{\theta}(x_{t}, t))$) but observed no significant difference in performance; further details are provided in \appref{app:more_incond}. Additionally, for the aligned models, we extend our evaluation to include DINOv2 features (\fdino) as an external conditioning.

\begin{table*}[t]
\centering
\caption{\textbf{Distribution-level comparison of \ours generations on ImageNet \cite{deng2009imagenet}.}  We compare the standard SiT backbone~\cite{ma_sit_2024} (no representation alignment) against REPA \cite{wang_repa_2025} and REPA-E \cite{leng_repa-e_2025} variants. 
For each method block, the first row denotes results for unconditional generation, serving as a baseline, and \textcolor{ourscolor}{\textbf{\fbase}} and \textcolor{ourscolor}{\textbf{\fdino}} for \ours. 
}
\label{tab:ipa_imagenet_fids}

\resizebox{\linewidth}{!}{%
\begin{tabular}{ll|ccccc|ccccc|ccccc}
\toprule
& & \multicolumn{5}{c|}{\textbf{Full Feature Map}}
& \multicolumn{5}{c|}{\textbf{Masked Feature Map}}
& \multicolumn{5}{c}{\textbf{Average Feature Map}} \\
 \cmidrule(lr){3-7} \cmidrule(lr){8-12} \cmidrule(lr){13-17}

\textbf{Model} & \textbf{Cond.}
& \textbf{FID}$\downarrow$ & \textbf{sFID}$\downarrow$ & \textbf{IS}$\uparrow$ & \textbf{Prec.}$\uparrow$ & \textbf{Rec.}$\uparrow$
& \textbf{FID}$\downarrow$ & \textbf{sFID}$\downarrow$ & \textbf{IS}$\uparrow$ & \textbf{Prec.}$\uparrow$ & \textbf{Rec.}$\uparrow$
& \textbf{FID}$\downarrow$ & \textbf{sFID}$\downarrow$ & \textbf{IS}$\uparrow$ & \textbf{Prec.}$\uparrow$ & \textbf{Rec.}$\uparrow$ \\
\midrule

\multirow{2}{*}{SiT}
    & -

            &                     \textbf{45.02}
            &                     \textbf{9.02}
            &                     \textbf{22.33}
            &                     \textbf{0.50}
            &                     \textbf{0.63}
            &                     \textbf{45.02}
            &                     \textbf{9.02}
            &                     22.33
            &                     \textbf{0.50}
            &                     \textbf{0.63}
            &                     \textbf{45.02}
            &                     \textbf{9.02}
            &                     \textbf{22.33}
            &                     \textbf{0.50}
            &                     \textbf{0.63}
 \\

    & \fbase

            &                     71.35
            &                     72.08
            &                     14.73
            &                     0.33
            &                     0.54
            &                     45.85
            &                     19.23
            &                     \textbf{29.36}
            &                     0.40
            &                     0.62
            &                     65.99
            &                     22.52
            &                     16.69
            &                     0.36
            &                     0.52
 \\

\midrule
\multirow{3}{*}{REPA}
    & -

            &                     26.58
            &                     6.85
            &                     41.31
            &                     0.56
            &                     \textbf{0.70}
            &                     26.58
            &                     6.85
            &                     41.31
            &                     0.56
            &                     \textbf{0.70}
            &                     26.58
            &                     6.85
            &                     41.31
            &                     0.56
            &                     \textbf{0.70}
 \\

    & \textcolor{ourscolor}{\textbf{\fdino}}

            &                     7.23
            &                     8.86
            &                     199.69
            &                     0.65
            &                     0.69
            &                     14.17
            &                     10.59
            &                     135.08
            &                     0.58
            &                     0.64
            &                     29.06
            &                     16.79
            &                     83.85
            &                     0.46
            &                     0.69
 \\

    & \textcolor{ourscolor}{\textbf{\falign}} 

            &                     \textbf{2.09}
            &                     \textbf{6.17}
            &                     \textbf{260.97}
            &                     \textbf{0.74}
            &                     \textbf{0.70}
            &                     \textbf{2.67}
            &                     \textbf{4.86}
            &                     \textbf{222.71}
            &                     \textbf{0.74}
            &                     0.69
            &                     \textbf{6.26}
            &                     \textbf{6.14}
            &                     \textbf{159.14}
            &                     \textbf{0.68}
            &                     \textbf{0.70}
 \\



\midrule
\multirow{3}{*}{REPA-E}
    & -

            &                     15.07
            &                     4.46
            &                     55.14
            &                     0.65
            &                     \textbf{0.69}
            &                     15.07
            &                     4.46
            &                     55.14
            &                     0.65
            &                     \textbf{0.69}
            &                     15.07
            &                     \textbf{4.46}
            &                     55.14
            &                     0.65
            &                     \textbf{0.69}
 \\

    & \textcolor{ourscolor}{\textbf{\fdino}}

            &                     \textbf{1.45}
            &                     \textbf{4.07}
            &                     \textbf{264.69}
            &                     \textbf{0.76}
            &                     \textbf{0.69}
            &                     2.30
            &                     4.84
            &                     212.25
            &                     0.75
            &                     0.67
            &                     3.24
            &                     5.20
            &                     188.92
            &                     0.73
            &                     0.67
 \\
 
    & \textcolor{ourscolor}{\textbf{\falign}}

            &                     2.15
            &                     6.64
            &                     264.48
            &                     0.75
            &                     0.68
            &                     \textbf{1.79}
            &                     \textbf{4.13}
            &                     \textbf{237.37}
            &                     \textbf{0.76}
            &                     0.68
            &                     \textbf{2.50}
            &                     4.81
            &                     \textbf{201.13}
            &                     \textbf{0.74}
            &                     \textbf{0.69}
 \\



\bottomrule
\end{tabular}}
\end{table*}

\begin{table*}[t]
\centering
\caption{\textbf{Instance-level comparison of \ours generations on ImageNet \cite{deng2009imagenet}.} Setup is the same as in Table~\ref{tab:ipa_imagenet_fids}. Alignment is measured with DINOv2 \cite{oquab2024dinov2}, JEPA \cite{jepa}, and CLIP \cite{radford2021learning} feature spaces, supplemented by PSNR for pixel-level fidelity. We highlight ours conditioning methods with \textcolor{ourscolor}{\textbf{\fbase}} and \textcolor{ourscolor}{\textbf{\fdino}}.
}
\label{tab:ipa_imagenet_sim}

\resizebox{\linewidth}{!}{%
\begin{tabular}{ll|cccc|cccc|cccc}
\toprule
& 
& \multicolumn{4}{c|}{\textbf{Full Feature Map}}
& \multicolumn{4}{c|}{\textbf{Masked Feature Map}}
& \multicolumn{4}{c}{\textbf{Average Feature Map}} \\
 \cmidrule(lr){3-6} \cmidrule(lr){7-10} \cmidrule(lr){11-14}

\textbf{Model} & \textbf{Cond.}
& \textbf{DINOv2} & \textbf{JEPA} & \textbf{CLIP} & \textbf{PSNR}
& \textbf{DINOv2} & \textbf{JEPA} & \textbf{CLIP} & \textbf{PSNR}
& \textbf{DINOv2} & \textbf{JEPA} & \textbf{CLIP} & \textbf{PSNR} \\
\midrule

\multirow{ 1 }{*}{SiT}
    
 &
    \fbase 

            & 
                    { 0.27 }
            & 
                    { 0.36 }
            & 
                    { 0.43 }
            & 
                    { 15.35 }
            & 
                    { 0.44 }
            & 
                   { 0.48 }
            & 
                    { 0.46 }
            & 
                    { 17.98 }
            & 
                   { 0.12 }
            & 
                    { 0.35 }
            & 
                    { 0.83 }
            & 
                    { 7.74 }
    \\
\midrule
\multirow{ 2 }{*}{REPA}

 &
    \textcolor{ourscolor}{\textbf{\fdino}} 

            & 
                    0.75
            & 
                    0.61
            & 
                    0.58
            & 
                    11.02
            & 
                    0.77
            & 
                    0.60
            & 
                    0.57
            & 
                    10.63
            & 
                    0.76
            & 
                    0.69
            & 
                    0.92
            & 
                    8.00
    \\
&
    \textcolor{ourscolor}{\textbf{\falign}}

            & 
                    \textbf{ 0.85 }
            & 
                    \textbf{ 0.78 }
            & 
                    \textbf{ 0.71 }
            & 
                    \textbf{ 20.46 }
            & 
                    \textbf{ 0.87 }
            & 
                    \textbf{ 0.79 }
            & 
                    \textbf{ 0.71 }
            & 
                    \textbf{ 20.72 }
            & 
                    \textbf{ 0.84 }
            & 
                    \textbf{ 0.84 }
            & 
                    \textbf{ 0.95 }
            & 
                    \textbf{10.82}
    \\

\midrule
\multirow{ 2 }{*}{REPA-E}
    
  &
    \textcolor{ourscolor}{\textbf{\fdino}} 

            & 
                    0.83
            & 
                    0.69
            & 
                    0.64
            & 
                    15.23
            & 
                    0.85
            & 
                    0.68
            & 
                    0.63
            & 
                    14.36
            & 
                    \textbf{ 0.91 }
            & 
                    0.83
            & 
                    \textbf{0.96}
            & 
                    10.97
    \\
 &
    \textcolor{ourscolor}{\textbf{\falign}}

            & 
                    \textbf{ 0.86 }
            & 
                    \textbf{ 0.76 }
            & 
                    \textbf{ 0.71 }
            & 
                    \textbf{ 18.68 }
            & 
                    \textbf{ 0.88 }
            & 
                    \textbf{ 0.78 }
            & 
                    \textbf{ 0.71 }
            & 
                    \textbf{ 19.07 }
            & 
                   \textbf{ 0.91}
            & 
                    \textbf{ 0.88 }
            & 
                    \textbf{ 0.96 }
            & 
                    \textbf{ 11.96 }
    \\


\bottomrule
\end{tabular}}
\end{table*}
\begin{table*}[t]
\centering
\caption{\textbf{Distribution-level zero-shot evaluation on the COCO~\cite{coco} dataset.} The setup is the same as in Table~\ref{tab:ipa_imagenet_fids}. Within each block, the first row reports unconditional generation results as a baseline, while \textcolor{ourscolor}{\textbf{\fbase}} and \textcolor{ourscolor}{\textbf{\fdino}} denote our method.}
\label{tab:ipa_coco_fids}

\resizebox{\linewidth}{!}{%
\begin{tabular}{ll|ccccc|ccccc|ccccc}
\toprule
& & \multicolumn{5}{c|}{\textbf{Full Feature Map}}
& \multicolumn{5}{c|}{\textbf{Masked Feature Map}}
& \multicolumn{5}{c}{\textbf{Average Feature Map}} \\
 \cmidrule(lr){3-7} \cmidrule(lr){8-12} \cmidrule(lr){13-17}

\textbf{Model} & \textbf{Cond.}
& \textbf{FID}$\downarrow$ & \textbf{sFID}$\downarrow$ & \textbf{IS}$\uparrow$ & \textbf{Prec.}$\uparrow$ & \textbf{Rec.}$\uparrow$
& \textbf{FID}$\downarrow$ & \textbf{sFID}$\downarrow$ & \textbf{IS}$\uparrow$ & \textbf{Prec.}$\uparrow$ & \textbf{Rec.}$\uparrow$
& \textbf{FID}$\downarrow$ & \textbf{sFID}$\downarrow$ & \textbf{IS}$\uparrow$ & \textbf{Prec.}$\uparrow$ & \textbf{Rec.}$\uparrow$ \\
\midrule

\multirow{2}{*}{SiT}
    
    & -

            &  \textbf{  45.13  }               
            &  \textbf{  30.15  }            
            &  \textbf{  22.33  }               
            &  \textbf{  0.42   }              
            &  \textbf{  0.54   }           
            &  \textbf{  45.13  }               
            &  \textbf{  30.15  }            
            &  \textbf{  22.33  }               
            &  \textbf{  0.42   }              
            &  \textbf{  0.54   }  
            &  \textbf{  45.13  }               
            &  \textbf{  30.15  }            
            &  \textbf{  22.33  }               
            &  \textbf{  0.42   }              
            &  \textbf{  0.54   }  
 \\

    & \fbase

            &                     {77.26}
            &                     {108.79}
            &                     {9.86}
            &                     {0.25}
            &                     {0.37}
            &                     {51.88}
            &                     {42.36}
            &                     {15.76}
            &                     {0.27}
            &                     {0.52}
            &                     {66.09}
            &                     {46.88}
            &                     {16.67}
            &                     {0.28}
            &                     {0.45}
 \\

\midrule
\multirow{3}{*}{REPA}
   
    & -

            &                    37.85
            &                    29.11
            &                    \textbf{41.31}
            &                    0.45
            &                    0.58
            &                    37.85
            &                    29.11
            &                    \textbf{41.31}
            &                    0.45
            &                    0.58
            &                    37.85
            &                    29.11
            &                    \textbf{41.31}
            &                    0.45
            &                    0.58
 \\

     & \textcolor{ourscolor}{\textbf{\fdino}}

            &                     12.96
            &                     29.05
            &                     31.11
            &                     0.55
            &                     0.57
            &                     20.51
            &                     31.04
            &                     26.15
            &                     0.45
            &                     0.53
            &                     32.04
            &                     39.05
            &                     20.59
            &                     0.35
            &                     0.53
 \\

    & \textcolor{ourscolor}{\textbf{\falign}}

            &                     \textbf{6.18}
            &                     \textbf{25.35}
            &                     33.08
            &                     \textbf{0.65}
            &                     \textbf{0.62}
            &                     \textbf{6.63}
            &                     \textbf{24.37}
            &                     32.05
            &                     \textbf{0.65}
            &                     \textbf{0.61}
            &                     \textbf{9.46}
            &                     \textbf{25.89}
            &                     29.43
            &                     \textbf{0.60}
            &                     \textbf{0.60}
 \\



\midrule
\multirow{3}{*}{REPA-E}
 
    & -

            &                    36.23
            &                    29.98
            &                    \textbf{55.13}
            &                    0.51
            &                    0.59
            &                    36.23
            &                    29.98
            &                    \textbf{55.13}
            &                    0.51
            &                    0.59
            &                    36.23
            &                    29.98
            &                    \textbf{55.13}
            &                    0.51
            &                    0.59
 \\

    & \textcolor{ourscolor}{\textbf{\fdino}}

            &                     \textbf{4.61}
            &                     \textbf{23.27}
            &                     35.97
            &                     \textbf{0.67}
            &                     \textbf{0.64}
            &                     5.65
            &                     23.35
            &                     34.54
            &                     0.64
            &                     0.63
            &                     6.07
            &                     24.09
            &                     32.76
            &                     0.63
            &                     \textbf{0.63}
 \\

    & \textcolor{ourscolor}{\textbf{\falign}}

            &                     5.70
            &                     25.55
            &                     34.29
            &                     0.66
            &                     0.62
            &                     \textbf{4.92}
            &                     \textbf{23.03}
            &                     34.78
            &                     \textbf{0.66}
            &                     \textbf{0.63}
            &                     \textbf{5.45}
            &                     \textbf{23.98}
            &                     33.81
            &                     \textbf{0.64}
            &                     \textbf{0.63}
 \\



\bottomrule
\end{tabular}}
\end{table*}

\textbf{Distribution-level evaluation.}
We first evaluate the generated outputs at the distribution level using standard metrics, FID \cite{heusel2017gans}, sFID \cite{nash2021generating}, IS \cite{salimans2016improved}, Precision, and Recall \cite{kynkaanniemi2019improved}. Distribution level evaluation is non-trivial, as a clear target conditional density is unavailable. To address this, we approximate the unconditional target distribution by aggregating the conditional densities across anchor features. We consider three feature granularities: $(i)$ full feature maps, $(ii)$ masked feature maps, and $(iii)$ average features. These conditioning signals are extracted from 50,000 anchor images sampled uniformly across all classes from the ImageNet training set. Using these anchor features, we generate 50,000 images with \ours and compare them against the whole dataset.

We report results on ImageNet (Table \ref{tab:ipa_imagenet_fids}) and present a zero-shot evaluation on COCO (Table \ref{tab:ipa_coco_fids}) using the same models. The first rows show that SiT does not respond well on \ours due to an ambiguous feature space, as the FID increases with IPA. In contrast, REPA and REPA-E benefit from feature alignment, allowing them to better react to the conditioning across all granularity levels. 

\textbf{Instance-level evaluation.}
To verify that the generated samples follow the given conditions, we evaluate quality at the instance level using feature extractors DINOv2 \cite{oquab2024dinov2}, JEPA \cite{jepa}, and CLIP \cite{radford2021learning}. We compute the average alignment score (i.e., cosine similarity) between the features of each anchor image and the corresponding generated image across three granularities. For the average feature map, alignment is calculated between global average features; for the full feature map, it is computed as the mean of all patch-wise similarities; and for the masked setup, it is the mean of patch similarities within the masked region. Furthermore, we report PSNR to measure the low-level pixel fidelity. 

\begin{figure*}[t]
\centering

\setlength{\fboxsep}{0pt} 
\setlength{\fboxrule}{1pt}  

\setlength{\tabcolsep}{1pt} 
\renewcommand{\arraystretch}{1.0}
\begin{minipage}[c]{0.49\linewidth}
\begin{tabular}{@{}ccccc@{}}
\textbf{Anchor} & \textbf{Mask} & \textbf{Full} & \textbf{Masked} & \textbf{Average} \\
\makecell{
\includegraphics[width=0.19\textwidth]{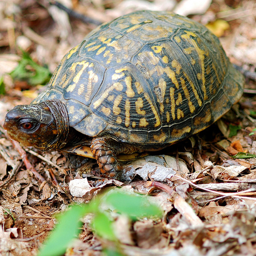}\\
\includegraphics[width=0.19\textwidth]{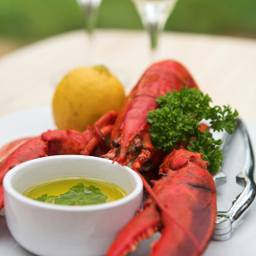}
} &
\makecell{
\includegraphics[width=0.19\textwidth]{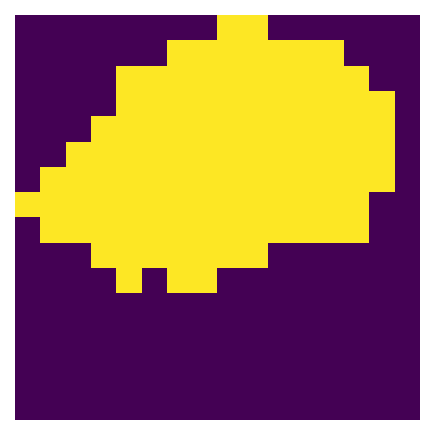}\\
\includegraphics[width=0.19\textwidth]{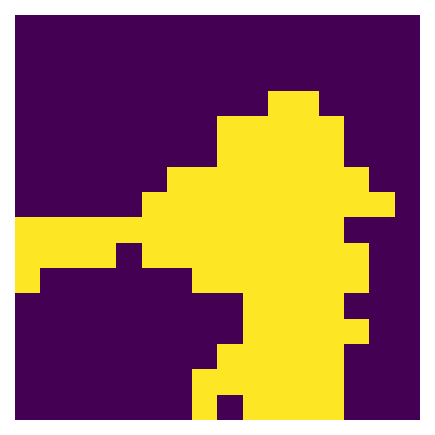}
} &
\makecell{
\includegraphics[width=0.19\textwidth]{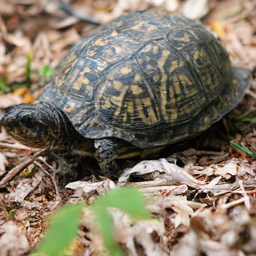}\\
\includegraphics[width=0.19\textwidth]{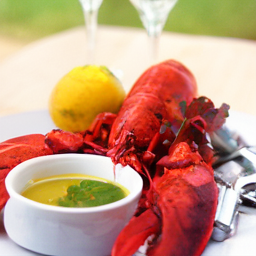}
} &
\makecell{
\includegraphics[width=0.19\textwidth]{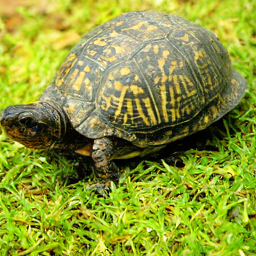}\\
\includegraphics[width=0.19\textwidth]{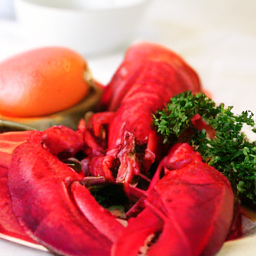}
} &
\makecell{
\includegraphics[width=0.19\textwidth]{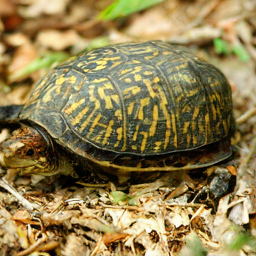}\\
\includegraphics[width=0.19\textwidth]{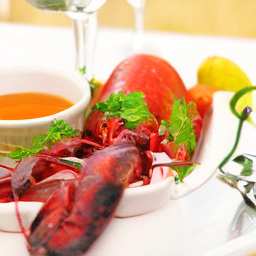}
} \\
\makecell{
\includegraphics[width=0.19\textwidth]{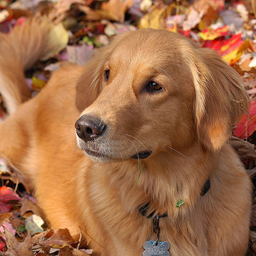}\\
\includegraphics[width=0.19\textwidth]{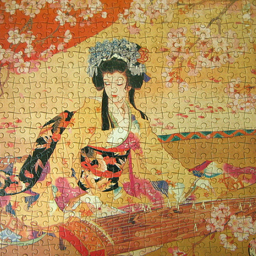}
} &
\makecell{
\includegraphics[width=0.19\textwidth]{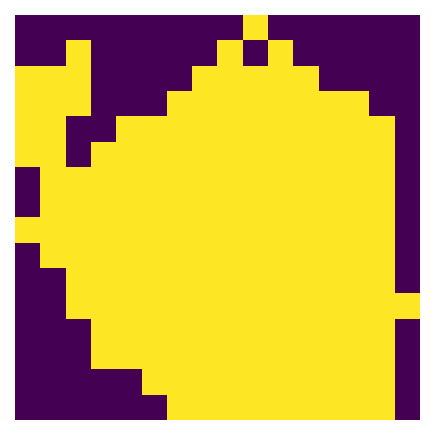}\\
\includegraphics[width=0.19\textwidth]{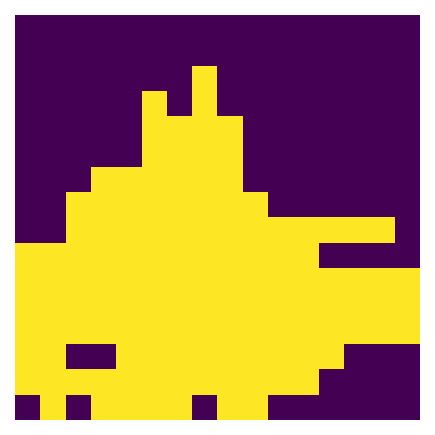}
} &
\makecell{
\includegraphics[width=0.19\textwidth]{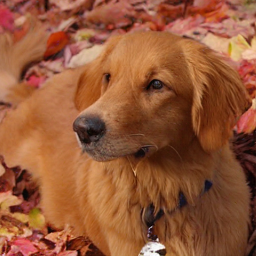}\\
\includegraphics[width=0.19\textwidth]{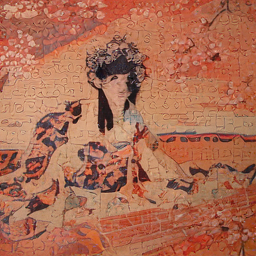}
} &
\makecell{
\includegraphics[width=0.19\textwidth]{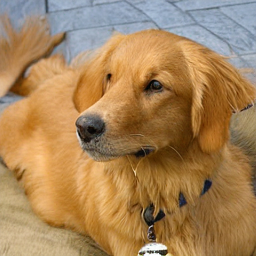}\\
\includegraphics[width=0.19\textwidth]{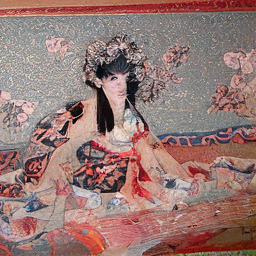}
} &
\makecell{
\includegraphics[width=0.19\textwidth]{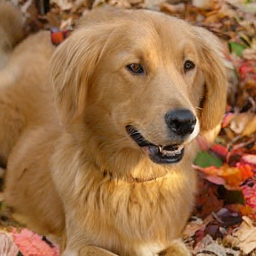}\\
\includegraphics[width=0.19\textwidth]{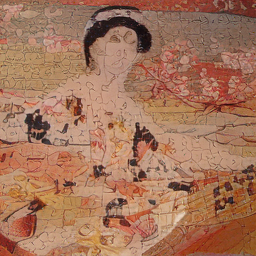}
}
\end{tabular}
\end{minipage}%
\hfill
\begin{minipage}[c]{0.49\linewidth}
\begin{minipage}[c]{0.10\textwidth}
    \includegraphics[width=\textwidth]{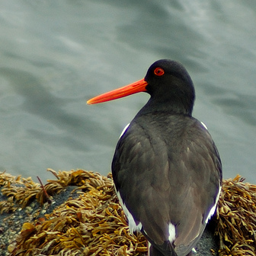}\\
    \includegraphics[width=\textwidth]{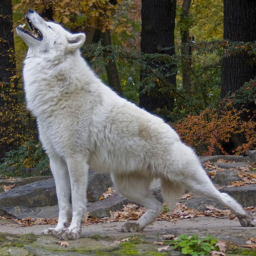}
\end{minipage}
\hspace{-1.5pt}
\begin{minipage}[c]{0.20\textwidth}
    \includegraphics[width=\textwidth]{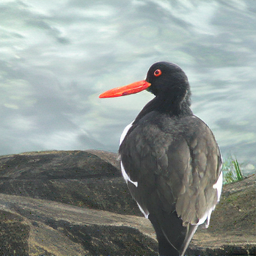}
\end{minipage}
\begin{minipage}[c]{0.10\textwidth}
    \includegraphics[width=\textwidth]{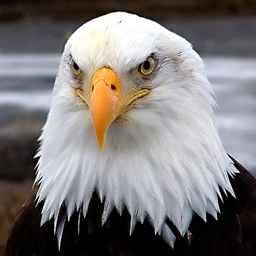}\\
    \includegraphics[width=\textwidth]{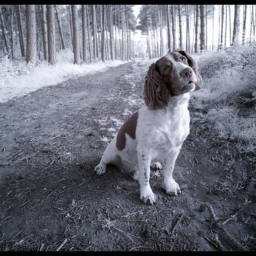}
\end{minipage}
\hspace{-1.5pt}
\begin{minipage}[c]{0.2\textwidth}
    \includegraphics[width=\textwidth]{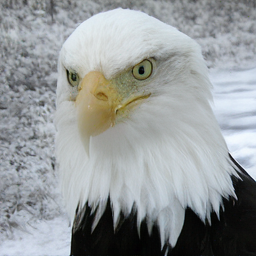}
\end{minipage}
\begin{minipage}[c]{0.10\textwidth}
    \includegraphics[width=\textwidth]{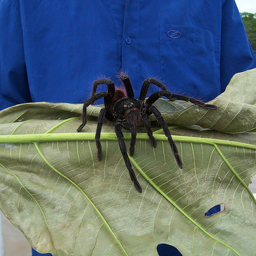}\\
    \includegraphics[width=\textwidth]{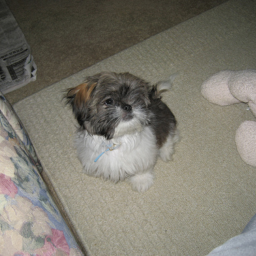}
\end{minipage}
\hspace{-1.5pt}
\begin{minipage}[c]{0.20\textwidth}
    \includegraphics[width=\textwidth]{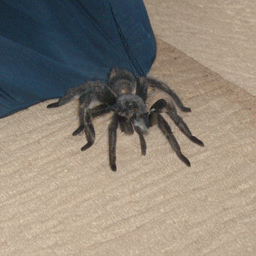}
\end{minipage}

\begin{minipage}[c]{0.10\textwidth}
    \includegraphics[width=\textwidth]{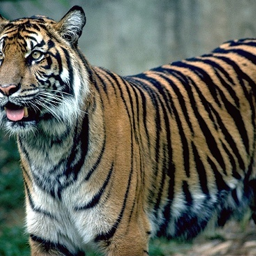}\\
    \includegraphics[width=\textwidth]{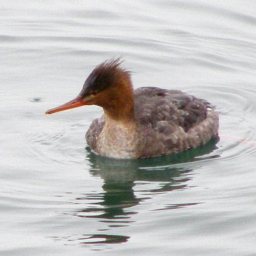}
\end{minipage}
\hspace{-1.5pt}
\begin{minipage}[c]{0.2\textwidth}
    \includegraphics[width=\textwidth]{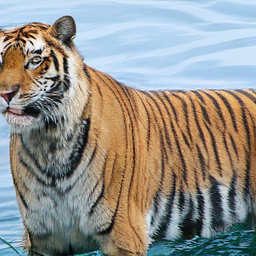}
\end{minipage}
\begin{minipage}[c]{0.10\textwidth}
    \includegraphics[width=\textwidth]{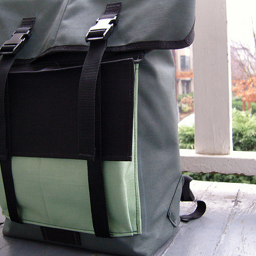}\\
    \includegraphics[width=\textwidth]{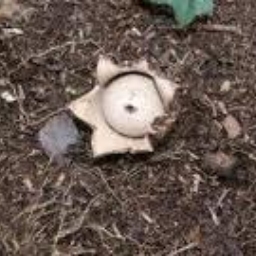}
\end{minipage}
\hspace{-1.5pt}
\begin{minipage}[c]{0.2\textwidth}
    \includegraphics[width=\textwidth]{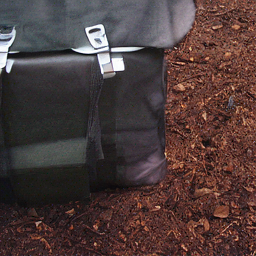}
\end{minipage}
\begin{minipage}[c]{0.10\textwidth}
    \includegraphics[width=\textwidth]{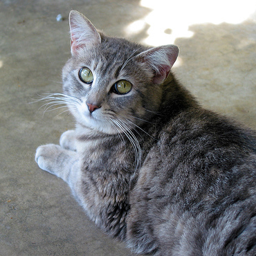}\\
    \includegraphics[width=\textwidth]{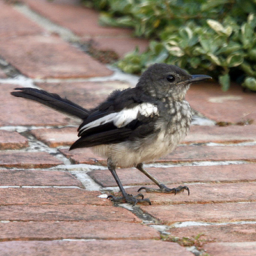}
\end{minipage}
\hspace{-1.5pt}
\begin{minipage}[c]{0.2\textwidth}
    \includegraphics[width=\linewidth]{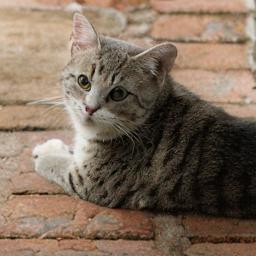}
\end{minipage}

\vspace{1pt}
\begin{minipage}[c]{0.10\textwidth}
    \includegraphics[width=\linewidth]{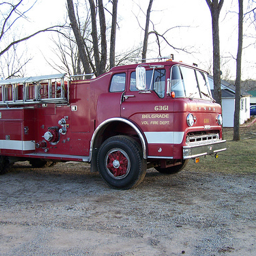}\\
    \includegraphics[width=\linewidth]{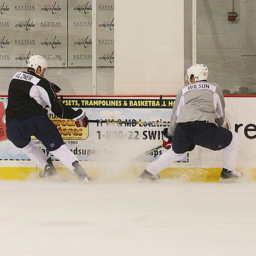}
\end{minipage}
\hspace{-1.5pt}
\begin{minipage}[c]{0.2\textwidth}
    \includegraphics[width=\linewidth]{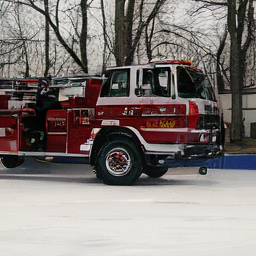}
\end{minipage}
\begin{minipage}[c]{0.10\textwidth}
    \includegraphics[width=\linewidth]{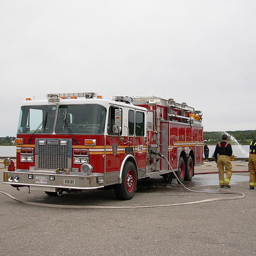}\\
    \includegraphics[width=\linewidth]{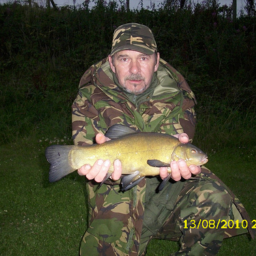}
\end{minipage}
\hspace{-1.5pt}
\begin{minipage}[c]{0.2\textwidth}
    \includegraphics[width=\linewidth]{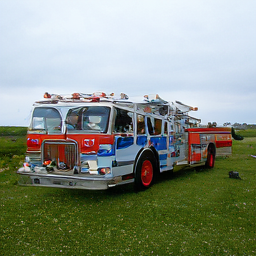}
\end{minipage}
\begin{minipage}[c]{0.10\textwidth}
    \includegraphics[width=\linewidth]{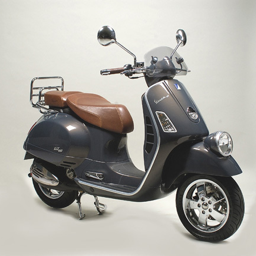}\\
    \includegraphics[width=\linewidth]{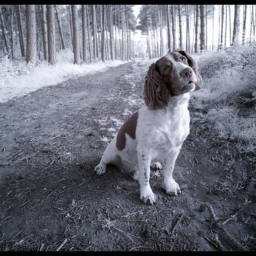}
\end{minipage}
\hspace{-1.5pt}
\begin{minipage}[c]{0.2\textwidth}
    \includegraphics[width=\linewidth]{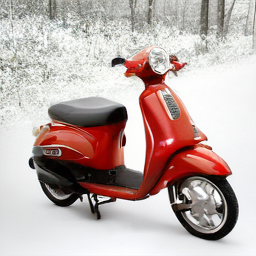}
\end{minipage}

\begin{minipage}[c]{0.10\textwidth}
    \includegraphics[width=\linewidth]{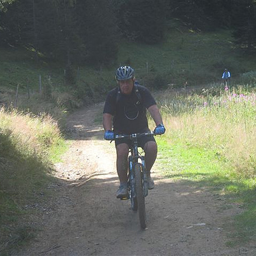}\\
    \includegraphics[width=\linewidth]{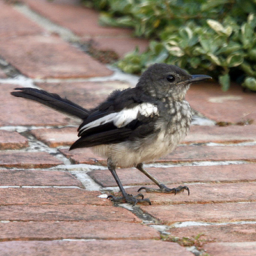}
\end{minipage}
\hspace{-1.5pt}
\begin{minipage}[c]{0.2\textwidth}
    \includegraphics[width=\linewidth]{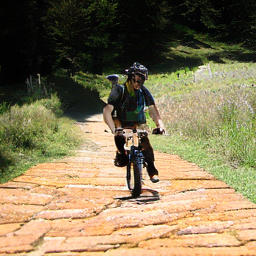}
\end{minipage}
\begin{minipage}[c]{0.10\textwidth}
    \includegraphics[width=\linewidth]{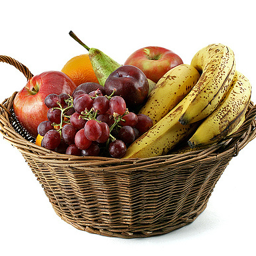}\\
    \includegraphics[width=\linewidth]{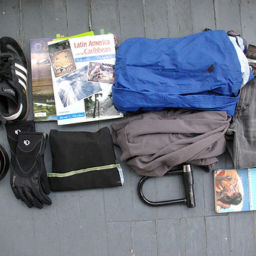}
\end{minipage}
\hspace{-1.5pt}
\begin{minipage}[c]{0.2\textwidth}
    \includegraphics[width=\linewidth]{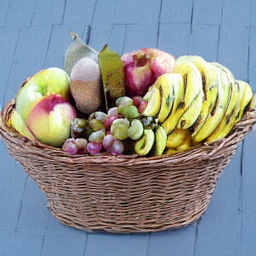}
\end{minipage}
\begin{minipage}[c]{0.10\textwidth}
    \includegraphics[width=\linewidth]{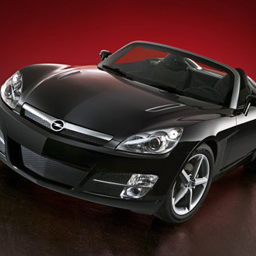}\\
    \includegraphics[width=\linewidth]{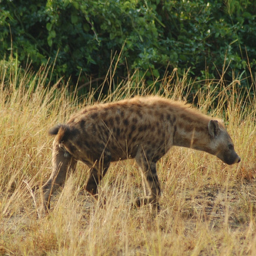}
\end{minipage}
\hspace{-1.5pt}
\begin{minipage}[c]{0.2\textwidth}
    \includegraphics[width=\linewidth]{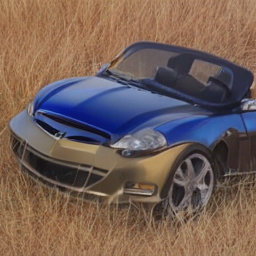}
\end{minipage}

\end{minipage}

\caption{\textbf{Qualitative comparison on ImageNet \cite{deng2009imagenet} using REPA-E \cite{leng_repa-e_2025} with DINOv2 features.} \textbf{(Left)} We show single-source conditioning across three levels of granularity: full, masked, and averaged feature maps given \emph{anchor image} and \emph{input mask}. \textbf{(Right)} We illustrate multi-source composition. Within each group, the anchor object and target background are shown at a small scale alongside the enlarged generation. We use full features from the anchor and a single feature patch sampled from the target's background.}\label{fig:main_quals}
\vspace{-5mm}
\end{figure*}

The results in Table~\ref{tab:ipa_imagenet_sim} indicate that achieving high similarity to the anchor images requires a well-structured representation space. The vanilla SiT model fails to follow the anchor images, whereas REPA and REPA-E nearly perfectly reconstruct the condition across all levels of granularity. Figure~\ref{fig:main_quals} (left) presents qualitative results of IPA, demonstrating that our method effectively conditions generation across each setting (additional examples provided in \appref{app:more_qual}).

\textbf{Information density of visual features and text.} Visual features are expected to provide more information than long and detailled caption. To confirm this intuition, we compare \ours against the text-to-image (T2I) model CAD-I \cite{cad} as a baseline in Table~\ref{tab:t2i}. Unlike most T2I models trained on massive, scene-centric datasets, CAD-I is specifically trained on ImageNet. We use long captions extracted by \cite{cad} from the COCO validation set as conditioning signals for CAD-I to generate the described scenes. For our method, we utilize both full and average feature maps extracted from the internal SiT representations before the projection layer trained with REPA-E to generate the corresponding images. The performance gap between text-to-image and visual feature conditioning shows that visual features provide as expected a denser, more informative signal than text captions.

\begin{table}[t]
\centering
\begin{minipage}[t]{0.48\linewidth}
\centering
\caption{\textbf{Comparison with text-to-image (T2I) generation.} We compare our method against CAD-I \cite{cad}, an ImageNet-trained T2I baseline, using COCO validation captions as conditioning signals. Our method utilizes full and average feature maps extracted from REPA-E \cite{leng_repa-e_2025}.}
\label{tab:t2i}
\renewcommand{\arraystretch}{1.2}
\resizebox{\linewidth}{!}{%
\begin{tabular}{llccccc}
\toprule
\textbf{Model} &
\textbf{Cond.} &
\textbf{FID $\downarrow$} &
\textbf{sFID $\downarrow$} &
\textbf{IS $\uparrow$} &
\textbf{Prec. $\uparrow$} &
\textbf{Rec. $\uparrow$} \\
\midrule
CAD-I  & Text         & 37.98 & 29.87& 25.60 & 0.46 & 0.37 \\
REPA-E  & Avg. Feat.  & 3.58 & 11.10 & 32.69 & 0.75 & 0.78 \\
REPA-E & Full Feat.  & \textbf{2.11} & \textbf{8.35}  & \textbf{36.01} &  \textbf{0.98} & \textbf{0.96}\\
\bottomrule
\end{tabular}}
\end{minipage}\hfill
\begin{minipage}[t]{0.48\linewidth}
\centering
\caption{\textbf{Compositional generation evaluation on ImageNet~\cite{deng2009imagenet}.} We assess visual quality with PickScore and concept adherence using CLIP, and compare \ours variants using IPA and SPA against standard baselines.}
\label{tab:multipot-metrics}
\resizebox{\linewidth}{!}{%
\begin{tabular}{lccccc}
\toprule
\multirow{2}{*}{\textbf{Cond.}} & \textbf{Pick} & \textbf{CLIP} & \textbf{Target}& \textbf{Anchor} & \textbf{Combined} \\
& \textbf{Score} & \textbf{Score} & \textbf{Sim.}& \textbf{Sim.} & \textbf{Sim.} \\
\midrule
Uncond. & 0.175  & 0.100  & 0.098 & 0.113 & 0.105 \\ %
Class Cond. & \textbf{0.198 } & 0.220  & 0.110 & 0.243  & 0.177 \\ 
Interp. & 0.188 & 0.190  & 0.156  & 0.183 & 0.170 \\
\midrule
IPA\textrightarrow IPA & \uline{0.196 } & \uline{0.232} & \uline{0.372} & \textbf{0.452} & \uline{0.412} \\
IPA\textrightarrow SPA & \uline{0.196} & \textbf{0.238} & \textbf{0.510} & \uline{0.422 } & \textbf{0.466} \\
\bottomrule
\end{tabular}
}
\end{minipage}
\end{table}

\textbf{Resolutions scalability.}
Although 256$\times$256 resolution is the standard benchmark in generative modeling research, we additionally fine-tune our 256$\times$256 model to generate 512$\times$512 outputs for 40k steps in order to evaluate scalability to higher resolutions. The results in \appref{app:scalability} confirm that our method scales effectively to higher resolutions, reducing FID from 33 to 5. 

\textbf{Generalization across semantic vision encoders.}
To assess whether \ours depends on a specific semantic vision encoder, we evaluate it on three SiT-XL/2 models trained with REPA and aligned with DINOv2, CLIP, and Perception Encoder~\cite{bolya2025PerceptionEncoder} (PE), respectively. As shown in Table~\ref{tab:backbones}, \ours consistently improves generation quality across all three representation spaces. Notably, these models are trained for only 400k steps, 10$\times$ fewer than our main models. This indicates that effective conditioning does not require fully converged representation alignment; once the internal features acquire sufficient semantic organization, they already provide a useful guidance signal.

\textbf{Generalization beyond REPA-style models.}
Our guidance mechanism is not inherently restricted to models trained with a REPA-style objective. Rather, it requires internal representations that form a semantically organized and sufficiently noise-robust feature space, such that feature similarity provides a meaningful guidance signal. To support this broader applicability, we evaluate our method on Self-Flow~\citep{CheferEsser2026selfflow}, which learns semantic internal representations without relying on the standard REPA objective or an additional pretrained self-supervised backbone. Self-Flow employs a teacher--student framework with asynchronous noise timesteps, where the student predicts representations produced by the teacher from a less corrupted input. Our guidance transfers directly to these learned representations, and, as shown in Table~\ref{tab:selfflow}, full-map guidance substantially improves generation quality. These results indicate that the proposed guidance is not tied to a particular representation-alignment objective, but can be applied more broadly when the generative model provides a semantically structured internal feature space. We provide qualitative results in Figure \ref{fig:tfg_visual}.

\begin{table}[tbh]
    \centering
    \begin{minipage}[t]{0.48\linewidth}
    \caption{\textbf{Generalization across semantic alignment backbones.} We present the FID and DINOv2 Alignment scores for REPA models aligned with CLIP, PE, and DINOv2, evaluating unconditional generation and different feature-conditioning granularities (Full, Average, Masked).}
    \resizebox{\textwidth}{!}{
    \begin{tabular}{lcccccc}
        \toprule
        \multirow{2}{*}{\textbf{Cond.}} & \multicolumn{3}{c}{\textbf{FID $\downarrow$}} & \multicolumn{3}{c}{\textbf{DINOv2 Align. $\uparrow$}} \\
        \cmidrule(lr){2-4} \cmidrule(lr){5-7}
        & \textbf{CLIP} & \textbf{PE} & \textbf{DINOv2} & \textbf{CLIP} & \textbf{PE} & \textbf{DINOv2} \\
        \midrule
        - & 40.0 & 44.6 & 36.2 & - & - & - \\
        Full & 8.3 & 22.2 & 4.8 & 0.5 & 0.4 & 0.7 \\
        Average & 18.7 & 19.1 & 9.8 & 0.6 & 0.6 & 0.7 \\
        Masked & 12.0 & 18.1 & 20.6 & 0.6 & 0.4 & 0.7 \\
        \bottomrule
    \end{tabular}}
    \label{tab:backbones}
    \end{minipage}\hfill
    \begin{minipage}[t]{0.48\linewidth}
    \caption{\textbf{Generalization beyond REPA-style models.} We apply REPA-G on Self-Flow~\cite{CheferEsser2026selfflow} with full feature map conditioning and show that it can extend to model trained with non REPA-style objective provided that the learned internal representation is semantic and noise-robust.}
    \resizebox{\textwidth}{!}{
\begin{tabular}{l r r r r r r}
    \toprule
    \textbf{Cond.} & \textbf{FID}$\downarrow$ & \textbf{sFID}$\downarrow$ & \textbf{IS}$\uparrow$ & \textbf{Prec.}$\uparrow$ & \textbf{Rec.}$\uparrow$ & \makecell{\textbf{DINOv2} \\ \textbf{Align.}$\uparrow$} \\
    \midrule
    \multicolumn{7}{l}{\textbf{\textit{REPA-E}}} \\
    - & 15.07 & 4.46 & 55.14 & 0.65 & 0.69 & - \\
    Full Feat. & 2.15 & 6.64 & 264.48 & 0.75 & 0.68 & 0.86 \\
    \midrule
    \multicolumn{7}{l}{\textbf{\textit{Self-Flow}}} \\
    - & 35.51 & 5.79 & 27.62 & 0.55 & 0.63 & - \\
    Full Feat. & 5.49 & 4.71 & 161.02 & 0.71 & 0.68 & 0.72 \\
    \bottomrule
\end{tabular}}
    \label{tab:selfflow}
    \end{minipage}

\end{table}

\textbf{Comparison to training-free baseline.}
To the best of our knowledge, TFG~\citep{ye_tfg_2024} is the only fair baseline that, similar to \ours, does not require additional parameters or training steps. We adapt the official implementation of TFG to support feature-based guidance, using DINOv2~\citep{oquab2024dinov2} as the target predictor.
However, in our setup, TFG does not produce images of sufficient visual quality, with FID scores around 80. We provide the full experimental setup and results in \appref{app:tfg}. 

\textbf{Comparison with adapter-based conditioning.}
IP-Adapter-style methods provide a relevant reference-based generation baseline, but differ substantially from \ours in both their assumptions and training requirements. Despite requiring additional training, IP-Adapter \cite{ye2023ip-adapter} yields an FID that is 6 points higher, 25\% lower DINOv2 alignment on ImageNet, and lower generation throughput than \ours. Further details are provided in \appref{app:adapter}.

\subsection{Test-time guidance with multiple sources}

We evaluate concept-blending by composing anchor images (source objects) with target features (backgrounds). We apply two simultaneous conditioning: first, an object-level IPA using the anchor's DINO features, and a target background condition implemented via IPA or SPA. In this multi-source setting, in every denoising step, we sum up both potentials and compute once the gradient of the resulting joint potential that incorporates both the anchor and the target.
 The resulting images should preserve the anchor object while adopting the target background's characteristics. We evaluate on a 50-class ImageNet subset representing diverse objects. For each class, 100 random samples are combined with 8 hand-selected target background features, yielding 40,000 images per method. 

As novel image compositions are steering away from the real distribution, FID is inapplicable. We therefore evaluate image quality via Pick Score~\citep{kirstain2023pick} and concept adherence via CLIP Score~\citep{hessel2021clipscore}, using the prompt: \textit{An image of a/an [Anchor Class] with a [Target Class] background}. Additionally, we introduce patch-wise similarity metrics: each generated patch is assigned to the anchor or target based on maximum cosine similarity. We then compute Anchor, Target, and Combined similarities to quantify feature adherence to each source.
 
We compare our method against three baselines: $(i)$ unconditional, $(ii)$ class-conditional (using the anchor class), and $(iii)$ interpolated conditional generation, which blends class embeddings via adapted Classifier-Free Guidance. Our compositional conditioning is evaluated using two variants: IPA followed by IPA (IPA\textrightarrow IPA) and IPA followed by SPA (IPA\textrightarrow SPA).
As shown in~\autoref{tab:multipot-metrics}, \ours significantly outperforms baselines in prompt alignment and image quality. While class-conditional baselines are limited by an inability to encode specific target backgrounds (e.g., a tiger with a water background), \ours leverages DINO features to achieve superior anchor and target alignment. Notably, the SPA variant is most effective at blending sources, yielding the highest CLIP Score (see \appref{app:patch_metrics_app}). Figure~\ref{fig:main_quals} (right) highlights that \ours enables multi-source composition.

\textbf{Hyperparameters.}
We run an hyparameter sweep for single and multi-source conditioning. 
Similar to classifier guidance~\cite{dhariwal2021diffusion}, the parameter $\lambda$ controls the strength of the conditioning. For single-source conditioning, we find that increasing $\lambda$ in \eqref{eq:ttvc} improves FID up to a saturation point, after which performance plateaus, while preserving both precision and recall thanks to the anchor. Results are shown in \appref{app:hyperparameter_sweep}.

For multi-source conditioning, we found a low optimal value for temperature $T=0.1$ indicating that SPA is indeed beneficial as a large temperature would recover the IPA\textrightarrow IPA variant. While in practice the user may have to tune the $\lambda$ of each energy to balance conditioning contributions, we provide a default setting used for the experiment with both guidance scale set to $\lambda=10{,}000$.

\textbf{Runtime.}
\ours predicts both the features and the velocity field in a single forward pass, with feature extraction occurring early in the network at the $8^{\text{th}}$ transformer block. This significantly reduces the cost of the backward pass making \ours only 26\% slower than unconditional generation and faster than other guidance mechanisms. We provide a detailed runtime analysis in \appref{app:runtime_analysis}.

\textbf{Reconstruction--generation trade-off.}
Full-map conditioning with large $\lambda$ approaches stochastic reconstruction, while weaker guidance or average-feature conditioning permits increasingly diverse generations. 
Reducing $\lambda$ or using average-feature conditioning relaxes spatial constraints and increases variation across samples from the same anchor, yielding a controllable reconstruction-generation trade-off.
We provide the full analysis, quantitative and qualitative results in \appref{app:reconstruction}.

\section{Conclusion}
\label{sec:conclusion}

We introduced \ours, an inference-time framework that leverages aligned self-supervised representations to condition diffusion models. By guiding the sampling process, \ours enables  semantic control without the need for retraining. 
Our work show that feature conditioning, while impracticable with a vanilla model, becomes possible with a well-aligned feature space and this without being the initial motivation for such a training. While REPA-style methods are primarily introduced for training, we show that they can also provide inference-time guidance at no additional cost.

\textbf{Limitations.} For multi-source composition, complex blending scenarios may require the user to tune the $\lambda$ hyperparameter, which introduces a slight overhead. 
Besides, more strongly conflicting conditions remain challenging: incompatible identities may blend, spatial or viewpoint constraints can induce competing gradients, one condition may dominate, pose control requires spatially corresponding features, and far out-of-distribution combinations may produce artifacts (see \appref{app:failures} for failure cases).
Addressing these limitations and integrating our guidance with orthogonal conditioning methods represent promising directions for future research.

\textbf{Impact statement.} This paper presents an image conditioning method which unlocks the full potential of image generation models, but these capabilities may be misused to produce harmful content and can amplify biases present in the training data.
\begin{ack}
We acknowledge EuroHPC Joint Undertaking for awarding the project ID EHPC-REG-2024R02-234 access to Karolina, Czech Republic. 
We thank the reviewers for their insightful comments.
\end{ack}
\nocite{*}

{
\small
\bibliographystyle{unsrtnat}
\bibliography{ref}
}

\newpage
\appendix
\newpage
\appendix
\onecolumn

This appendix supplements the main paper with additional details regarding;

\begin{itemize}
\item Complementary Details on \textit{Preliminaries} (\appref{app:proof_prelim})

\item Complementary Details on \textit{Guiding Generation at Inference} (\appref{apx:guidance}) 
\begin{itemize}
    \item Sampling from the tilted distribution (\appref{apx:proof_lemma})
    \item Experiments on toy data to compare with the tilted distribution (\appref{apx:toy})
    \item Our algorithm for test-time conditioning (\appref{app:algorithm})
\end{itemize}

\item Complementary Details on \textit{Properties of Representation Space}
\begin{itemize}
    \item Impact of representation alignment on ImageNet Clusters (\appref{app:clusteringimagenet})
    \item More technical details on Euclidean Embedding Property (\appref{apx:eep})
\end{itemize}

\item Complementary Details on Design of Potential $\mathscr{V}$ (\appref{app:more_potential})

\item More Analysis and Experimental Results (\appref{app:more_exps_analysis})
\begin{itemize}
    \item More on hyperparameters (\appref{app:hyperparameter_sweep})
    \item More results on unconditional generation (\appref{app:more_incond})
    \item Results on conditional generation  (\appref{app:more_cond})
    \item Analysis of the balance between stochastic reconstruction and generation diversity (\appref{app:reconstruction})
    \item Comparison to adapted training free guidance baseline with visual feature (\appref{app:tfg})
    \item Comparison with adapter-based conditioning (\appref{app:adapter})
    \item Results on high resolution generation (\appref{app:scalability})
    \item Detailed runtime analysis (\appref{app:runtime_analysis})
    \item More technical details on multiple-Source compositional generation (\appref{app:patch_metrics_app}).
    \item Failure cases for compositional generation (\appref{app:failures}).
    \item More qualitative results (\appref{app:more_qual}).
    
\end{itemize}
 \item{Licences of assets and codebase used (\appref{app:licenses})}
\end{itemize}

\section{Complementary details on \textit{Preliminaries} (Section \ref{sec:preliminary})}\label{app:proof_prelim}

\subsection{Flow matching training and sampling}
Let $p_{1} = \mathcal{N}(0, I)$ be a standard Gaussian prior on $\cX$. 
We define an interpolation process by sampling independently $x_{0} \sim p_{0}$ and $x_{1} \sim p_{1}$ and set $x_{t} = \alpha_{t} x_{0} + \sigma_{t} x_{1}$, with typically $\alpha_{t} = 1 - t$ and $\sigma_{t} = t$. 
For such a process, there exists a \textit{probability-flow} ordinary differential equation (PF-ODE) $\dot{x} = v^{\star}(x,t)$ with a velocity field $v^{\star}(x,t)$ such that the probability flow $p_t$ induced by the PF-ODE at time $t$ is the time-marginal density of $x_t$. Conditional flow models train a velocity field $v_{\theta}: (x, t) \in \cX \times [0,1] \rightarrow v_{\theta}(x, t) \in \cX$ with
\begin{equation}
\mathcal{L}^{\mathrm{diff}}(\theta) = \mathbb{E}_{t \sim \mathcal{U}(0,1), (x_0, x_1) \sim p_{0} \times p_{1}} \left[\left\|v_\theta\left(x_t, t\right)- \dot{x}_t \right\|_2^2\right]\;.
\label{eq:condition_flow}
\end{equation}
A necessary condition for $\mathcal{L}^{\mathrm{diff}}$ to reach a global optimum at $\theta = \theta^{\star}$ is that, for each $x$ and $t$, we have:
\begin{equation}
v_{\theta}(x,t) = v^{\star}\left(x, t\right) = \mathbb{E}_{(x_0, x_1) \sim p_{0} \times p_{1}}[\dot{x}_t \mid x_{t} = x]\,.
\label{eq:optima_flow}
\end{equation}
Using Tweedie's formula \cite{efron2011tweedie}, we can relate $v^{\star}$ to the \textit{score} $\nabla_x \log p_t(x)$ of the marginal distribution $p_{t}$ of $x_{t}$. In the particular case of $(\alpha_{t}, \sigma_{t}) = (1-t,t)$, we have: 
\begin{equation}
    v^{\star}\left(x, t\right) = -\frac{1}{1-t} x-\frac{t}{1-t} \nabla_x \log p_t(x)\;.
    \label{eq:score}
\end{equation}
To sample from $p_{0}$, one can draw $x_{1} \sim p_{1}$ and integrate the PF-ODE backward from $t=1$ to $t=0$ using, e.g., Euler's method. In practice, a Stochastic Sampler (e.g., Euler-Maruyama) can be used to improve sample quality during inference as in \cite{yu2025representation}, integrating a reverse-time stochastic differential equation (SDE) of the form
\begin{equation}
\mathrm{d}x_{t} = \left( v^{\star}(x_t, t) - t \nabla_x \log p_t(x_t) \right) \mathrm{d}t + \sqrt{2 t} \mathrm{d}\bar{W}_{t}\;,
\label{eq:sde}
\end{equation}
which shares the same marginals $p_{t}$ as the PF-ODE. The term in parentheses corresponds to the \textit{drift}, while the stochastic term $\sqrt{2 t}\,\mathrm{d}\bar{W}_t$ is the \textit{diffusion}, where $\bar{W}_t = W_{1-t}$ denotes a (time-reversed) standard Brownian motion. In \eqref{eq:sde}, $v^{\star}$ is replaced by $v_{\theta}$ and the estimated score $\nabla_x \log p_t(x)$ is deduced from \eqref{eq:score}, where $t$ is clipped to $\varepsilon > 0$ to avoid instabilities.

\section{Complementary details on \textit{Guiding generation at inference} (Section \ref{sec:guidance})}
\label{apx:guidance}

\subsection{Sampling the tilted distribution}\label{apx:proof_lemma}

General classifier guidance does not exactly target the tilted distribution~\cite{chidambaram_what_2024}, defined as:
\begin{equation}
\tilde{p}_{0}(x ; x_{c}) \propto p_{0}(x) e^{\lambda \mathscr{V}\left(\phi(x), \phi(x_{c})\right)}\;.
\label{eq:tilt_p}
\end{equation}

Below, we present two assumptions under which modifying the score function via the gradient of the potential $\mathscr{V}$, as done in \eqref{eq:ttvc}, would theoretically allow sampling from this tilted distribution \eqref{eq:tilt_p}. While our method operates within the classifier guidance framework and does not strictly require exact sampling from this density, we provide these assumptions as supplementary grounding for the score modification \eqref{eq:ttvc}. This complements the standard justification derived from Bayes' rule~\cite{song2021denoising}.

\subsubsection{Assumptions}
Mirroring \eqref{eq:optima_flow}, we can show the following proposition.
\begin{proposition}
    A necessary condition for \eqref{eq:alignment} to reach a global optimum when $\mathscr{V}$ is an average patch-wise similarity with unit length vectors is that for each $x \in \cX$ and $t \in [0,1]$:
\begin{equation}
(h_{\theta} \circ f_{\theta})(x, t) = \mathbb{E}_{(x_0,x_1) \sim p_0 \times p_1} \left[ \phi(x_{0}) \mid x_{t} = x \right]\;.
\label{eq:repa_opt}
\end{equation}
\label{prop:repa_opt}
\end{proposition}

We use the notation established in Section \ref{sec:preliminary}. Below, we provide the derivation for Proposition \ref{prop:repa_opt}, assuming $N=1$ without loss of generality.
\begin{manualproof}[Proof of proposition \ref{prop:repa_opt}] The potential $\mathscr{V}$ can be expressed in terms of the $L_2$ distance as follows:
\begin{align*}
\mathscr{V}\big( (h_{\theta} \circ f_{\theta})(x_{t}, t),\, \phi(x_0) \big) &= \langle (h_{\theta} \circ f_{\theta})(x_{t}, t),  \phi(x_0) \rangle \\
&= 1 - \frac{1}{2} \| (h_{\theta} \circ f_{\theta})(x_{t}, t) - \phi(x_0)  \|_{2}^{2}\,,
\end{align*}
where the second equality holds because the vectors in the scalar product are constrained to unit length. Consequently, minimizing the alignment loss $\mathcal{L}^{\mathrm{align}}$ defined in \eqref{eq:alignment} is equivalent to minimizing the following objective over $\theta$:
$$\mathbb{E}_{\substack{
t \sim \mathcal{U}(0,1) \\
(x_0, x_1) \sim p_{0} \times p_{1}
}}
\Big[
\| (h_{\theta} \circ f_{\theta})(x_{t}, t) - \phi(x_0)  \|_{2}^{2} \,
\Big] .$$
From the properties of conditional expectation in minimum mean square error (MMSE) estimation, this quantity is minimized when, for every $x \in \cX$ and $t \in [0,1]$:$$(h_{\theta} \circ f_{\theta})(x, t) = \mathbb{E}_{x_0, x_1 \sim p_0 \times p_1} [\phi(x_0) \mid (1 - t) x_{0} + t x_{1} = x]\;.$$
\end{manualproof}

\begin{assumption}[Global Optimality]
We assume the model trained with $\mathcal{L}^{\mathrm{diff}}(\theta) + \beta \mathcal{L}^{\mathrm{align}}(\theta)$ has reached a global optimum, such that the conditions \eqref{eq:optima_flow} and \eqref{eq:repa_opt} are satisfied.
\label{asm:local_opt}
\end{assumption}
\begin{assumption}[Vanishing Gap]
Let $Z_{t}(x ; x_{c}) \triangleq \mathbb{E}_{(x_{0},x_{1}) \sim p_{0} \times p_{1}} \left[ e^{ \lambda \mathscr{V}\left(\phi(x_{0}),\phi(x_c)\right)} \mid x_{t} = x \right]$ denotes the expected energy at state $x_{t}$. By Jensen's inequality:
\begin{equation}
\log Z_{t}(x ; x_{c}) \geq \lambda \mathbb{E}_{(x_{0},x_{1})} \left[ \mathscr{V}\left(\phi(x_{0}),\phi(x_c)\right) \mid x_{t} = x \right]\,.
\end{equation}
We assume this bound is tight (i.e., the Jensen Gap vanishes).
\label{asm:jensen_gap}
\end{assumption}

\subsubsection{Guidance of the process}

Let us consider the following forward time SDE
\begin{equation}
    \mathrm{d}x_{t} = \left( v^{\star}(x_t, t) + t \nabla_x \log p_t(x_t) \right) \mathrm{d}t + \sqrt{2 t} \mathrm{d} W_{t}\;.
    \label{eqapx:forward_sde}
\end{equation}

Starting from the Fokker-Plank equation \cite{fokker1914mittlere, planck1917ueber}, we have:
\begin{align*}
    \frac{\partial p_t }{\partial t}  &= -\nabla \cdot (p_t (v^{\star}_t + t \nabla \log p_t)) + t\Delta p_t \\
    &= -\nabla \cdot (p_t v^{\star}_t) -\nabla \cdot (t p_t\nabla \log p_t)) + t\Delta p_t \\
    &= -\nabla \cdot (p_t v^{\star}_t) -\nabla \cdot (t \nabla p_t)) + t\Delta p_t \\
    &= -\nabla \cdot (p_t v^{\star}_t) - t\Delta p_t + t\Delta p_t \\
    &= -\nabla \cdot (p_t v^{\star}_t) \;.
\end{align*}
We see that the forward SDE \eqref{eqapx:forward_sde} shares the same marginals as the ODE $\dot{x}_t=v^{\star}(x_t,t)$, which we use during the forward.

We utilize the following Lemma, established by \cite{rogers2000diffusions, didi2023framework}, to connect modified SDEs to tilted distributions.
\begin{lemma}[Adapted from \cite{didi2023framework}]
Given a backward SDE of the form $\mathrm{d}x_t = (v^{\star}(x_t, t) - t \nabla_x \log p_t(x_t)) \mathrm{d}t + \sqrt{2t} \ \mathrm{d}\bar{W_t}$ with $x_{1} \sim p_{1}$, the reverse-time SDE:
\begin{equation}
\mathrm{d}x_t = \left(v^{\star}(x_t, t) - t \nabla_x \log p_t(x_t) - 2 t \nabla_{x} \log Z_{t}(x_{t} ; x_{c}) \right) \mathrm{d}t + \sqrt{2t} \mathrm{d}\bar{W}_t\;,
\end{equation}
satisfies $\mathrm{Law}(x{_0}) = \tilde{p}_{0}(\cdot ; x_{c})$, where $\tilde{p}_{0}(x ; x_{c}) \propto p_{0}(x) e^{\lambda \mathscr{V}\left(\phi(x), \phi(x_{c})\right)}$.
\label{lemma:didi}
\end{lemma}
\begin{proposition}
Assume $\mathscr{V}(h, h^{\star}) = \langle h, h^{\star} \rangle$. Under Assumptions \ref{asm:local_opt} and \ref{asm:jensen_gap}, for a flow model $v_{\theta}$ trained with representation alignment, the SDE:
\begin{equation}
\mathrm{d}x_{t} = \left( v^{\star}(x_t, t) - t \nabla_x \log p_t(x_t) - 2\lambda t \nabla \mathscr{V}(x_t, \phi(x_{c})) \right)\mathrm{d}t +  \sqrt{2 t} \, \mathrm{d}\bar{W}_{t}\;, 
\label{eq:sde_guidance}
\end{equation}
with $x_1 \sim p_1$, produces samples distributed according to the tilted distribution $\tilde{p}_{0}(\cdot ; x_{c})$.
\label{prop:law}
\end{proposition}

\begin{manualproof}[Proof of lemma \ref{lemma:didi}]
Consider the forward SDE \eqref{eqapx:forward_sde}, where the marginals $p_t$ are induced by the interpolation process $x_t = \alpha_t x_0 +\sigma_t x_1$, with $x_0\sim p_0$ and $x_1\sim p_1$. We define the tilting term $p(\phi(x_{c})\mid x_0) = e^{\lambda \mathscr{V}\left(\phi(x_0), \phi(x_{c})\right)}$ and the corresponding time-dependent potential $Z_{t}(x ; x_{c}) \triangleq \mathbb{E}_{x_{0} \sim p(x_{0} \mid x_{t})} \left[ e^{ \lambda \mathscr{V}\left(\phi(x_{0}),\phi(x_c)\right)} \mid x_{t} = x \right]$. Applying Proposition 2.3. from~\cite{didi2023framework}, the reverse-time SDE is given by:
\begin{align*}
    \mathrm{d}x_{t} =& \left( v^{\star}(x_t, t) + t \nabla_x \log p_t(x_t) - 2t( \nabla_x \log p_t(x_t) +\nabla_x \log Z_{t}(x ; x_{c}))\right) \mathrm{d}t + \sqrt{2 t} \mathrm{d} \bar{W_{t}} \\
    =& \left( v^{\star}(x_t, t) - t (\nabla_x \log p_t(x_t) + 2\nabla_x \log Z_{t}(x ; x_{c}))\right) \mathrm{d}t + \sqrt{2 t} \mathrm{d} \bar{W_{t}}
\end{align*}
Initialised at $x_1\sim p_1$, this SDE yields samples $x_0 \sim \tilde{p}_{0}(x ; x_{c}) \propto p_{0}(x) e^{\lambda \mathscr{V}\left(\phi(x), \phi(x_{c})\right)}$, effectively sampling from the tilted distribution.
\end{manualproof}
\begin{manualproof}[Proof of proposition \ref{prop:law}]
By Assumption \ref{asm:local_opt}, the optimality condition \eqref{eq:repa_opt} holds, which implies that the network recovers the conditional expectation: $f_{\theta}(x, t) = \mathbb{E}_{(x_0,x_1) \sim p_0 \times p_1} [\phi(x_0) \mid x_t = x]$. Consequently:
\begin{equation}
\mathscr{V}(f_{\theta}(x, t), \phi(x_{c})) = \mathbb{E}_{(x_0,x_1) \sim p_0 \times p_1} [ \langle \phi(x_0), \phi(x_c) \rangle \mid x_t = x]\;.
\end{equation}
Under Assumption \ref{asm:jensen_gap}, the gradient of this expectation is equivalent to $\nabla_x \log Z_t(x ; x_c)$. Substituting this result into the formulation of the reverse SDE from Lemma \ref{lemma:didi} completes the proof.\end{manualproof}

\subsection{Toy example}
\label{apx:toy}

To visualize the capacity of \ours for conditional generation, we consider a simple 2D distribution $p_0(x) = \frac{1}{2}\mathbf{1}_{}[x\in C_1 \cup C_2 ]$, where $C_1 = \{x=(x_1,x_2);-1\le x_1 \le 0 \, , 0\le x_2 \le 1\}$ and $C_2 = \{x=(x_1,x_2);0\le x_1 \le 1 \, , -1\le x_2 \le 0\}$. To learn a velocity field mapping $p_0$ to the standard normal distribution $p_1 = \mathcal{N}(0,I)$, we train a 5-layer MLP with ReLU activation functions following the REPA procedure~\cite{yu2025representation}. We used Adam \cite{kingma2015adam} optimizer with learning rate of $10^{-3}$, a batch size of $512$, and $300$ epochs, with a dataset size of $100{,}000$ points. The model utilizes 512-channel intermediate layers, taking a 2D coordinate and time as input to output a 2D velocity vector. We extract representations from the third layer and map them to the alignment target space using a 2-layer MLP projection head.

To mimic the $\ell_2$ normalization used in REPA, we design a 2D target feature space constrained to the unit circle. Specifically, the target embedding $\phi(x)$ is defined as:
\begin{equation}
\phi(x) = [\frac{t}{\sqrt{t^2 + (h-w)^2}},\frac{h-w}{\sqrt{t^2 + (h-w)^2}}],
\end{equation}
where $t$, $h$, and $w$ represent the normalized distances from the cell center, the nearest vertical edge, and the nearest horizontal edge, respectively. The model is trained for 300 epochs using the composite loss $\mathcal{L}^{\mathrm{diff}} + \beta \mathcal{L}^{\mathrm{align}}$ with $\beta=0.5$.

\begin{figure*}[t]
    \centering
\begin{tabular}{@{}c@{} c@{} c@{} c@{} | c@{}}
    
    \begin{minipage}[c]{0.02\textwidth}
        \centering
        \rotatebox{90}{\textbf{GT}}
    \end{minipage} & 
    \begin{minipage}{0.23\textwidth}
        \centering
        Unconditional
        \includegraphics[width=\linewidth]{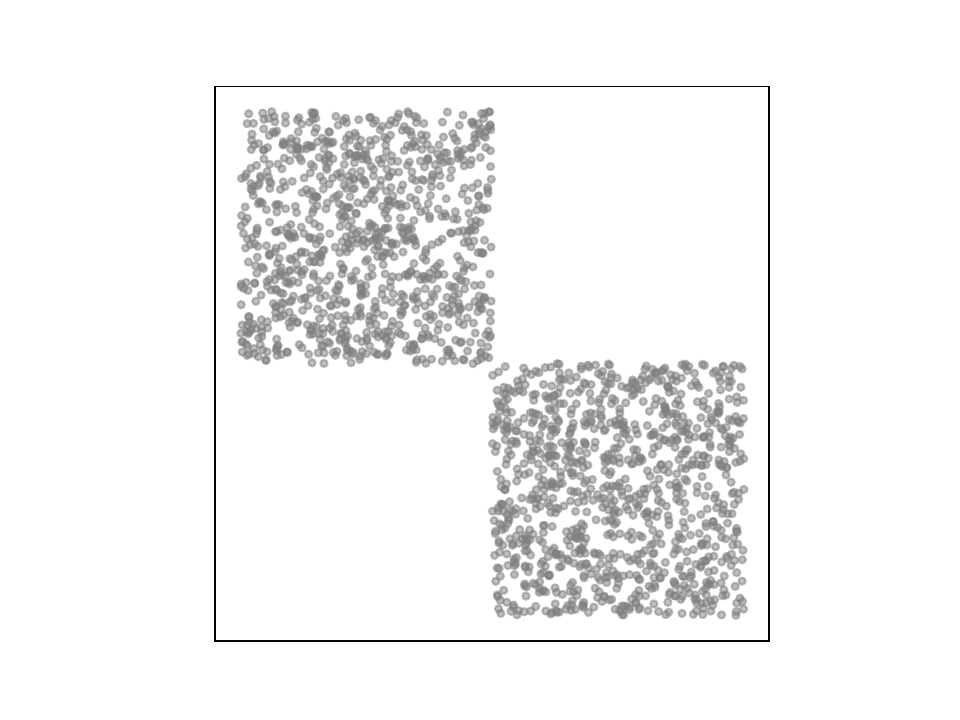}\\  
    \end{minipage} & 
    \begin{minipage}{0.23\textwidth}
        \centering
        Conditional
        \includegraphics[width=\linewidth]{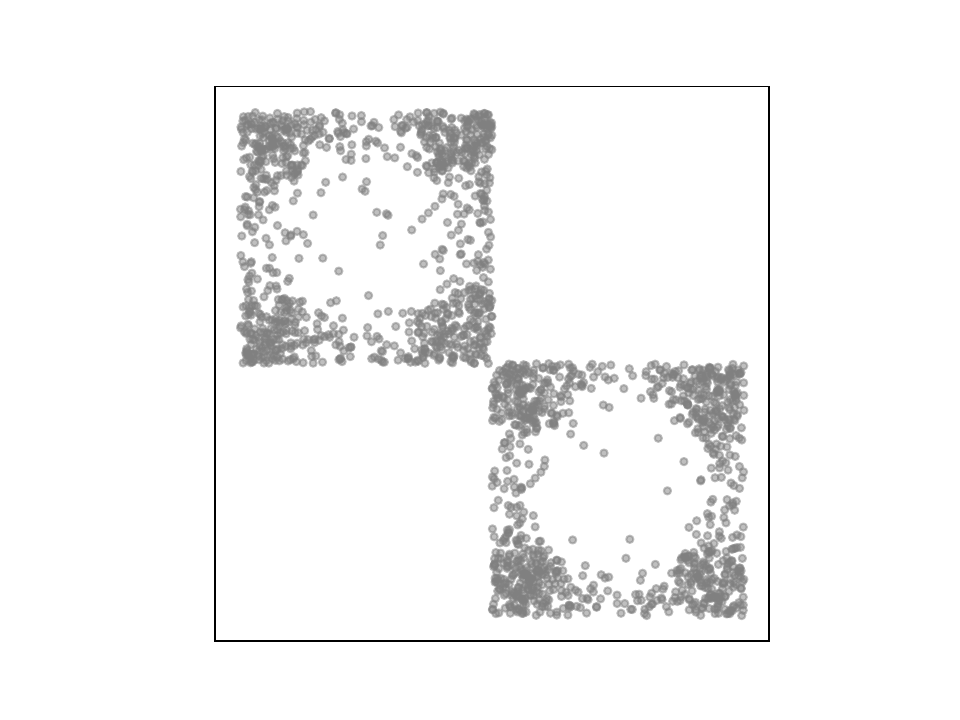}\\
    \end{minipage} & 
    \begin{minipage}{0.23\textwidth}
        \centering
        Conditional
        \includegraphics[width=\linewidth]{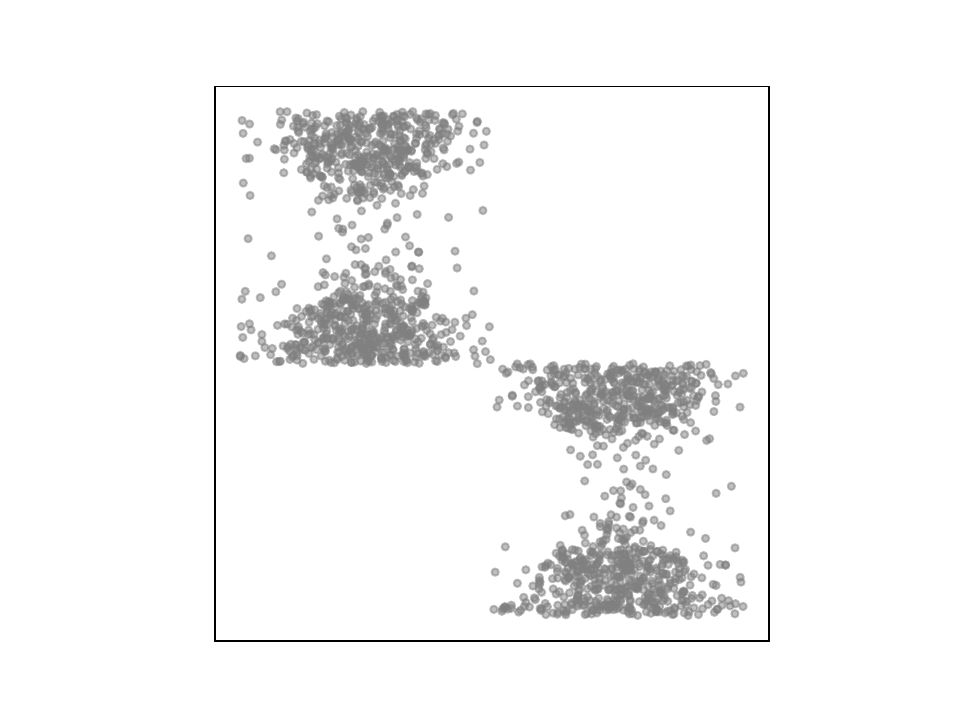}\\
    \end{minipage} & 
    \begin{minipage}{0.23\textwidth}
        \centering
        \includegraphics[width=\linewidth]{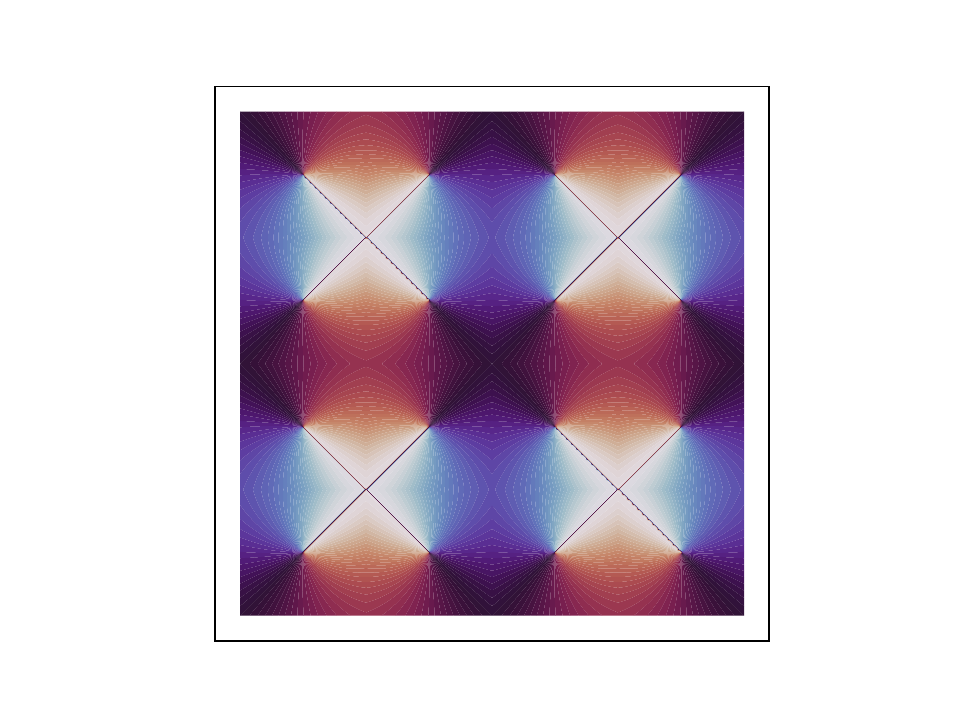}\\
        Feature Space Mapping
    \end{minipage} \\

    \begin{minipage}[c]{0.02\textwidth}
        \centering
        \rotatebox{90}{\textbf{Generated}}
    \end{minipage} & 
    \begin{minipage}{0.23\textwidth}
        \centering
        \includegraphics[width=\linewidth]{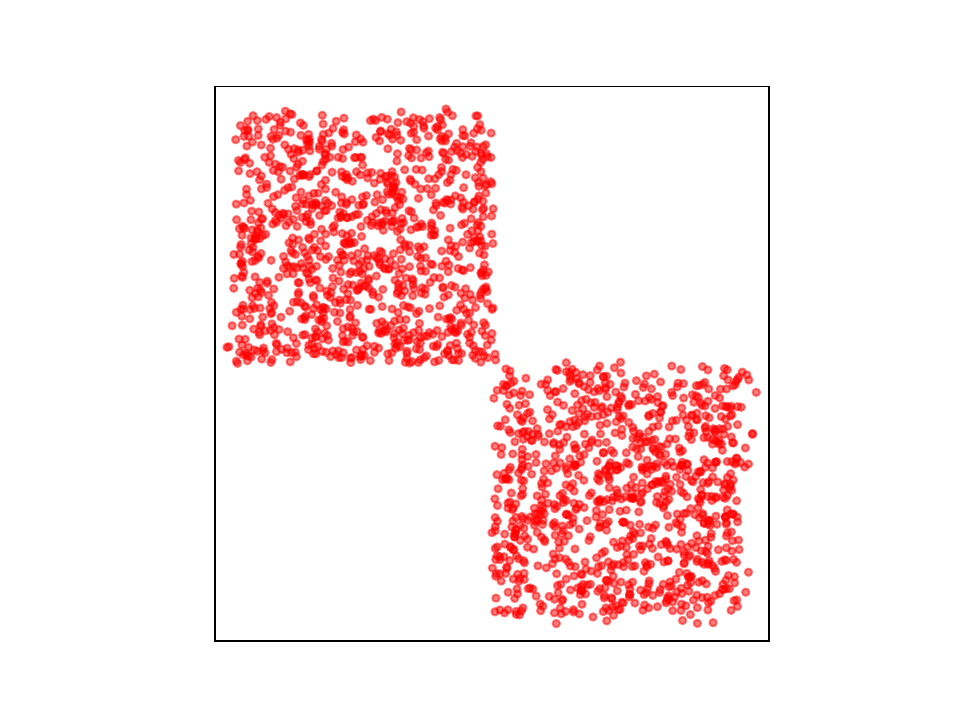}\\
    \end{minipage} & 
    \begin{minipage}{0.23\textwidth}
        \centering
        \includegraphics[width=\linewidth]{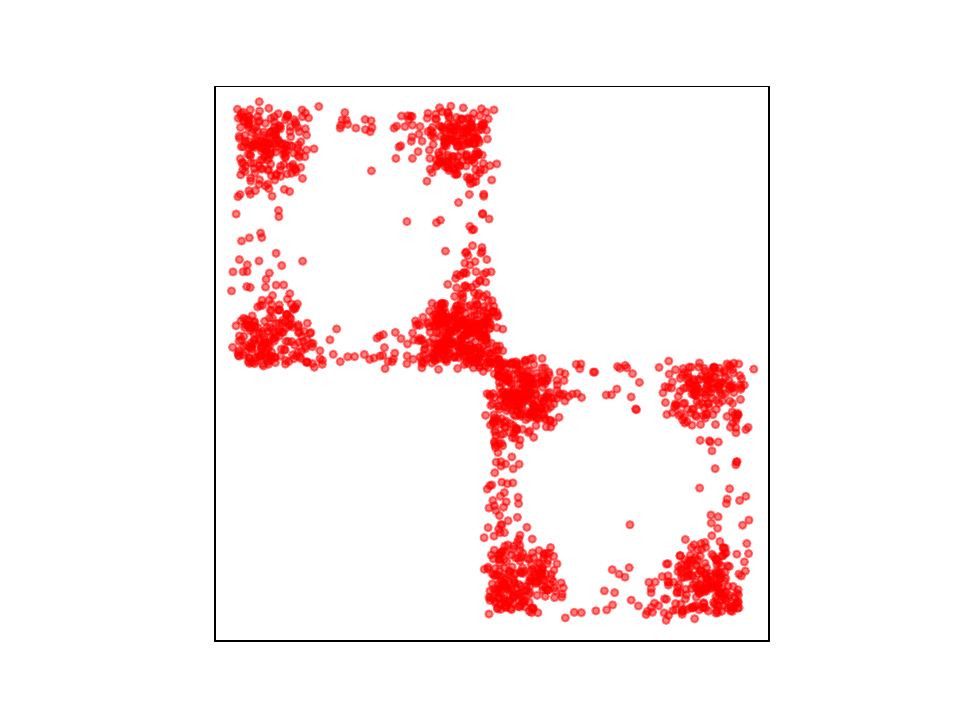}\\
    \end{minipage} & 
    \begin{minipage}{0.23\textwidth}
        \centering
        \includegraphics[width=\linewidth]{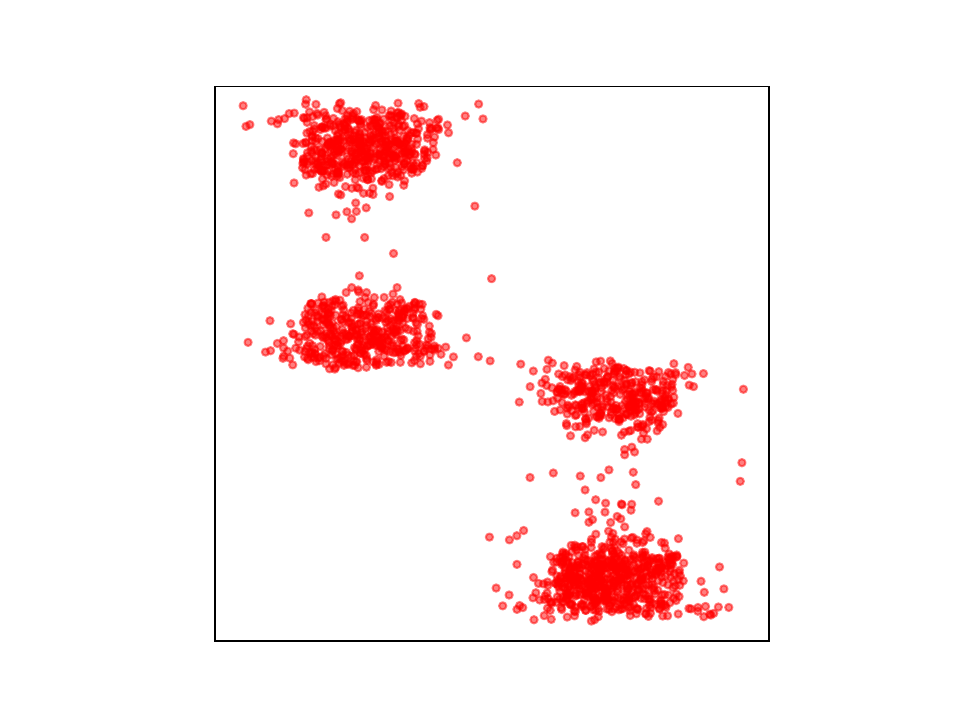}\\
    \end{minipage} & 
    \begin{minipage}{0.23\textwidth}
        \centering
        \includegraphics[width=\linewidth]{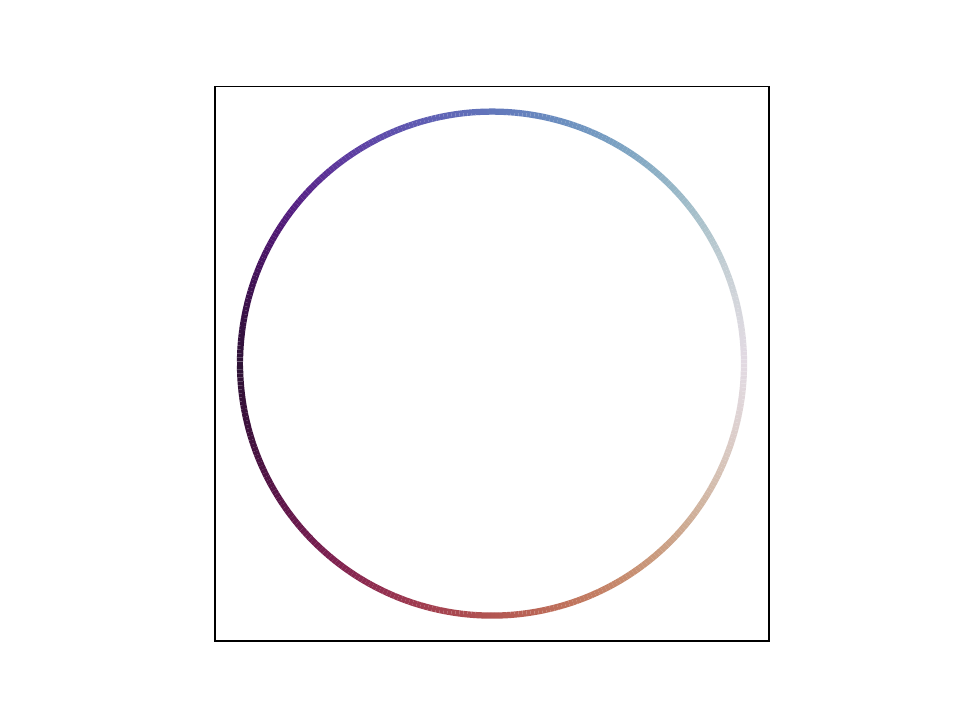}\\
        2D Feature Space
    \end{minipage} 
    \end{tabular}
    \caption{\textbf{Representation alignment with diffusion in 2D space.} Ground-truth samples are shown in grey, with corresponding samples generated via diffusion in red. The conditional results in the second and third columns are guided by target feature vectors $[-1,0]$ and $[0,-1]$, respectively. 
    The last column shows the feature space mapping, mapping each data points to its corresponding 2D feature.} 
    \label{fig:toy_app}
\end{figure*}

Figure~\ref{fig:toy_app} illustrates $2{,}000$ generated samples (red) alongside the ground-truth distribution (grey). For conditional density sampling, we employ a rejection algorithm~\cite{DEVROYE200683} to generate reference points from the tilted distribution. Diffusion sampling is performed via an Euler-Maruyama solver with 250 steps and an optional guidance scale of $\lambda=2$. The conditional results in the third and fourth columns are induced by target feature vectors $[-1,0]$ and $[0,-1]$, respectively.

\subsection{Algorithm}
\label{app:algorithm}

Algorithm~\ref{alg:conditioning} describes the test-time conditioning process using a potential function $\mathscr{V}$ that is parametrized by the conditioning features and the guidance scale $\lambda$. Optionally, it can be a combination of joint potential, i.e, $\mathscr{V}=\lambda^{1}\mathscr{V}^{1}+\lambda^{2}\mathscr{V}^{2}$.

\begin{algorithm}[h]
\caption{\ours}
\begin{algorithmic}[1]
\STATE \textbf{Input:} Initial noise $x_1 \sim \mathcal{N}(0, I)$, conditioning features $h^{\star}$, guidance scale $\lambda$, potential $\mathscr{V}$, and step size $\Delta t$.
\FOR{$t = 1$ \textbf{down to} $0$}
    \STATE Compute velocity $v_\theta(x_t, t)$ and internal representation $h_t = (h_\theta \circ f_\theta)(x_t, t)$. 
    \STATE Compute the score $\nabla_{x_t} \log p_t(x_t)$ from the velocity~\eqref{eq:score}.
    \STATE Compute guidance gradient $\nabla_{x_t} \mathscr{V}(h_t, h^{\star})$. 
    \STATE Sample $\epsilon \sim \mathcal{N}(0, I)$.
    \STATE Update $x_t$ using the Euler-Maruyama discretization of the reverse-time SDE~\eqref{eq:sde_guidance}:
    \STATE \quad $x_{t - \Delta t} = x_t - \left( v_\theta(x_t, t) - t \nabla_{x_t} \log p_t(x_t) - 2\lambda t \nabla_{x_t} \mathscr{V}(h_t, h^{\star}) \right)\Delta t + \sqrt{2t \Delta t} \, \epsilon$.
\ENDFOR
\STATE \textbf{Return} $x_0$.
\end{algorithmic}
\label{alg:conditioning}
\end{algorithm}

\section{Complementary details on \textit{Properties of representation space} (Section \ref{sec:properties})}
\subsection{Clustering of ImageNet}
\label{app:clusteringimagenet}
We perform k-means clustering across ImageNet~\cite{deng2009imagenet} using $k=1000$ clusters across four distinct feature spaces via the FAISS library~\cite{douze2024faiss}. We use the normalized global average of each image's spatial feature map. Features are extracted from: (i) Dinov2~\cite{oquab2024dinov2}, (ii) SiT-XL/2 with representation alignment~\cite{yu2025representation} at the 8th layer (both before and after projection), and (iii) the baseline SiT-XL/2 trained without alignment~\cite{ma_sit_2024} at the same layer.
Figure~\ref{fig:supp_clusters} provide two additional cluster comparisons for each feature space.

\subsection{Euclidean embedding property}\label{apx:eep}

Following \cite{mallat_ttc}, we investigate the Euclidean embedding property. As established in Section~\ref{sec:preliminary},  there must exist $0 < A \le B$ and where the ratio $B/A$ is reasonably small such that for all $x_1$, $x_2$:\begin{equation*}A \lVert \phi(x_1) - \phi(x_2) \rVert^2\le d^2(p_1, p_2)\le B \lVert \phi(x_1) - \phi(x_2) \rVert^2.\end{equation*}We now justify the choice of the metric $d^2$. Let $\phi_1$ and $\phi_2$ be two embeddings that induce two tilted conditional densities $p_1$ and $p_2$~\eqref{eq:tilt_p}, respectively.

\newcommand{\blockdinoflamingo}{
{%
\setlength{\tabcolsep}{0pt}%
\renewcommand{\arraystretch}{0}%
\begin{tabular}{@{}cccc@{}}
{\setlength{\fboxsep}{0.0pt}\fcolorbox{red}{white}{\img{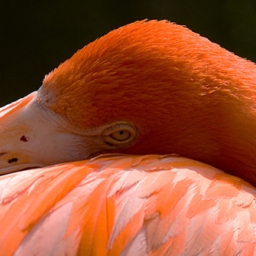}}} &
\img{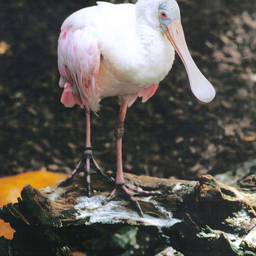} &
\img{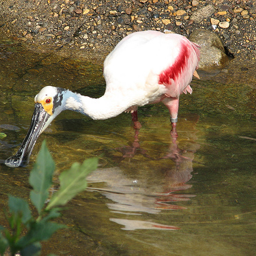} &
\img{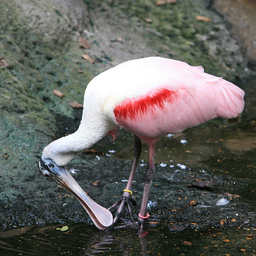} \\

\img{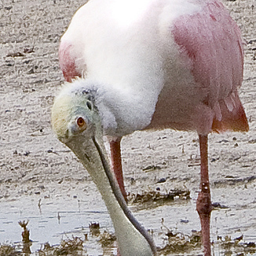} &
\img{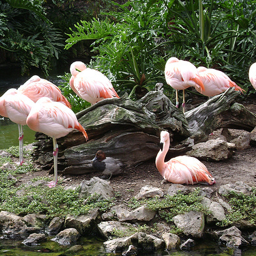} &
\img{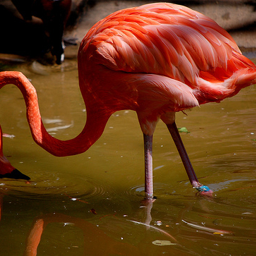} &
\img{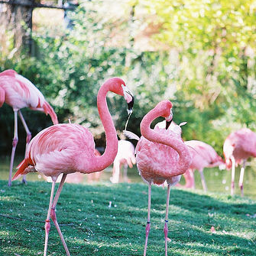} \\

\img{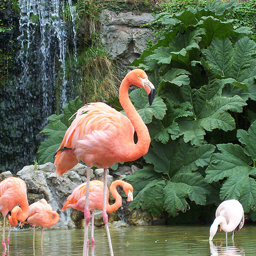} &
\img{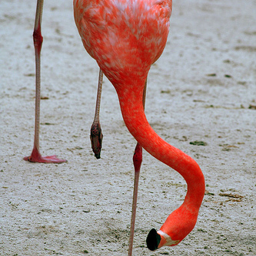} &
\img{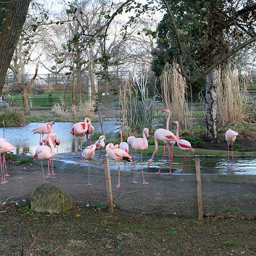} &
\img{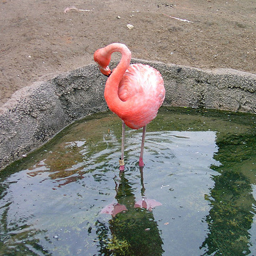} \\

\img{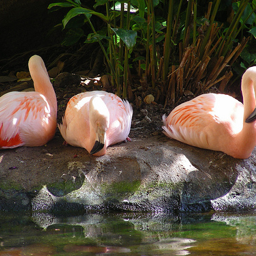} &
\img{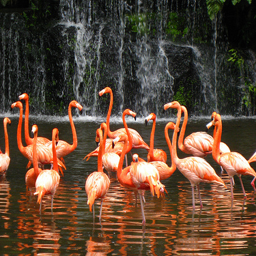} &
\img{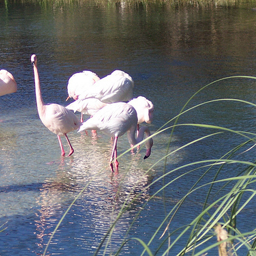} &
\img{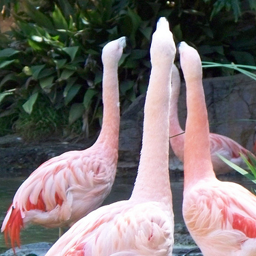}
\end{tabular}
}%
}

\newcommand{\blocksitflamingo}{
{%
\setlength{\tabcolsep}{0pt}%
\renewcommand{\arraystretch}{0}%
\begin{tabular}{@{}cccc@{}}
{\setlength{\fboxsep}{0.0pt}\fcolorbox{red}{white}{\img{fig/clusters/ref169195_20260122_180234/ref.jpg}}} &
\img{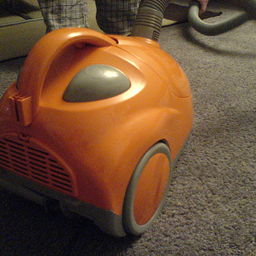} &
\img{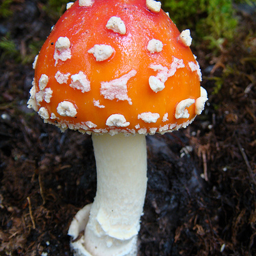} &
\img{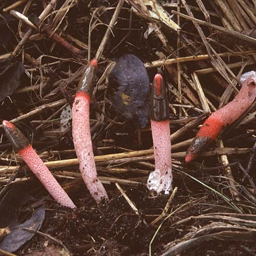} \\

\img{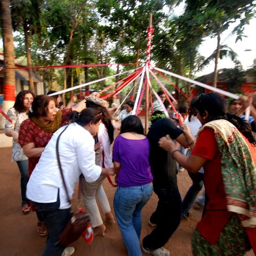} &
\img{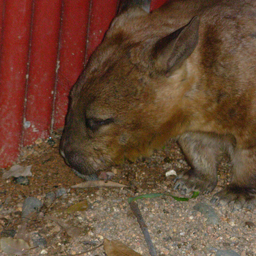} &
\img{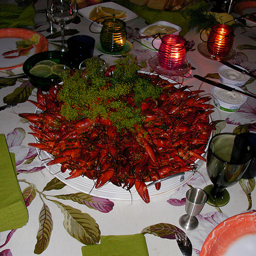} &
\img{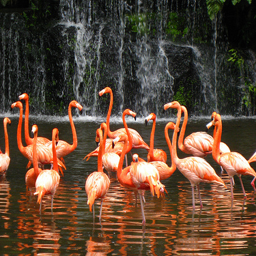} \\

\img{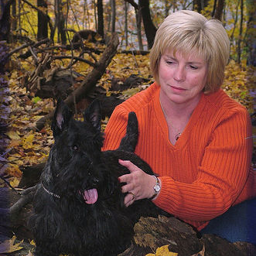} &
\img{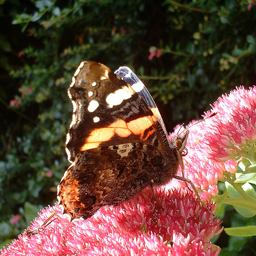} &
\img{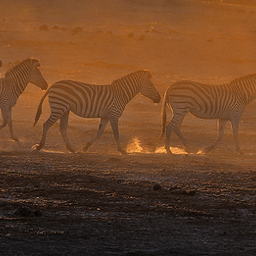} &
\img{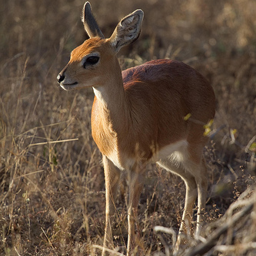} \\

\img{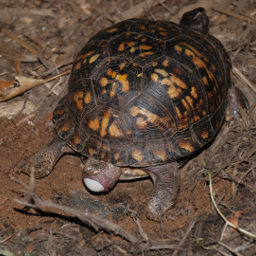} &
\img{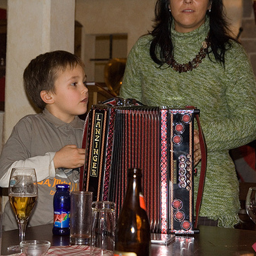} &
\img{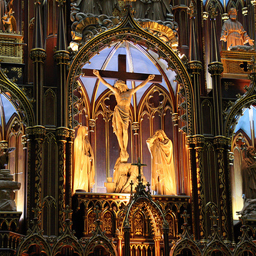} &
\img{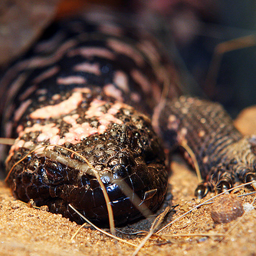}
\end{tabular}
}%
}

\newcommand{\blocksitrepaflamingo}{
{%
\setlength{\tabcolsep}{0pt}%
\renewcommand{\arraystretch}{0}%
\begin{tabular}{@{}cccc@{}}
{\setlength{\fboxsep}{0.0pt}\fcolorbox{red}{white}{\img{fig/clusters/ref169195_20260122_180234/ref.jpg}}}
 &
\img{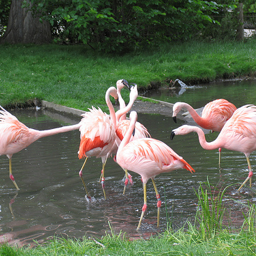} &
\img{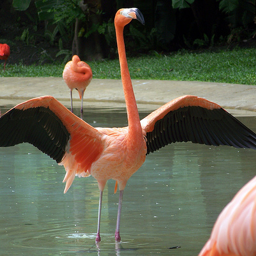} &
\img{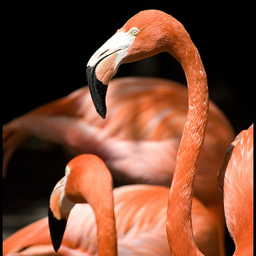} \\

\img{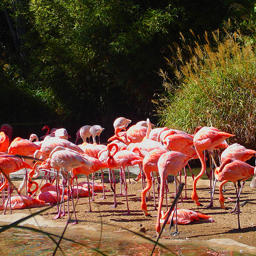} &
\img{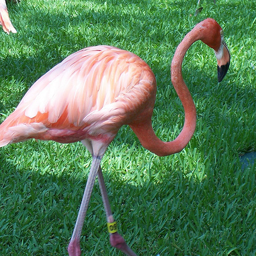} &
\img{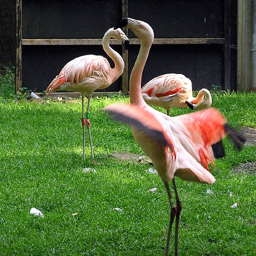} &
\img{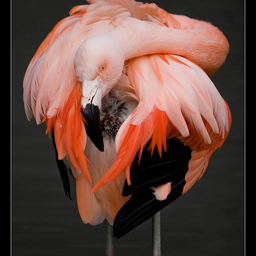} \\

\img{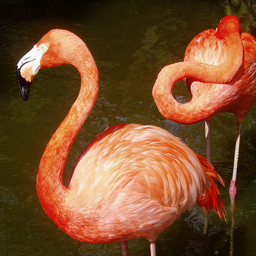} &
\img{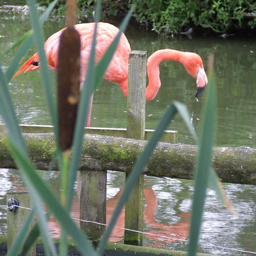} &
\img{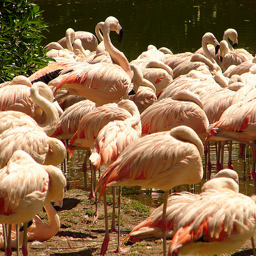} &
\img{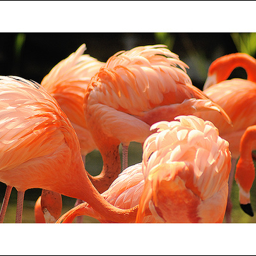} \\

\img{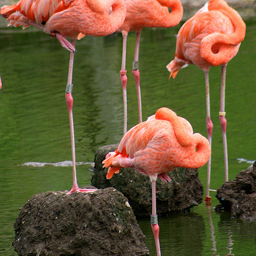} &
\img{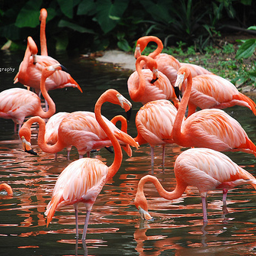} &
\img{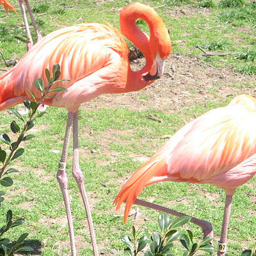} &
\img{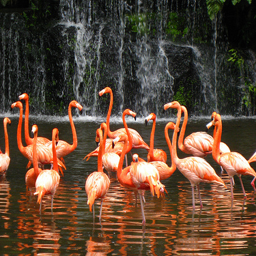}
\end{tabular}
}%
}

\newcommand{\blocksitrepaprojflamingo}{
{%
\setlength{\tabcolsep}{0pt}%
\renewcommand{\arraystretch}{0}%
\begin{tabular}{@{}cccc@{}}
{\setlength{\fboxsep}{0.0pt}\fcolorbox{red}{white}{\img{fig/clusters/ref169195_20260122_180234/ref.jpg}}}
 &
\img{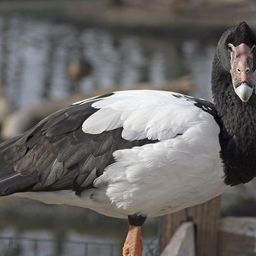} &
\img{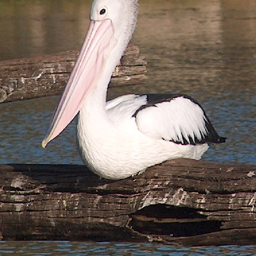} &
\img{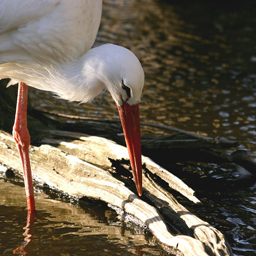} \\

\img{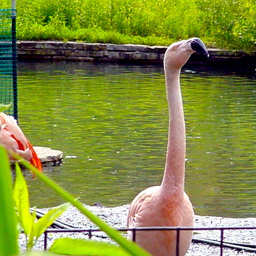} &
\img{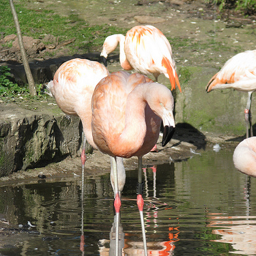} &
\img{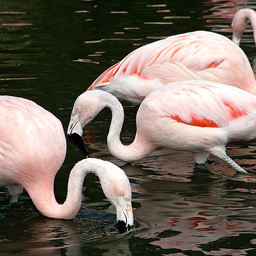} &
\img{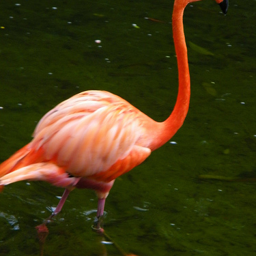} \\

\img{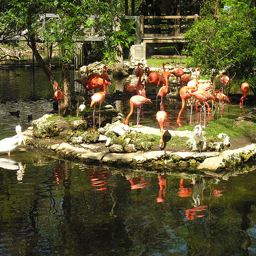} &
\img{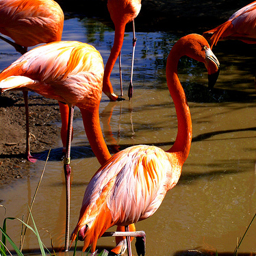} &
\img{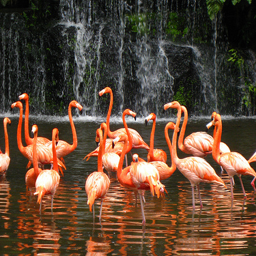} &
\img{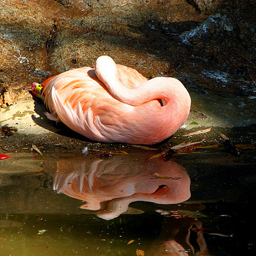} \\

\img{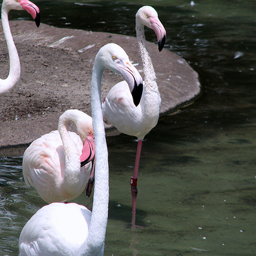} &
\img{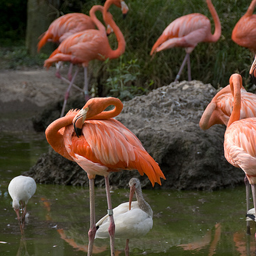} &
\img{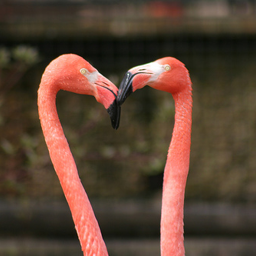} &
\img{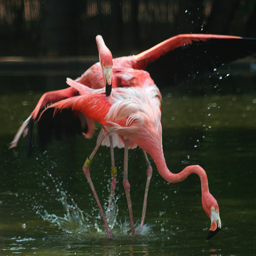}
\end{tabular}
}%
}

\newcommand{\blockdinoguitar}{
{%
\setlength{\tabcolsep}{0pt}%
\renewcommand{\arraystretch}{0}%
\begin{tabular}{@{}cccc@{}}
{\setlength{\fboxsep}{0.0pt}\fcolorbox{red}{white}{\img{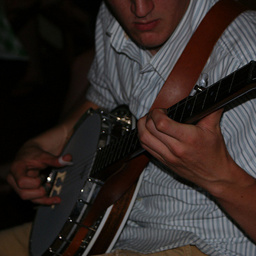}}}
 &
\img{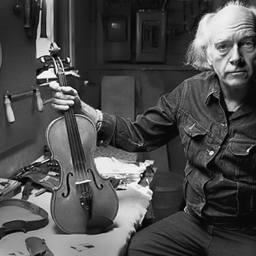} &
\img{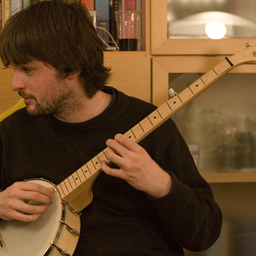} &
\img{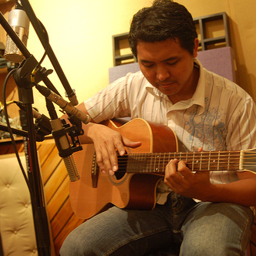} \\

\img{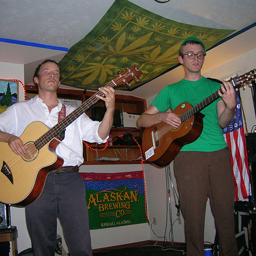} &
\img{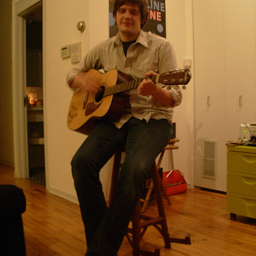} &
\img{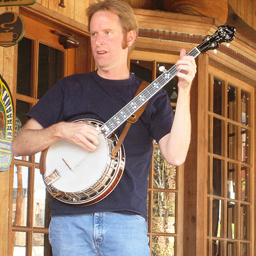} &
\img{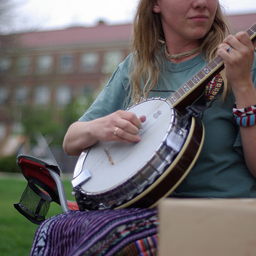} \\

\img{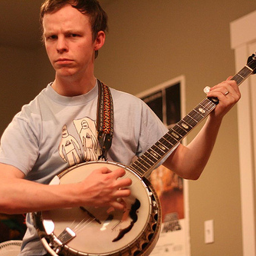} &
\img{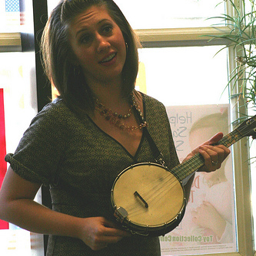} &
\img{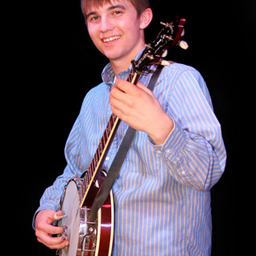} &
\img{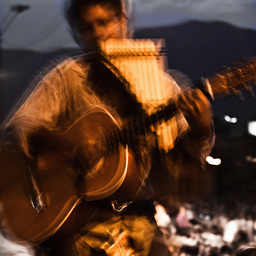} \\

\img{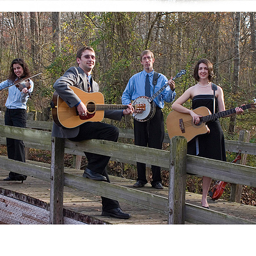} &
\img{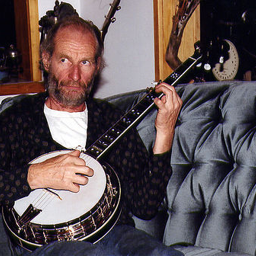} &
\img{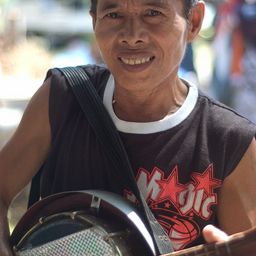} &
\img{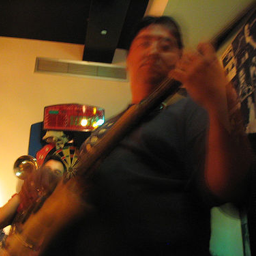} \\

\end{tabular}
}%
}

\newcommand{\blocksitguitar}{
{%
\setlength{\tabcolsep}{0pt}%
\renewcommand{\arraystretch}{0}%
\begin{tabular}{@{}cccc@{}}
{\setlength{\fboxsep}{0.0pt}\fcolorbox{red}{white}{\img{fig/clusters/ref539931_20260122_173703/dino/img_539931.png}}} &
\img{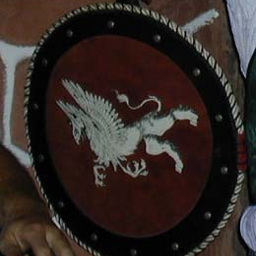} &
\img{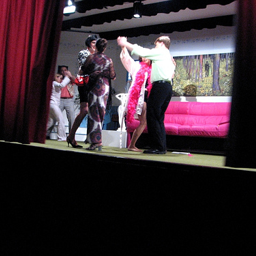} &
\img{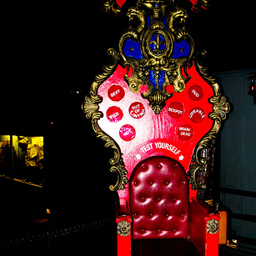} \\

\img{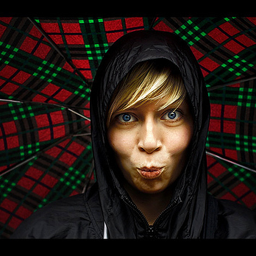} &
\img{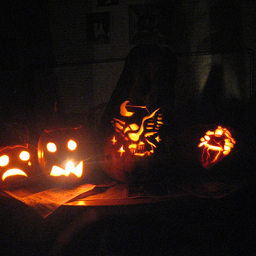} &
\img{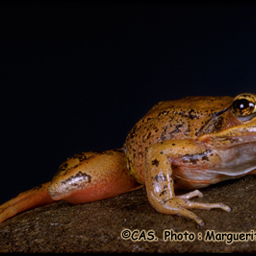} &
\img{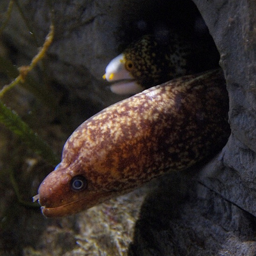} \\

\img{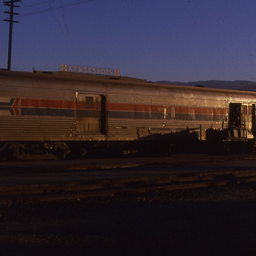} &
\img{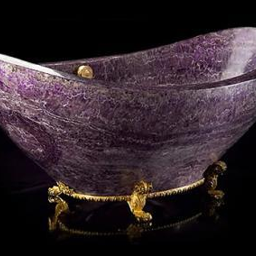} &
\img{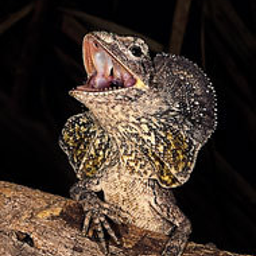} &
\img{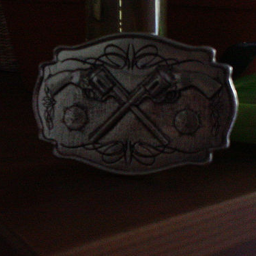} \\

\img{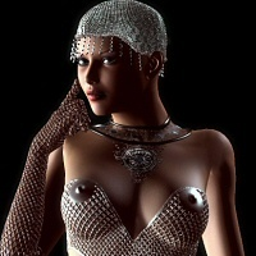} &
\img{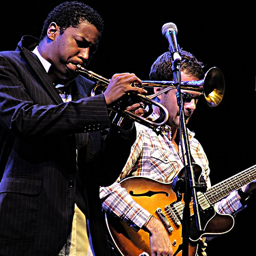} &
\img{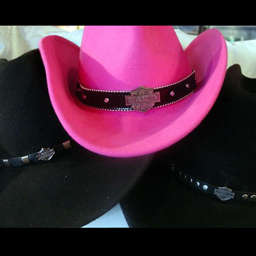} &
\img{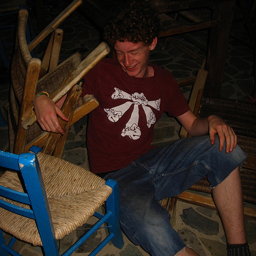} \\

\end{tabular}
}%
}

\newcommand{\blocksitrepaguitar}{
{%
\setlength{\tabcolsep}{0pt}%
\renewcommand{\arraystretch}{0}%
\begin{tabular}{@{}cccc@{}}
{\setlength{\fboxsep}{0.0pt}\fcolorbox{red}{white}{\img{fig/clusters/ref539931_20260122_173703/dino/img_539931.png}}} &
\img{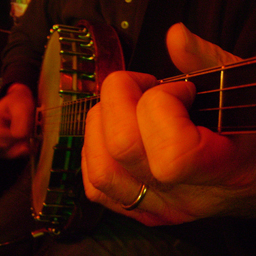} &
\img{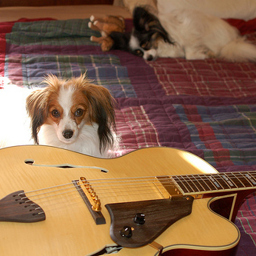} &
\img{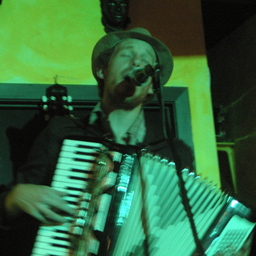} \\

\img{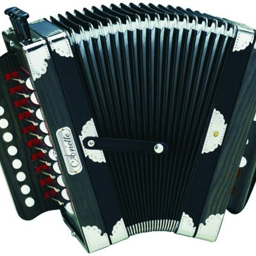} &
\img{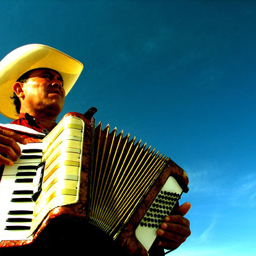} &
\img{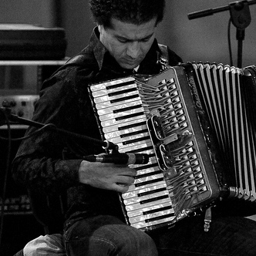} &
\img{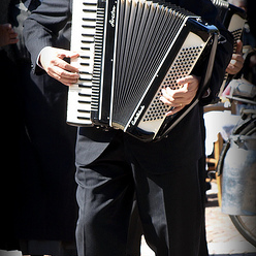} \\

\img{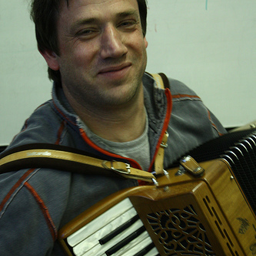} &
\img{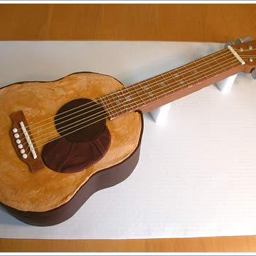} &
\img{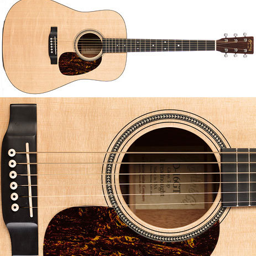} &
\img{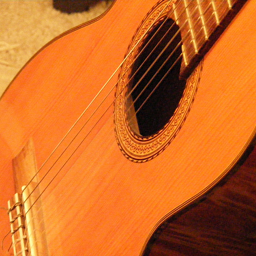} \\

\img{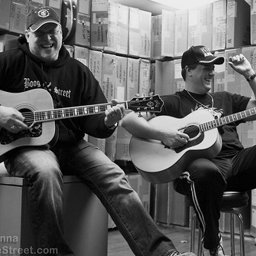} &
\img{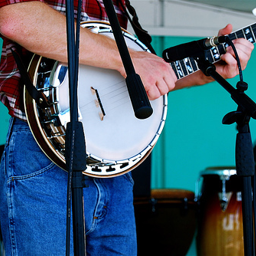} &
\img{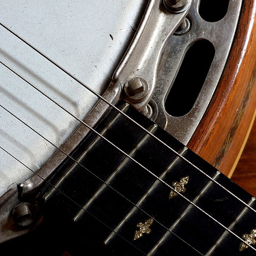} &
\img{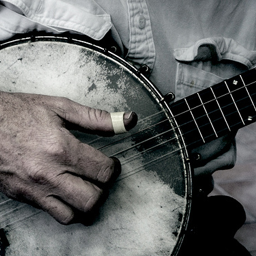} \\

\end{tabular}
}%
}

\newcommand{\blocksitrepaprojguitar}{
{%
\setlength{\tabcolsep}{0pt}%
\renewcommand{\arraystretch}{0}%
\begin{tabular}{@{}cccc@{}}
{\setlength{\fboxsep}{0.0pt}\fcolorbox{red}{white}{\img{fig/clusters/ref539931_20260122_173703/dino/img_539931.png}}} &
\img{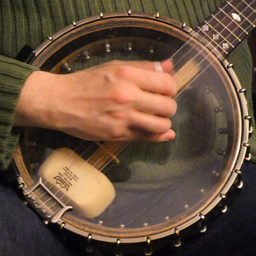} &
\img{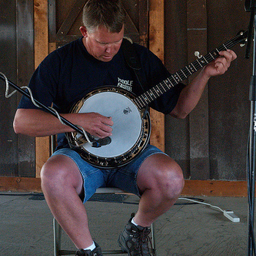} &
\img{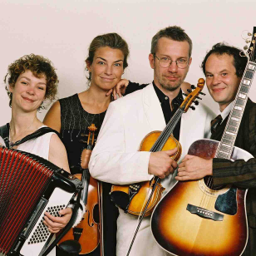} \\

\img{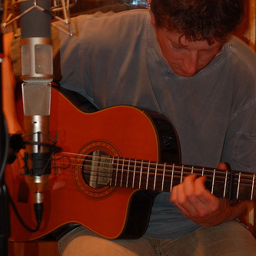} &
\img{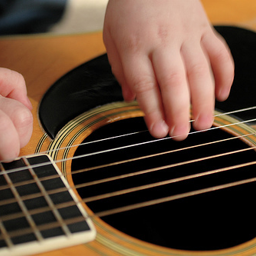} &
\img{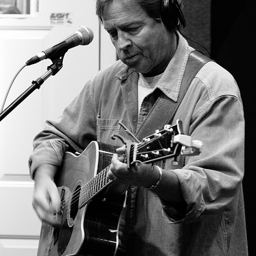} &
\img{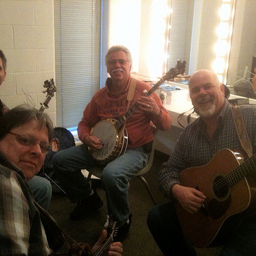} \\

\img{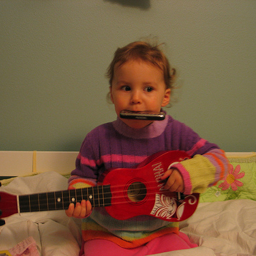} &
\img{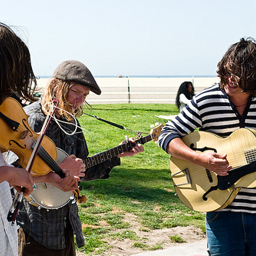} &
\img{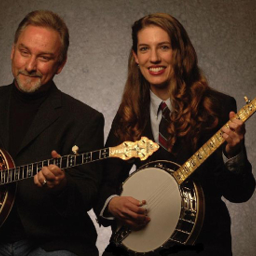} &
\img{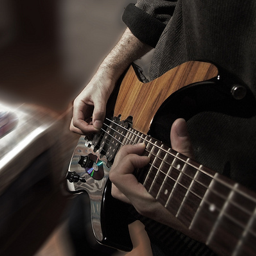} \\

\img{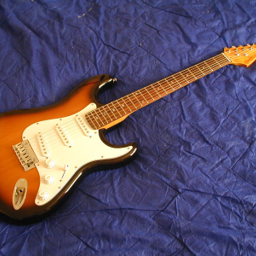} &
\img{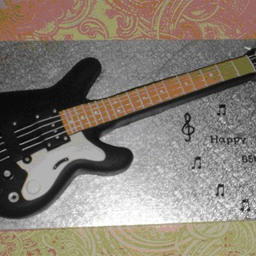} &
\img{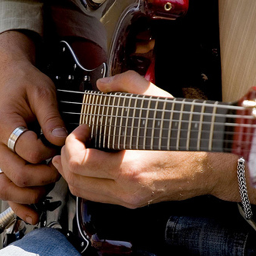} &
\img{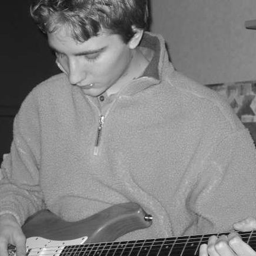} \\

\end{tabular}
}%
}

\begin{figure*}[!t]
\centering
\setlength{\tabcolsep}{1em}
\begin{tabular}{@{}c@{~}c@{~~~~}c@{~}c@{}}
\blockdinoflamingo        & \blocksitflamingo        &
\blockdinoguitar       & \blocksitguitar       \\[0.6em]

{\small DINOv2}             & {\small SiT}             &
{\small DINOv2}             & {\small SiT}             \\[0.4em]

\blocksitrepaflamingo     & \blocksitrepaprojflamingo &
\blocksitrepaguitar    & \blocksitrepaprojguitar \\[0.6em]

{\small SiT (Align)}         & {\small SiT (Align + Proj)}   &
{\small SiT (Align)}         & {\small SiT (Align + Proj)}
\end{tabular}
\caption{\textbf{Impact of representation alignment on SiT's feature space.} 
We compare clusters across ImageNet \cite{deng2009imagenet} using four feature spaces. For each space, we extract features for the entire ImageNet dataset and perform $k$-means clustering into 1,000 discrete clusters. Then, we identify the specific cluster associated with a given reference image (indicated by the red frame) and randomly sample other images from that same cluster.
Without alignment, the standard SiT model fails to form semantic clusters; features that are close in latent space represent unrelated concepts, which makes these internal representations unsuitable for conditioning. The aligned model successfully mimics the feature space of the visual teacher backbone both before and after the projection layer. This results in semantically consistent clusters. SiT (Align) refers to REPA \cite{singh2025irepa} features extracted before the projection layer, whereas SiT (Align + Proj) denotes features extracted after the projection layer.}
\label{fig:supp_clusters}
\end{figure*}

In this experiment, we utilize the full feature map to adhere strictly to our theoretical framework. For this setup, we define $d^2$ as the symmetric KL divergence:

\begin{align}
d^2(p_1,p_2)
:=& \mathrm{KL}(p_1 \| p_2) + \mathrm{KL}(p_2 \| p_1) \\
=& \mathbb{E}_{p_1}\!\left[\log \frac{p_1}{p_2}\right]
 + \mathbb{E}_{p_2}\!\left[\log \frac{p_2}{p_1}\right] \\
=& \mathbb{E}_{p_1}\!\left[\lambda \big(\mathscr{V}(\phi(x), \phi_1) - \mathscr{V}(\phi(x), \phi_2) \big)\right]
 + \mathbb{E}_{p_2}\!\left[\lambda \big(\mathscr{V}(\phi(x), \phi_2) - \mathscr{V}(\phi(x), \phi_1 )\big)\right] \\
=& \frac{\lambda}{N} \sum_{n=1}^N
\Big(
\mathbb{E}_{p_1}\!\left[\langle [\phi(x)]_n , [\phi_1 - \phi_2]_n \rangle\right] 
+ \mathbb{E}_{p_2}\!\left[\langle [\phi(x)]_n , [\phi_2 - \phi_1]_n \rangle\right]
\Big) \\
=& \frac{\lambda}{N} \sum_{n=1}^N
\left\langle
\big[ \mathbb{E}_{p_1}[\phi(x)] - \mathbb{E}_{p_2}[\phi(x)] \big]_n
,[\phi_1 - \phi_2]_n
\right\rangle .
\end{align}

\section{Complementary details on design of \textit{Potential} $\mathscr{V}$ (Section \ref{sec:method})}
\label{app:more_potential}
\subsection{Selective Patch Alignment (SPA)}
\label{apx:spa}

We define the Selective Patch Alignment (SPA) potential as:
\begin{equation}
\mathscr{V}^{\mathrm{SPA}}(h, h^{\star}) = T\log \left[ \sum_{n=1}^{N}\exp \left( \frac{\langle [h]_{n}, [h^{\star}]_{i}\rangle}{T}\right) \right]\;.
\label{eq:free_energy_app}
\end{equation}
As discussed in Section~\ref{sec:spa}, the temperature parameter $T$ modulates the sparsity of this selection. In the limit $T \rightarrow 0$, the potential is dominated by the patch most similar to $h^\star_i$, effectively performing a "hard" selection. Conversely, as $T \rightarrow \infty$, the patches are weighted equally, recovering the global averaging behavior of IPA. Consequently, SPA can be viewed as a generalization of IPA, where $T$ serves as a tunable parameter to optimize spatial correspondence. This relationship is further elucidated by the gradient:
\begin{equation}
\begin{aligned}
\nabla_{h_k} \mathscr{V}^{\mathrm{SPA}}(h, h^\star)
&= N \,
\sigma\!\left[
\frac{1}{T}
\bigl(
\langle h_n, h_i^\star \rangle
\bigr)_{n \in [1,N]}
\right]_k 
\nabla_{h_k} \mathscr{V}^{\mathrm{IPA}}(h, h^\star),
\end{aligned}
\end{equation}
where $\sigma(\cdot)$ denotes the softmax function. Here, the gradient of SPA is simply a reweighted version of the IPA gradient, where the weights are determined by the relative similarity of each patch to the target feature.

\section{More experimental results (Section \ref{sec:experiments})}
\label{app:more_exps_analysis}

\subsection{Hyperparameter selection}\label{app:hyperparameter_sweep}

\begin{table}[tbh]
    \centering
    \caption{\textbf{Guidance scale ($\lambda$) robustness.} FID and DINOv2 Alignment scores for single-source conditioning under various values of the guidance scale $\lambda$.}
    \resizebox{\linewidth}{!}{
    \begin{tabular}{lcccccccccccc}
        \toprule
        \multirow{2}{*}{\textbf{$\lambda$}} & \multicolumn{4}{c}{\textbf{Full}} & \multicolumn{4}{c}{\textbf{Masked}} & \multicolumn{4}{c}{\textbf{Average}} \\
        \cmidrule(lr){2-5} \cmidrule(lr){6-9} \cmidrule(lr){10-13}
        & \textbf{FID ($\downarrow$)} & \textbf{Prec. ($\uparrow$)} & \textbf{Rec. ($\uparrow$)} & \textbf{Align. ($\uparrow$)} & \textbf{FID ($\downarrow$)} & \textbf{Prec. ($\uparrow$)} & \textbf{Rec. ($\uparrow$)} & \textbf{Align. ($\uparrow$)} & \textbf{FID ($\downarrow$)} & \textbf{Prec. ($\uparrow$)} & \textbf{Rec. ($\uparrow$)} & \textbf{Align. ($\uparrow$)} \\
        \midrule
        5000 & 3.3 & 0.72 & 0.67 & 0.6 & 3.7 & 0.72 & 0.66 & 0.7 & 3.2 & 0.76 & 0.64 & 0.9 \\
        10000 & 2.8 & 0.75 & 0.67 & 0.7 & 2.8 & 0.74 & 0.66 & 0.8 & 2.8 & 0.76 & 0.65 & 0.9 \\
        20000 & 1.6 & 0.76 & 0.69 & 0.8 & 2.4 & 0.75 & 0.66 & 0.8 & 2.7 & 0.75 & 0.67 & 0.9 \\
        50000 & 1.4 & 0.76 & 0.69 & 0.8 & 2.3 & 0.75 & 0.67 & 0.9 & 3.3 & 0.73 & 0.68 & 0.9 \\
        100000 & 1.5 & 0.75 & 0.70 & 0.9 & 2.6 & 0.74 & 0.68 & 0.9 & 4.5 & 0.71 & 0.69 & 0.9 \\
        \bottomrule
    \end{tabular}
    }
    \label{tab:lambda_robustness}
\end{table} 

\paragraph{Robustness in single-source guidance}
We present a quantitative evaluation of the impact of the guidance scale ($\lambda$) on generation quality and feature alignment for single-source conditioning on ImageNet in Table~\ref{tab:lambda_robustness}. Numbers are computed using the same $50{,}000$ image anchors evenly distributed across the class of ImageNet~\citep{deng2009imagenet}. Here we report FID, Precision and Recall alongside alignment scores across DINOv2 feature spaces. For full feature map conditioning, we observe that increasing $\lambda$ consistently improves both FID and alignment scores before plateauing. We keep $\lambda$ fixed at $50{,}000$ for all other single source conditioning settings for simplicity. It is also the best performing value for the masked conditioning setting, while a lower value of $\lambda=20{,}000$ seems to be slightly more effective for the average conditioning setting. Importantly, the performance of single source conditioning remains robust across a wide range of $\lambda$ values, showing that this parameter does not introduce an overhead of careful tuning.

\paragraph{Hyperparameters for multi-source guidance.} Hyperparameters $\lambda$ and temperature $T$ are set to $10{,}000$ and 0.1, respectively, based on a grid search over a diverse 5-class ImageNet subset (i.e., \textit{goldfish, shark, golden retriever, bald eagle, tiger}). For each class, 10 random samples are combined with 8 hand-selected target background features.

\begin{figure}[h]
    \centering
    \begin{minipage}{0.48\textwidth}
        \centering
        \includegraphics[width=\linewidth]{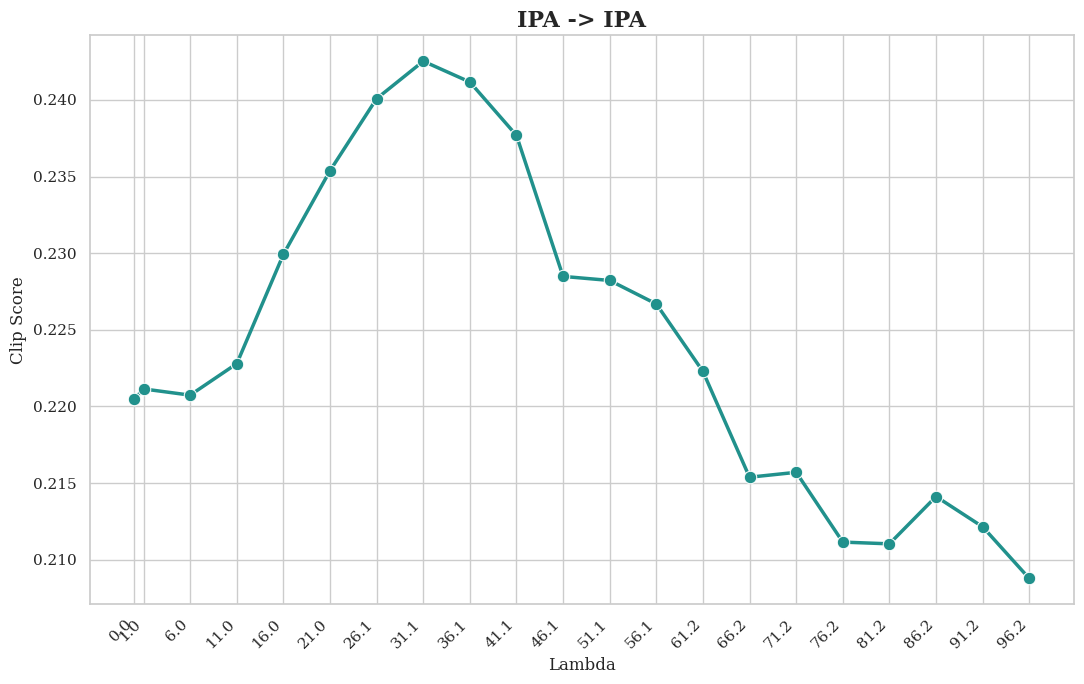}\\
        (a) CLIP Score (IPA)
    \end{minipage}\hfill
    \begin{minipage}{0.48\textwidth}
        \centering
        \includegraphics[width=\linewidth]{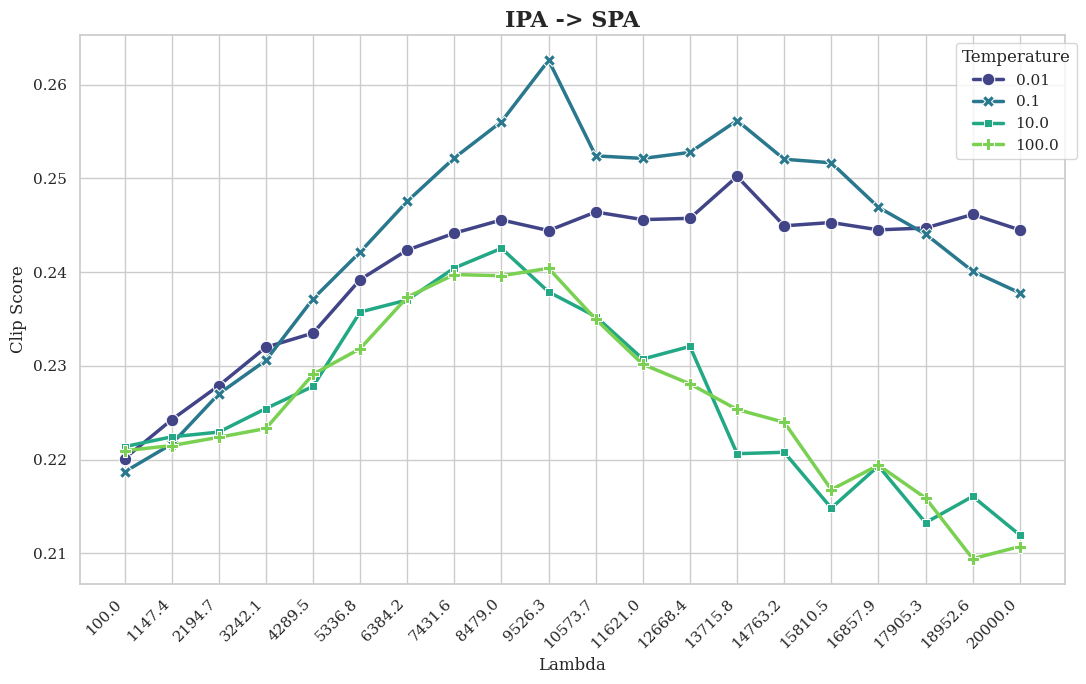}\\
        (b) CLIP Score (SPA)
    \end{minipage}
    
    \vspace{1em}
    \begin{minipage}{0.48\textwidth}
        \centering
        \includegraphics[width=\linewidth]{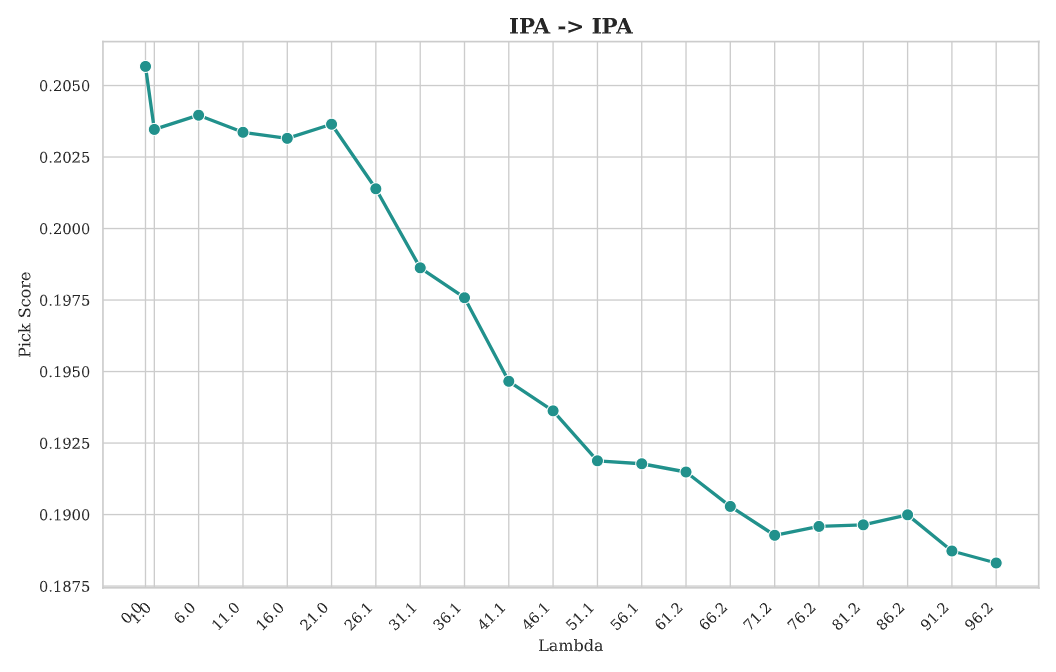}\\
        (c) Peak (IPA)
    \end{minipage}\hfill
    \begin{minipage}{0.48\textwidth}
        \centering
        \includegraphics[width=\linewidth]{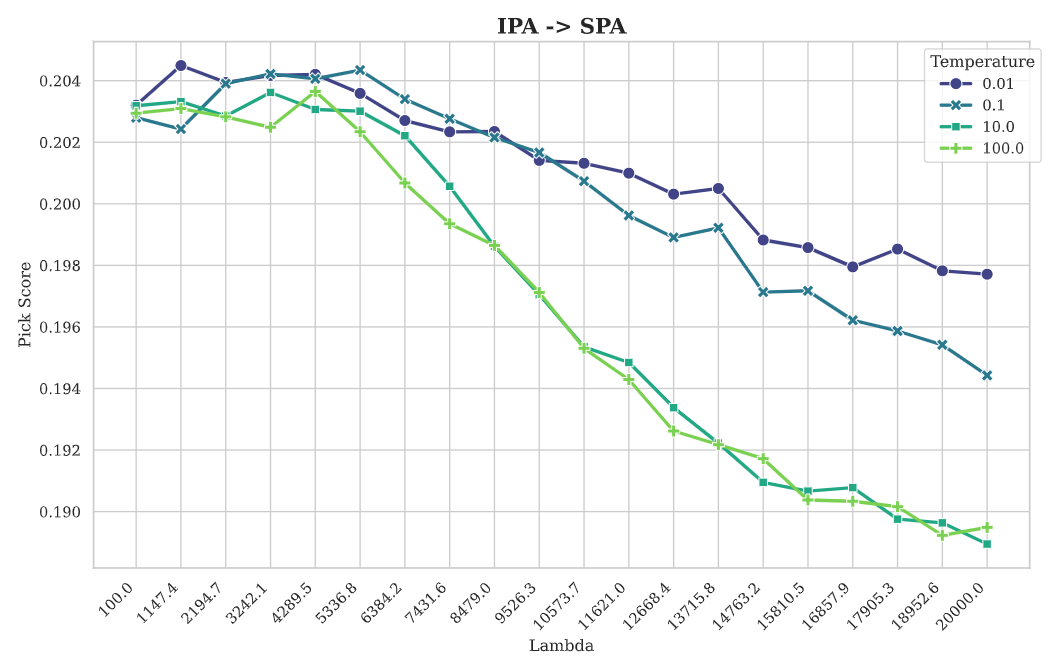}\\
        (d) Peak (SPA)
    \end{minipage}
    \caption{Sensitivity plots illustrating the relationship between $\lambda$, temperature, and output quality.}
    \label{fig:multi_source_exps}
\end{figure}

For multi-source composition (IPA $\rightarrow$ SPA), Figure~\ref{fig:multi_source_exps} shows tuning follows a smooth concave trend, making the effective range easy to identify. The best CLIP score, 0.263 is reached around  $\lambda \approx 9,500$ at temperature $0.1$. Performance is especially robust at lower temperatures  ($0.01$ to $0.1$), where CLIP scores remain high across a broad range of $\lambda$.

The sensitivity plots provide a tendency of  the relationship between $\lambda$, temperature, and output quality. While different compositions might require some tuning to balance the contributions of the conditions, adjusting conditioning strength is common when guiding diffusion models.

\subsection{More results on unconditional generation}\label{app:more_incond}
\begin{table*}[t]
\centering
\caption{\textbf{Distribution-level comparison of unconditional \ours generations on ImageNet \cite{deng2009imagenet}.}  We compare the standard SiT backbone~\cite{ma_sit_2024} (trained without representation alignment) against REPA \cite{wang_repa_2025} and REPA-E \cite{leng_repa-e_2025} variants. All models are trained on ImageNet. For each method block, the first row denotes metrics for unconditional generation, serving as a baseline for the \ours results. \textbf{\fproj} denotes features extracted after the projection layer in REPA-based models. \textcolor{gray!60}{Gray results} are from Table \ref{tab:ipa_imagenet_fids} in the main text for ease of comparison.}
\label{tab:ipa_imagenet_fids_proj}

\resizebox{\linewidth}{!}{%
\begin{tabular}{ll|ccccc|ccccc|ccccc}
\toprule
& & \multicolumn{5}{c|}{\textbf{Full Feature Map}}
& \multicolumn{5}{c|}{\textbf{Masked Feature Map}}
& \multicolumn{5}{c}{\textbf{Average Feature Map}} \\
 \cmidrule(lr){3-7} \cmidrule(lr){8-12} \cmidrule(lr){13-17}

\textbf{Model} & \textbf{Cond.}
& \textbf{FID}$\downarrow$ & \textbf{sFID}$\downarrow$ & \textbf{IS}$\uparrow$ & \textbf{Prec.}$\uparrow$ & \textbf{Rec.}$\uparrow$
& \textbf{FID}$\downarrow$ & \textbf{sFID}$\downarrow$ & \textbf{IS}$\uparrow$ & \textbf{Prec.}$\uparrow$ & \textbf{Rec.}$\uparrow$
& \textbf{FID}$\downarrow$ & \textbf{sFID}$\downarrow$ & \textbf{IS}$\uparrow$ & \textbf{Prec.}$\uparrow$ & \textbf{Rec.}$\uparrow$ \\
\midrule

\multirow{2}{*}{\textcolor{gray!60}{SiT}}
    & \textcolor{gray!60}{-}
            & \textcolor{gray!60}{\textbf{45.02}}
            & \textcolor{gray!60}{\textbf{9.02}}
            & \textcolor{gray!60}{\textbf{22.33}}
            & \textcolor{gray!60}{\textbf{0.50}}
            & \textcolor{gray!60}{\textbf{0.63}}
            & \textcolor{gray!60}{\textbf{45.02}}
            & \textcolor{gray!60}{\textbf{9.02}}
            & \textcolor{gray!60}{\textbf{22.33}}
            & \textcolor{gray!60}{\textbf{0.50}}
            & \textcolor{gray!60}{\textbf{0.63}}
            & \textcolor{gray!60}{\textbf{45.02}}
            & \textcolor{gray!60}{\textbf{9.02}}
            & \textcolor{gray!60}{\textbf{22.33}}
            & \textcolor{gray!60}{\textbf{0.50}}
            & \textcolor{gray!60}{\textbf{0.63}}
 \\
    & \textcolor{gray!60}{\fbase}
            & \textcolor{gray!60}{71.35}
            & \textcolor{gray!60}{72.08}
            & \textcolor{gray!60}{14.73}
            & \textcolor{gray!60}{0.33}
            & \textcolor{gray!60}{0.54}
            & \textcolor{gray!60}{51.88}
            & \textcolor{gray!60}{42.36}
            & \textcolor{gray!60}{15.76}
            & \textcolor{gray!60}{0.27}
            & \textcolor{gray!60}{0.52}
            & \textcolor{gray!60}{65.99}
            & \textcolor{gray!60}{22.52}
            & \textcolor{gray!60}{16.69}
            & \textcolor{gray!60}{0.36}
            & \textcolor{gray!60}{0.52}
 \\

\midrule
\multirow{4}{*}{\textcolor{gray!60}{REPA}}
    & \textcolor{gray!60}{-}
            & \textcolor{gray!60}{26.58}
            & \textcolor{gray!60}{6.85}
            & \textcolor{gray!60}{41.31}
            & \textcolor{gray!60}{0.56}
            & \textcolor{gray!60}{\textbf{0.70}}
            & \textcolor{gray!60}{26.58}
            & \textcolor{gray!60}{6.85}
            & \textcolor{gray!60}{41.31}
            & \textcolor{gray!60}{0.56}
            & \textcolor{gray!60}{\textbf{0.70}}
            & \textcolor{gray!60}{26.58}
            & \textcolor{gray!60}{6.85}
            & \textcolor{gray!60}{41.31}
            & \textcolor{gray!60}{0.56}
            & \textcolor{gray!60}{\textbf{0.70}}
 \\
    & \textcolor{gray!60}{\fdino}
            & \textcolor{gray!60}{7.23}
            & \textcolor{gray!60}{8.86}
            & \textcolor{gray!60}{199.69}
            & \textcolor{gray!60}{0.65}
            & \textcolor{gray!60}{0.69}
            & \textcolor{gray!60}{14.17}
            & \textcolor{gray!60}{10.59}
            & \textcolor{gray!60}{135.08}
            & \textcolor{gray!60}{0.58}
            & \textcolor{gray!60}{0.64}
            & \textcolor{gray!60}{29.06}
            & \textcolor{gray!60}{16.79}
            & \textcolor{gray!60}{83.85}
            & \textcolor{gray!60}{0.46}
            & \textcolor{gray!60}{0.69}
 \\
    & \textcolor{gray!60}{\falign}
            & \textcolor{gray!60}{\textbf{2.09}}
            & \textcolor{gray!60}{\textbf{6.17}}
            & \textcolor{gray!60}{\textbf{260.97}}
            & \textcolor{gray!60}{\textbf{0.74}}
            & \textcolor{gray!60}{\textbf{0.70}}
            & \textcolor{gray!60}{\textbf{2.67}}
            & \textcolor{gray!60}{\textbf{4.86}}
            & \textcolor{gray!60}{\textbf{222.71}}
            & \textcolor{gray!60}{\textbf{0.74}}
            & \textcolor{gray!60}{0.69}
            & \textcolor{gray!60}{\textbf{6.26}}
            & \textcolor{gray!60}{\textbf{6.14}}
            & \textcolor{gray!60}{\textbf{159.14}}
            & \textcolor{gray!60}{\textbf{0.68}}
            & \textcolor{gray!60}{\textbf{0.70}}
 \\
    & \fproj
            & 6.65
            & 7.59
            & 186.60
            & 0.66
            & 0.70
            & 12.29
            & 9.72
            & 135.77
            & 0.61
            & 0.65
            & 22.10
            & 13.86
            & 92.34
            & 0.54
            & 0.66
 \\

\midrule
\multirow{3}{*}{\textcolor{gray!60}{REPA-E}}
    & \textcolor{gray!60}{-}
            & \textcolor{gray!60}{15.07}
            & \textcolor{gray!60}{4.46}
            & \textcolor{gray!60}{55.14}
            & \textcolor{gray!60}{0.65}
            & \textcolor{gray!60}{\textbf{0.69}}
            & \textcolor{gray!60}{15.07}
            & \textcolor{gray!60}{4.46}
            & \textcolor{gray!60}{55.14}
            & \textcolor{gray!60}{0.65}
            & \textcolor{gray!60}{\textbf{0.69}}
            & \textcolor{gray!60}{15.07}
            & \textcolor{gray!60}{\textbf{4.46}}
            & \textcolor{gray!60}{55.14}
            & \textcolor{gray!60}{0.65}
            & \textcolor{gray!60}{\textbf{0.69}}
 \\
    & \textcolor{gray!60}{\fdino}
            & \textcolor{gray!60}{\textbf{1.45}}
            & \textcolor{gray!60}{\textbf{4.07}}
            & \textcolor{gray!60}{\textbf{264.69}}
            & \textcolor{gray!60}{\textbf{0.76}}
            & \textcolor{gray!60}{\textbf{0.69}}
            & \textcolor{gray!60}{2.30}
            & \textcolor{gray!60}{4.84}
            & \textcolor{gray!60}{212.25}
            & \textcolor{gray!60}{0.75}
            & \textcolor{gray!60}{0.67}
            & \textcolor{gray!60}{3.24}
            & \textcolor{gray!60}{5.20}
            & \textcolor{gray!60}{188.92}
            & \textcolor{gray!60}{0.73}
            & \textcolor{gray!60}{0.67}
 \\
    & \textcolor{gray!60}{\falign}
            & \textcolor{gray!60}{2.15}
            & \textcolor{gray!60}{6.64}
            & \textcolor{gray!60}{264.48}
            & \textcolor{gray!60}{0.75}
            & \textcolor{gray!60}{0.68}
            & \textcolor{gray!60}{\textbf{1.79}}
            & \textcolor{gray!60}{\textbf{4.13}}
            & \textcolor{gray!60}{\textbf{237.37}}
            & \textcolor{gray!60}{\textbf{0.76}}
            & \textcolor{gray!60}{0.68}
            & \textcolor{gray!60}{\textbf{2.50}}
            & \textcolor{gray!60}{4.81}
            & \textcolor{gray!60}{\textbf{201.13}}
            & \textcolor{gray!60}{\textbf{0.74}}
            & \textcolor{gray!60}{\textbf{0.69}}
 \\
    & \fproj
            & \textbf{1.42}
            & 4.10
            & 257.23
            & \textbf{0.76}
            & \textbf{0.70}
            & 2.01
            & 4.73
            & 216.19
            & 0.76
            & 0.67
            & 2.65
            & 5.05
            & 191.79
            & \textbf{0.76}
            & 0.65
 \\
\bottomrule
\end{tabular}}
\end{table*}

\begin{table*}[t]
\centering
\caption{\textbf{Instance-level comparison of unconditional \ours generations on ImageNet \cite{deng2009imagenet}.} Setup is the same as in Table~\ref{tab:ipa_imagenet_fids}. Alignment is measured with DINOv2 \cite{oquab2024dinov2}, JEPA \cite{jepa}, CLIP \cite{radford2021learning}, MAE \cite{he2022masked} and MoCov3 \cite{chen2021empirical} feature spaces, supplemented by PSNR for pixel-level fidelity. \textbf{\fproj} denotes features extracted after the projection layer in REPA-based models. \textcolor{gray!60}{Gray results} are from Table \ref{tab:ipa_imagenet_fids} in the main text for ease of comparison.}
\label{tab:ipa_imagenet_sim_w_other_feats}

\resizebox{\linewidth}{!}{%
\begin{tabular}{ll|cccccc|cccccc|cccccc}
\toprule
&  
& \multicolumn{6}{c|}{\textbf{Full Feature Map}}
& \multicolumn{6}{c|}{\textbf{Masked Feature Map}}
& \multicolumn{6}{c}{\textbf{Average Feature Map}} \\
\cmidrule(lr){3-8} \cmidrule(lr){9-14} \cmidrule(lr){15-20}
\textbf{Model} & \textbf{Feat}.
& \textcolor{gray!60}{\textbf{DINOv2}} & \textcolor{gray!60}{\textbf{JEPA}} & \textcolor{gray!60}{\textbf{CLIP}} & \textbf{MAE} & \textbf{MoCo} & \textcolor{gray!60}{\textbf{PSNR}}
& \textcolor{gray!60}{\textbf{DINOv2}} & \textcolor{gray!60}{\textbf{JEPA}} & \textcolor{gray!60}{\textbf{CLIP}} & \textbf{MAE} & \textbf{MoCo} & \textcolor{gray!60}{\textbf{PSNR}}
& \textcolor{gray!60}{\textbf{DINOv2}} & \textcolor{gray!60}{\textbf{JEPA}} & \textcolor{gray!60}{\textbf{CLIP}} & \textbf{MAE} & \textbf{MoCo} & \textcolor{gray!60}{\textbf{PSNR}} \\
\midrule

\multirow{1}{*}{\textcolor{gray!60}{SiT}} & \textcolor{gray!60}{\fbase} 
    & \textcolor{gray!60}{\textbf{0.27}} & \textcolor{gray!60}{\textbf{0.36}} & \textcolor{gray!60}{\textbf{0.43}} & \textbf{0.93} & \textbf{0.78} & \textcolor{gray!60}{\textbf{15.35}}
    & \textcolor{gray!60}{\textbf{0.44}} & \textcolor{gray!60}{\textbf{0.48}} & \textcolor{gray!60}{\textbf{0.46}} & \textbf{0.95} & \textbf{0.82} & \textcolor{gray!60}{\textbf{17.98}}
    & \textcolor{gray!60}{\textbf{0.12}} & \textcolor{gray!60}{\textbf{0.35}} & \textcolor{gray!60}{\textbf{0.83}} & \textbf{0.99} & \textbf{0.95} & \textcolor{gray!60}{\textbf{7.74}}
    \\
    
\midrule
\multirow{3}{*}{\textcolor{gray!60}{REPA}}
 & \textcolor{gray!60}{\fdino} 
    & \textcolor{gray!60}{0.75} & \textcolor{gray!60}{0.61} & \textcolor{gray!60}{0.58} & 0.94 & 0.86 & \textcolor{gray!60}{11.02}
    & \textcolor{gray!60}{0.77} & \textcolor{gray!60}{0.60} & \textcolor{gray!60}{0.57} & 0.94 & 0.84 & \textcolor{gray!60}{10.63}
    & \textcolor{gray!60}{0.76} & \textcolor{gray!60}{0.69} & \textcolor{gray!60}{0.92} & 0.99 & 0.98 & \textcolor{gray!60}{8.00}
    \\
 & \textcolor{gray!60}{\fbase}
    & \textcolor{gray!60}{\textbf{0.85}} & \textcolor{gray!60}{\textbf{0.78}} & \textcolor{gray!60}{\textbf{0.71}} & \textbf{0.98} & \textbf{0.95} & \textcolor{gray!60}{\textbf{20.46}}
    & \textcolor{gray!60}{\textbf{0.87}} & \textcolor{gray!60}{\textbf{0.79}} & \textcolor{gray!60}{\textbf{0.71}} & \textbf{0.98} & \textbf{0.94} & \textcolor{gray!60}{\textbf{20.72}}
    & \textcolor{gray!60}{\textbf{0.84}} & \textcolor{gray!60}{\textbf{0.84}} & \textcolor{gray!60}{\textbf{0.95}} & \textbf{0.99} & \textbf{0.990} & \textcolor{gray!60}{\textbf{10.82}}
    \\
 & \fproj 
    & {0.72} & {0.63} & {0.59} & 0.94 & 0.87 & {11.32}
    & {0.74} & {0.63} & {0.58} & 0.94 & 0.86 & {11.02}
    & {0.77} & {0.72} & {0.92} & 0.99 & 0.98 & {8.01}
    \\
\midrule
\multirow{3}{*}{\textcolor{gray!60}{REPA-E}}
 & \textcolor{gray!60}{\fdino} 
    & \textcolor{gray!60}{0.83} & \textcolor{gray!60}{0.69} & \textcolor{gray!60}{0.64} & 0.96 & 0.90 & \textcolor{gray!60}{15.23}
    & \textcolor{gray!60}{0.85} & \textcolor{gray!60}{0.68} & \textcolor{gray!60}{0.63} & 0.95 & 0.88 & \textcolor{gray!60}{14.36}
    & \textcolor{gray!60}{\textbf{0.91}} & \textcolor{gray!60}{0.83} & \textcolor{gray!60}{0.96} & 0.99 & 0.99 & \textcolor{gray!60}{10.97}
    \\
 & \textcolor{gray!60}{\fbase}
    & \textcolor{gray!60}{\textbf{0.86}} & \textcolor{gray!60}{\textbf{0.76}} & \textcolor{gray!60}{\textbf{0.71}} & \textbf{0.97} & \textbf{0.94} & \textcolor{gray!60}{\textbf{18.68}}
    & \textcolor{gray!60}{\textbf{0.89}} & \textcolor{gray!60}{\textbf{0.78}} & \textcolor{gray!60}{\textbf{0.71}} & \textbf{0.98} & \textbf{0.94} & \textcolor{gray!60}{\textbf{19.07}}
    & \textcolor{gray!60}{0.91} & \textcolor{gray!60}{\textbf{0.88}} & \textcolor{gray!60}{\textbf{0.96}} & \textbf{0.99} & \textbf{0.99} & \textcolor{gray!60}{\textbf{11.96}}
    \\
 & \fproj 
    & 0.81 & 0.70 & 0.65 & 0.96 & 0.91 & 15.91
    & 0.83 & 0.71 & 0.65 & 0.96 & 0.90 & 15.13
    & 0.90 & 0.86 & 0.96 & 0.99 & 0.99 & 11.12
    \\
\bottomrule
\end{tabular}}
\end{table*}

\begin{table*}[htb]
\centering
\caption{\textbf{Zero-shot evaluation on COCO \cite{coco} dataset using instance-level metrics.} Alignment is measured with DINOv2 \cite{oquab2024dinov2}, JEPA \cite{jepa}, CLIP \cite{radford2021learning}, MAE \cite{he2022masked} and MoCov3 \cite{chen2021empirical} feature spaces, supplemented by PSNR for pixel-level fidelity. \textbf{\fproj} denotes features extracted after the projection layer in REPA-based models.}
\label{tab:ipa_coco_sim}

\resizebox{\linewidth}{!}{%
\begin{tabular}{ll|SSSSSS[table-format=2.2]|SSSSSS[table-format=2.2]|SSSSSS[table-format=2.2]}
\toprule
& 
& \multicolumn{6}{c|}{\textbf{Full Feature Map}}
& \multicolumn{6}{c|}{\textbf{Masked Feature Map}}
& \multicolumn{6}{c}{\textbf{Average Feature Map}} \\
\cmidrule(lr){3-8} \cmidrule(lr){9-14} \cmidrule(lr){15-20}
\textbf{Model} & \textbf{Cond.}
& \textbf{DINOv2} & \textbf{JEPA} & \textbf{CLIP} & \textbf{MAE} & \textbf{MoCo} & \textbf{PSNR}
& \textbf{DINOv2} & \textbf{JEPA} & \textbf{CLIP} & \textbf{MAE} & \textbf{MoCo} & \textbf{PSNR}
& \textbf{DINOv2} & \textbf{JEPA} & \textbf{CLIP} & \textbf{MAE} & \textbf{MoCo} & \textbf{PSNR} \\
\midrule

\multirow{ 1}{*}{ SiT}
     &
    \fbase

            & 
                    \textbf{\round{0.246}}
            & 
                    \textbf{\round{0.348}}
            & 
                    \textbf{\round{0.401}}
            & 
                    \textbf{\round{0.929}}
            & 
                    \textbf{\round{0.764}}
            & 
                    \textbf{\round{15.176}}
            & 
                    \textbf{\round{0.405}}
            & 
                    \textbf{\round{0.452}}
            & 
                    \textbf{\round{0.439}}
            & 
                    \textbf{\round{0.948}}
            & 
                    \textbf{\round{0.818}}
            & 
                    \textbf{\round{18.556}}
            & 
                    \textbf{\round{0.110}}
            & 
                    \textbf{\round{0.378}}
            & 
                    \textbf{\round{0.814}}
            & 
                    \textbf{0.99} 
            & 
                    \textbf{\round{0.953}}
            & 
                    \textbf{\round{7.661}}
    \\
\midrule
\multirow{ 3}{*}{ REPA}
&
    \fdino 

            & 
                    0.737
            & 
                    0.603
            & 
                    0.564
            & 
                    0.935
            & 
                    0.847
            & 
                    10.718
            & 
                    0.746
            & 
                    0.561
            & 
                    0.553
            & 
                    0.933
            & 
                    0.835
            & 
                    10.846
            & 
                    0.736
            & 
                    0.679
            & 
                    0.900
            & 
                    0.992
            & 
                    0.979
            & 
                    7.724
    \\
 &
    \fbase 

            & 
                    \textbf{\round{0.843}}
            & 
                    \textbf{\round{0.780}}
            & 
                    \textbf{\round{0.699}}
            & 
                    \textbf{\round{0.977}}
            & 
                    \textbf{\round{0.949}}
            & 
                    \textbf{\round{20.609}}
            & 
                    \textbf{\round{0.841}}
            & 
                    \textbf{\round{0.760}}
            & 
                    \textbf{\round{0.687}}
            & 
                    \textbf{\round{0.977}}
            & 
                    \textbf{\round{0.945}}
            & 
                    \textbf{\round{21.382}}
            & 
                    \textbf{\round{0.802}}
            & 
                    \textbf{\round{0.825}}
            & 
                    \textbf{\round{0.924}}
            & 
                    \textbf{0.99}
            & 
                    \textbf{\round{0.991}}
            & 
                    \textbf{\round{10.474}}
    \\
&
    \fproj

            & 
                    0.697
            & 
                    0.627
            & 
                    0.564
            & 
                    0.939
            & 
                    0.866
            & 
                    11.007
            & 
                    0.705
            & 
                    0.593
            & 
                    0.556
            & 
                    0.938
            & 
                    0.857
            & 
                    11.275
            & 
                    0.732
            & 
                    0.701
            & 
                    0.902
            & 
                    0.993
            & 
                    0.982
            & 
                    7.727
    \\
\midrule
\multirow{ 3}{*}{ REPA-E}
    &
    \fdino 

            & 
                    0.821
            & 
                    0.677
            & 
                    0.626
            & 
                    0.951
            & 
                    0.892
            & 
                    14.776
            & 
                    0.838
            & 
                    0.642
            & 
                    0.616
            & 
                    0.949
            & 
                    0.882
            & 
                    14.715
            & 
                    \textbf{\round{0.901}}
            & 
                    0.821
            & 
                    0.943
            & 
                    0.996
            & 
                    0.989
            & 
                    10.663
    \\
 &
    \fbase 

            & 
                    \textbf{\round{0.845}}
            & 
                    \textbf{\round{0.765}}
            & 
                    \textbf{\round{0.696}}
            & 
                    \textbf{\round{0.973}}
            & 
                    \textbf{\round{0.939}}
            & 
                    \textbf{\round{18.581}}
            & 
                    \textbf{\round{0.856}}
            & 
                    \textbf{\round{0.752}}
            & 
                    \textbf{\round{0.692}}
            & 
                    \textbf{\round{0.975}}
            & 
                    \textbf{\round{0.937}}
            & 
                    \textbf{\round{19.668}}
            & 
                    0.886
            & 
                    \textbf{\round{0.858}}
            & 
                    \textbf{\round{0.946}}
            & 
                    \textbf{0.99}
            & 
                    \textbf{\round{0.992}}
            & 
                    \textbf{\round{11.627}}
    \\
 &
    \fproj

            & 
                    0.791
            & 
                    0.696
            & 
                    0.631
            & 
                    0.954
            & 
                    0.905
            & 
                    15.570
            & 
                    0.807
            & 
                    0.670
            & 
                    0.625
            & 
                    0.954
            & 
                    0.898
            & 
                    15.617
            & 
                    0.884
            & 
                    0.835
            & 
                    0.943
            & 
                    0.997
            & 
                    0.991
            & 
                    10.887
    \\

\bottomrule
\end{tabular}}
\end{table*}
\begin{table*}[t]
\centering
\caption{\textbf{Distribution-level comparison of class-conditional \ours generations on ImageNet \cite{deng2009imagenet}.}  We compare the standard SiT backbone~\cite{ma_sit_2024} (trained without representation alignment) against REPA \cite{wang_repa_2025} and REPA-E \cite{leng_repa-e_2025} variants. All models are trained on ImageNet. For each method block, the first row denotes metrics for unconditional generation, serving as a baseline for the \ours results. \fproj denotes features extracted after the projection layer in REPA-based models.}
\label{tab:cond_ipa_imagenet_fids}

\resizebox{\linewidth}{!}{%
\begin{tabular}{ll|ccccc|ccccc|ccccc}
\toprule
& &  \multicolumn{5}{c|}{\textbf{Full Feature Map}}
& \multicolumn{5}{c|}{\textbf{Masked Feature Map}}
& \multicolumn{5}{c}{\textbf{Average Feature Map}} \\
 \cmidrule(lr){3-7} \cmidrule(lr){8-12} \cmidrule(lr){13-17}
\textbf{Model} &  \textbf{Cond.}
& \textbf{FID}$\downarrow$ & \textbf{sFID}$\downarrow$ & \textbf{IS}$\uparrow$ & \textbf{Prec.}$\uparrow$ & \textbf{Rec.}$\uparrow$
& \textbf{FID}$\downarrow$ & \textbf{sFID}$\downarrow$ & \textbf{IS}$\uparrow$ & \textbf{Prec.}$\uparrow$ & \textbf{Rec.}$\uparrow$
& \textbf{FID}$\downarrow$ & \textbf{sFID}$\downarrow$ & \textbf{IS}$\uparrow$ & \textbf{Prec.}$\uparrow$ & \textbf{Rec.}$\uparrow$ \\
\midrule

\multirow{2}{*}{SiT}

    & -

            &                     \textbf{10.17}
            &                     \textbf{8.53}
            &                     \textbf{124.93}
            &                     \textbf{0.67}
            &                     \textbf{0.67}
            &                     \textbf{10.17}
            &                     \textbf{8.53}
            &                     \textbf{124.93}
            &                     \textbf{0.67}
            &                     \textbf{0.67}
            &                     \textbf{10.17}
            &                     \textbf{8.53}
            &                     \textbf{124.93}
            &                     \textbf{0.67}
            &                     \textbf{0.67}
 \\

    & \fbase

            &                     71.05
            &                     49.21
            &                     17.42
            &                     0.29
            &                     0.58
            &                     62.65
            &                     13.98
            &                     19.82
            &                     0.34
            &                     0.65
            &                     130.14
            &                     53.16
            &                     2.08
            &                     0.50
            &                     0.04
 \\

\midrule
\multirow{4}{*}{REPA}

    & -

            &                     5.92
            &                     \textbf{5.92}
            &                     157.10
            &                     0.70
            &                     \textbf{0.68}
            &                     5.92
            &                     5.92
            &                     157.10
            &                     0.70
            &                     \textbf{0.68}
            &                     \textbf{5.92}
            &                     \textbf{5.92}
            &                     \textbf{157.10}
            &                     \textbf{0.70}
            &                     \textbf{0.68}
 \\

    & \fdino

            &                     6.24
            &                     8.43
            &                     197.64
            &                     0.69
            &                     0.64
            &                     13.68
            &                     10.21
            &                     127.72
            &                     0.61
            &                     0.60
            &                     27.94
            &                     15.32
            &                     84.04
            &                     0.49
            &                     0.66
 \\

    & \fbase

            &                     \textbf{2.14}
            &                     6.71
            &                     \textbf{254.99}
            &                     \textbf{0.77}
            &                     0.65
            &                     \textbf{3.34}
            &                     \textbf{5.68}
            &                     \textbf{199.55}
            &                     \textbf{0.77}
            &                     0.63
            &                     11.24
            &                     8.59
            &                     122.54
            &                     0.63
            &                     0.67
 \\

    & \fproj

            &                     5.60
            &                     7.14
            &                     186.45
            &                     0.71
            &                     0.64
            &                     11.77
            &                     9.41
            &                     131.34
            &                     0.64
            &                     0.60
            &                     21.98
            &                     13.61
            &                     91.11
            &                     0.56
            &                     0.62
 \\

\midrule
\multirow{4}{*}{REPA-E}

    & -

            &                     1.71
            &                     4.19
            &                     220.60
            &                     0.77
            &                     0.66
            &                     \textbf{1.71}
            &                     \textbf{4.19}
            &                     220.60
            &                     0.77
            &                     \textbf{0.66}
            &                     \textbf{1.71}
            &                     \textbf{4.19}
            &                     \textbf{220.60}
            &                     \textbf{0.77}
            &                     0.66
 \\

    & \fdino

            &                     1.53
            &                     4.10
            &                     266.85
            &                     0.78
            &                     0.66
            &                     3.46
            &                     5.03
            &                     188.62
            &                     0.74
            &                     0.65
            &                     8.07
            &                     7.41
            &                     160.08
            &                     0.66
            &                     \textbf{0.69}
 \\

    & \fbase

            &                     1.97
            &                     6.67
            &                     \textbf{269.26}
            &                     0.78
            &                     0.66
            &                     1.98
            &                     4.24
            &                     \textbf{223.53}
            &                     \textbf{0.77}
            &                     0.65
            &                     3.23
            &                     5.39
            &                     192.17
            &                     0.72
            &                     0.67
 \\

    & \fproj

            &                     \textbf{1.53}
            &                     \textbf{4.04}
            &                     257.50
            &                     \textbf{0.78}
            &                     \textbf{0.66}
            &                     2.85
            &                     4.76
            &                     193.55
            &                     0.75
            &                     0.65
            &                     6.12
            &                     7.37
            &                     165.95
            &                     0.70
            &                     0.66
 \\

\bottomrule
\end{tabular}}
\end{table*}
\begin{table*}[b]
\centering
\caption{\textbf{Distribution-level comparison of class-conditional \ours generations on ImageNet \cite{deng2009imagenet}.}  We compare the standard SiT backbone~\cite{ma_sit_2024} (trained without representation alignment) against REPA \cite{wang_repa_2025} and REPA-E \cite{leng_repa-e_2025} variants. All models are trained on ImageNet. For each method block, the first row denotes metrics for unconditional generation, serving as a baseline for the \ours results. \textbf{\fproj} denotes features extracted after the projection layer in REPA-based models.}
\label{tab:cond_ipa_imagenet_sim}

\resizebox{\linewidth}{!}{%
\begin{tabular}{ll|SSSSSS[table-format=2.2]|SSSSSS[table-format=2.2]|SSSSSS[table-format=2.2]}
\toprule
&   
& \multicolumn{6}{c|}{\textbf{Full Feature Map}}
& \multicolumn{6}{c|}{\textbf{Masked Feature Map}}
& \multicolumn{6}{c}{\textbf{Average Feature Map}} \\
 \cmidrule(lr){3-8} \cmidrule(lr){9-14} \cmidrule(lr){15-20}
\textbf{Model}  & \textbf{Cond.}
& \textbf{DINO} & \textbf{JEPA} & \textbf{CLIP} & \textbf{MAE} & \textbf{MoCo} & \textbf{PSNR}
& \textbf{DINO} & \textbf{JEPA} & \textbf{CLIP} & \textbf{MAE} & \textbf{MoCo} & \textbf{PSNR}
& \textbf{DINO} & \textbf{JEPA} & \textbf{CLIP} & \textbf{MAE} & \textbf{MoCo} & \textbf{PSNR} \\
\midrule

\multirow{ 1}{*}{ SiT}

&    
    \fbase 

            & 
                    \textbf{\round{0.305}}
            & 
                    \textbf{\round{0.393}}
            & 
                    \textbf{\round{0.432}}
            & 
                    \textbf{\round{0.936}}
            & 
                    \textbf{\round{0.785}}
            & 
                    \textbf{\round{14.794}}
            & 
                    \textbf{\round{0.4036}}
            & 
                    \textbf{\round{0.4441}}
            & 
                    \textbf{\round{0.4360}}
            & 
                    \textbf{\round{0.9456}}
            & 
                    \textbf{\round{0.8117}}
            & 
                    \textbf{\round{17.66}}
            & 
                    \textbf{\round{0.129}}
            & 
                    \textbf{\round{0.368}}
            & 
                    \textbf{\round{0.828}}
            & 
                    \textbf{\round{0.991}}
            & 
                    \textbf{\round{0.953}}
            & 
                    \textbf{\round{8.645}}
    \\
\midrule
\multirow{ 3}{*}{ REPA}
      &
     
    \fdino 

            & 
                    0.737
            & 
                    0.593
            & 
                    0.574
            & 
                    0.935
            & 
                    0.848
            & 
                    10.467
            & 
                    0.757
            & 
                    0.583
            & 
                    0.564
            & 
                    0.931
            & 
                    0.832
            & 
                    10.303
            & 
                    0.753
            & 
                    0.678
            & 
                    0.917
            & 
                    \textbf{0.99}
            & 
                    0.977
            & 
                    8.047
    \\
&    
    \fbase 

            & 
                    \textbf{\round{0.843}}
            & 
                    \textbf{\round{0.766}}
            & 
                    \textbf{\round{0.697}}
            & 
                    \textbf{\round{0.976}}
            & 
                    \textbf{\round{0.942}}
            & 
                    \textbf{\round{19.601}}
            & 
                    \textbf{\round{0.856}}
            & 
                    \textbf{\round{0.769}}
            & 
                    \textbf{\round{0.695}}
            & 
                    \textbf{\round{0.976}}
            & 
                    \textbf{\round{0.936}}
            & 
                    \textbf{\round{20.005}}
            & 
                    \textbf{\round{0.787}}
            & 
                    \textbf{\round{0.800}}
            & 
                    \textbf{\round{0.933}}
            & 
                    \textbf{0.99}
            & 
                    \textbf{\round{0.986}}
            & 
                    \textbf{\round{10.361}}
    \\
&    
    \fproj 

            & 
                    0.706
            & 
                    0.613
            & 
                    0.578
            & 
                    0.938
            & 
                    0.862
            & 
                    10.714
            & 
                    0.727
            & 
                    0.608
            & 
                    0.570
            & 
                    0.935
            & 
                    0.849
            & 
                    10.595
            & 
                    0.758
            & 
                    0.696
            & 
                    0.922
            & 
                    \textbf{0.99}
            & 
                    0.980
            & 
                    8.024
    \\
\midrule
\multirow{ 3}{*}{REPA-E}

&    
    \fdino 

            & 
                    0.821
            & 
                    0.675
            & 
                    0.634
            & 
                    0.952
            & 
                    0.892
            & 
                    14.785
            & 
                    0.839
            & 
                    0.656
            & 
                    0.620
            & 
                    0.947
            & 
                    0.873
            & 
                    13.835
            & 
                    0.882
            & 
                    0.788
            & 
                    0.949
            & 
                    \textbf{0.99}
            & 
                    0.985
            & 
                    10.340
    \\
&   
    \fbase 

            & 
                    \textbf{\round{0.853}}
            & 
                    \textbf{\round{0.761}}
            & 
                    \textbf{\round{0.702}}
            & 
                    \textbf{\round{0.974}}
            & 
                    \textbf{\round{0.938}}
            & 
                    \textbf{\round{18.567}}
            & 
                    \textbf{\round{0.873}}
            & 
                    \textbf{\round{0.772}}
            & 
                    \textbf{\round{0.705}}
            & 
                    \textbf{\round{0.974}}
            & 
                    \textbf{\round{0.931}}
            & 
                    \textbf{\round{18.806}}
            & 
                    \textbf{\round{0.892}}
            & 
                    \textbf{\round{0.854}}
            & 
                    \textbf{\round{0.959}}
            & 
                    \textbf{0.99}
            & 
                    \textbf{\round{0.989}}
            & 
                    \textbf{\round{11.585}}
    \\
&    
    \fproj 

            & 
                    0.800
            & 
                    0.691
            & 
                    0.641
            & 
                    0.955
            & 
                    0.902
            & 
                    15.330
            & 
                    0.819
            & 
                    0.679
            & 
                    0.631
            & 
                    0.951
            & 
                    0.885
            & 
                    14.414
            & 
                    0.879
            & 
                    0.801
            & 
                    0.952
            & 
                    \textbf{0.99}
            & 
                    0.987
            & 
                    10.450
    \\

\bottomrule
\end{tabular}}
\end{table*}

As detailed in Section \ref{sec:experiments}, we also evaluated features extracted after the projection layer for REPA-based models, but observed no significant performance difference. These results, provided in Table \ref{tab:ipa_imagenet_fids_proj}, complement those in Table \ref{tab:ipa_imagenet_fids}.

We further complement Table \ref{tab:ipa_imagenet_sim} by providing alignment scores within the MAE \cite{he2022masked} and MoCov3 \cite{chen2021empirical} feature spaces. As shown in Table \ref{tab:ipa_imagenet_sim_w_other_feats}, these scores remain consistent with the trends observed across other feature representations.
Similarly, we report alignment scores for the COCO dataset \cite{coco} in Table~\ref{tab:ipa_coco_sim}.

\subsection{Results on conditional generation} \label{app:more_cond}
Analogous to our unconditional experiments, we evaluate conditional generation by providing the class index alongside the noisy latent as input. Results for this setup are reported in Table \ref{tab:cond_ipa_imagenet_fids} for distribution-level metrics and Table \ref{tab:cond_ipa_imagenet_sim} for instance-level alignment.

\subsection{Analysis of the balance between stochastic reconstruction and generation diversity}
\label{app:reconstruction}

With full-feature-map conditioning and sufficiently large $\lambda$, \ours approaches stochastic reconstruction. This setting provides a useful sanity check for the guidance mechanism: recovering the spatial and semantic structure of an anchor image from Gaussian noise is not guaranteed, and both vanilla SiT and the adapted TFG baseline (\appref{app:tfg}) fail under the same feature-conditioning setup. Successful reconstruction therefore indicates that representation alignment produces an internal feature space capable of meaningfully steering the denoising trajectory.

Stochastic reconstruction is also useful for precise image manipulation. When combined with masks or additional conditions, it enables selected content to be preserved while allowing other regions to change. More generally, reconstruction and semantic generation form a continuum controlled by both the guidance strength $\lambda$ and the granularity of the conditioning features. Reducing $\lambda$ weakens the attraction toward the anchor and increases stochastic variation, while average-feature conditioning removes patch-wise spatial correspondence and retains primarily global semantic information, thereby producing more diverse samples. This behavior is analogous to the fidelity--diversity relationship commonly observed with classifier and classifier-free guidance.

Our qualitative results illustrate this continuum. In the multi-source examples in Figure~\ref{fig:main_quals}, full-feature conditioning preserves the semantic identity of the reference while still allowing substantial variations in properties such as color and shape. Similarly, in Figure~\ref{fig:teaser_cond_comparison}, combining rabbit and lava features produces interactions, such as orange illumination reflected on the rabbit's fur, that are not directly copied from either source but instead emerge from the generative prior.

We further quantify the ability of \ours to generate diverse samples from the same condition. For each of the 1,000 ImageNet classes, we randomly select one anchor image and generate 10 samples using different random seeds. In Figure~\ref{fig:diversity}, we report the mean pairwise LPIPS distance, pairwise DINOv2 patch cosine distance, and DINOv2-based Vendi score, where larger values indicate greater sample diversity. Across all metrics, two consistent trends emerge: diversity decreases as $\lambda$ increases, while average-feature conditioning remains substantially more diverse than full-map conditioning, particularly at large guidance scales, due to the removal of explicit patch-wise spatial constraints.

We repeat the same analysis using 1,000 randomly selected COCO anchors as a held-out dataset, which is out of distribution for the ImageNet-trained generator. The same trends are observed on COCO, with no reduction in diversity relative to ImageNet. Together with the quantitative COCO evaluation and the in-the-wild references in Figure~1, these results indicate that \ours is not simply retrieving or memorizing ImageNet training examples. Additional qualitative results with a lower guidance scale of $\lambda=2,000$ are shown in Figure~\ref{fig:diversity_spatial}.

\begin{figure}[tbh]
    \centering
    \textbf{ImageNet}\par\vspace{0.2cm}
    \begin{tikzpicture}
        \begin{axis}[
            width=0.32\textwidth,
            title={LPIPS Distance},
            xlabel={$\lambda$},
            xmode=log,
            xtick={2000, 10000, 50000},
            xticklabels={2k, 10k, 50k},
            legend pos=south west,
            legend style={nodes={scale=0.7, transform shape}, font=\footnotesize}
        ]
        \addplot+[error bars/.cd, y dir=both, y explicit] coordinates {
            (2000, 0.62) +- (0, 0.06)
            (10000, 0.57) +- (0, 0.07)
            (50000, 0.53) +- (0, 0.08)
        };
        \addlegendentry{avg.}
        \addplot+[error bars/.cd, y dir=both, y explicit] coordinates {
            (2000, 0.62) +- (0, 0.06)
            (10000, 0.46) +- (0, 0.09)
            (50000, 0.34) +- (0, 0.08)
        };
        \addlegendentry{full}
        \end{axis}
    \end{tikzpicture}%
    \hfill
    \begin{tikzpicture}
        \begin{axis}[
            width=0.32\textwidth,
            title={DINOv2 Cosine},
            xlabel={$\lambda$},
            xmode=log,
            xtick={2000, 10000, 50000},
            xticklabels={2k, 10k, 50k}
        ]
        \addplot+[error bars/.cd, y dir=both, y explicit] coordinates {
            (2000, 0.63) +- (0, 0.06)
            (10000, 0.55) +- (0, 0.07)
            (50000, 0.50) +- (0, 0.07)
        };
        \addplot+[error bars/.cd, y dir=both, y explicit] coordinates {
            (2000, 0.60) +- (0, 0.07)
            (10000, 0.33) +- (0, 0.07)
            (50000, 0.19) +- (0, 0.04)
        };
        \end{axis}
    \end{tikzpicture}%
    \hfill
    \begin{tikzpicture}
        \begin{axis}[
            width=0.32\textwidth,
            title={DINOv2 Vendi},
            xlabel={$\lambda$},
            xmode=log,
            xtick={2000, 10000, 50000},
            xticklabels={2k, 10k, 50k}
        ]
        \addplot+[error bars/.cd, y dir=both, y explicit] coordinates {
            (2000, 6.1) +- (0, 0.7)
            (10000, 5.2) +- (0, 0.8)
            (50000, 4.7) +- (0, 0.8)
        };
        \addplot+[error bars/.cd, y dir=both, y explicit] coordinates {
            (2000, 5.9) +- (0, 0.8)
            (10000, 3.3) +- (0, 0.6)
            (50000, 2.2) +- (0, 0.3)
        };
        \end{axis}
    \end{tikzpicture}
    
    \vspace{0.5cm}
    \textbf{COCO}\par\vspace{0.2cm}
    
    \begin{tikzpicture}
        \begin{axis}[
            width=0.32\textwidth,
            xlabel={$\lambda$},
            xmode=log,
            xtick={2000, 10000, 50000},
            xticklabels={2k, 10k, 50k}
        ]
        \addplot+[error bars/.cd, y dir=both, y explicit] coordinates {
            (2000, 0.64) +- (0, 0.05)
            (10000, 0.59) +- (0, 0.07)
            (50000, 0.56) +- (0, 0.08)
        };
        \addplot+[error bars/.cd, y dir=both, y explicit] coordinates {
            (2000, 0.64) +- (0, 0.05)
            (10000, 0.49) +- (0, 0.08)
            (50000, 0.35) +- (0, 0.07)
        };
        \end{axis}
    \end{tikzpicture}%
    \hfill
    \begin{tikzpicture}
        \begin{axis}[
            width=0.32\textwidth,
            xlabel={$\lambda$},
            xmode=log,
            xtick={2000, 10000, 50000},
            xticklabels={2k, 10k, 50k}
        ]
        \addplot+[error bars/.cd, y dir=both, y explicit] coordinates {
            (2000, 0.66) +- (0, 0.05)
            (10000, 0.58) +- (0, 0.06)
            (50000, 0.54) +- (0, 0.07)
        };
        \addplot+[error bars/.cd, y dir=both, y explicit] coordinates {
            (2000, 0.63) +- (0, 0.06)
            (10000, 0.38) +- (0, 0.06)
            (50000, 0.21) +- (0, 0.04)
        };
        \end{axis}
    \end{tikzpicture}%
    \hfill
    \begin{tikzpicture}
        \begin{axis}[
            width=0.32\textwidth,
            xlabel={$\lambda$},
            xmode=log,
            xtick={2000, 10000, 50000},
            xticklabels={2k, 10k, 50k}
        ]
        \addplot+[error bars/.cd, y dir=both, y explicit] coordinates {
            (2000, 6.5) +- (0, 0.6)
            (10000, 5.6) +- (0, 0.7)
            (50000, 5.1) +- (0, 0.7)
        };
        \addplot+[error bars/.cd, y dir=both, y explicit] coordinates {
            (2000, 6.3) +- (0, 0.8)
            (10000, 3.6) +- (0, 0.6)
            (50000, 2.3) +- (0, 0.3)
        };
        \end{axis}
    \end{tikzpicture}
    \caption{\textbf{Effect of guidance strength on generation diversity.} Impact of $\lambda$ on LPIPS distance, DINOv2 cosine distance, and DINOv2-based Vendi score for samples generated from ImageNet and COCO anchors.}
    \label{fig:diversity}
\end{figure}

\begin{figure}[tbh]
    \centering
    \setlength{\tabcolsep}{2pt}

\resizebox{\textwidth}{!}{
\begin{tabular}{c ccc ccc}
    \textbf{Anchor} & \multicolumn{3}{c}{\textbf{Average Feature}} & \multicolumn{3}{c}{\textbf{Full Feature Map}} \\
    \cmidrule(lr){2-4} \cmidrule(lr){5-7}

    \includegraphics[width=0.13\linewidth]{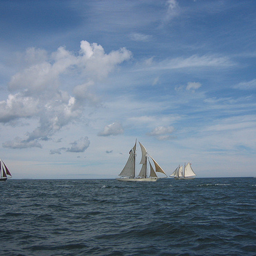} &
    \includegraphics[width=0.13\linewidth]{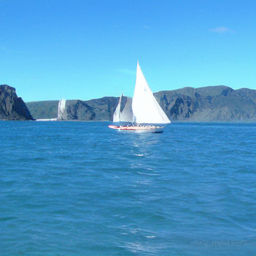} &
    \includegraphics[width=0.13\linewidth]{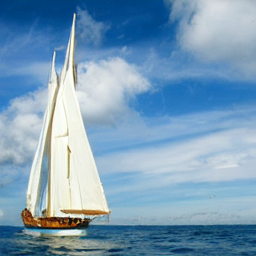} &
    \includegraphics[width=0.13\linewidth]{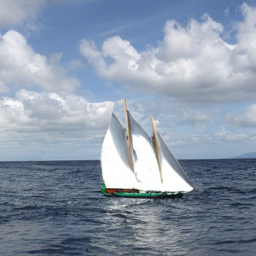} &
    \includegraphics[width=0.13\linewidth]{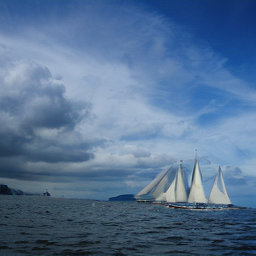} &
    \includegraphics[width=0.13\linewidth]{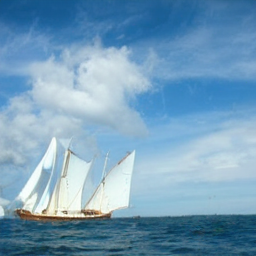} &
    \includegraphics[width=0.13\linewidth]{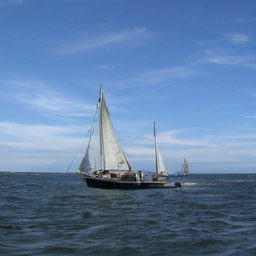} \\
    
    \addlinespace[4pt] 
    
    \includegraphics[width=0.13\linewidth]{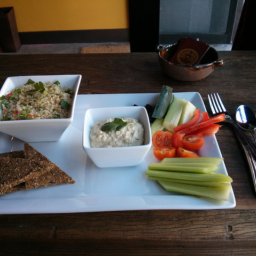} &
    \includegraphics[width=0.13\linewidth]{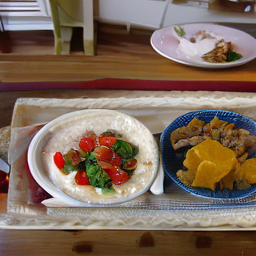} &
    \includegraphics[width=0.13\linewidth]{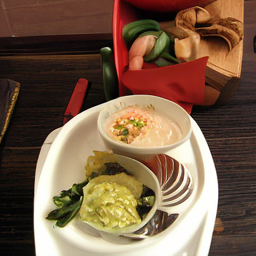} &
    \includegraphics[width=0.13\linewidth]{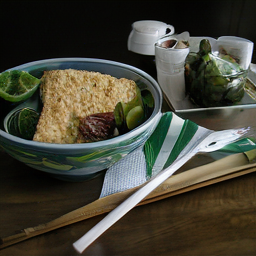} &
    \includegraphics[width=0.13\linewidth]{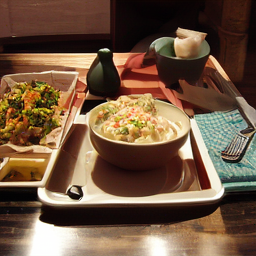} &
    \includegraphics[width=0.13\linewidth]{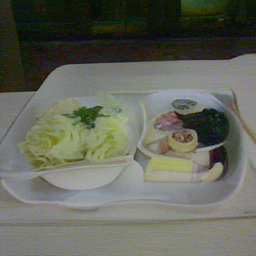} &
    \includegraphics[width=0.13\linewidth]{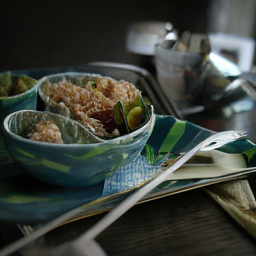} \\
\end{tabular}
}
\caption{\textbf{Qualitative results of generation diversity.} ImageNet examples are shown in the top row and COCO examples in the bottom row. For each example, the anchor image is shown on the left, followed by three samples generated with \textit{average feature} conditioning and three with \textit{full feature map} conditioning. We use a small guidance scale of $\lambda=2{,}000$.}
\label{fig:diversity_spatial}
\end{figure}

\subsection{Training free guidance with visual feature}\label{app:tfg}

TFG~\citep{ye_tfg_2024} presents a unified approach to bring together every training free guidance for diffusion models in the same framework. It can use as guidance signal the output of a classifier or any general energy function. Critically, contrarily to classifier guidance where the model is trained to classify noisy input, the guidance signal is extracted here from clean data. We introduce a new setting not tested in their work to see if TFG can be used with visual features as guidance signal. We use as target predictor the same energy function as in our method based on alignment of features patches. In \ours, the predicted features are extracted from the noisy latent with the aligned diffusion transformer. In contrast, they have to predict at each step an approximation of the clean latent sample and decode it to image space. Then, the it is possible to extract features from it with a vision encoder, here DINOv2, and compare them with target features. It introduces much more computational overhead, as the backpropagation is done through the vision encoder, the decoder and the entire diffusion network versus only the first 8 transformer blocks in our method.

We use their style transfer hyperparameters, as this is the closest task to ours. We set $N_{iter}=1$ and $\rho=0$ for faster sampling, and $\mu=1000$ to provide sufficiently strong conditioning. We use the REPA-E model for generation.

We show in Figure~\ref{fig:tfg_visual} that TFG fails in the simple setting of stochastic reconstruction of an image, i.e. guiding with a full feature map. While the condition is recognizable in the generated image, the quality is very poor due to visual artifacts. This behaviour is confirmed by the quantitative evaluation in Table~\ref{tab:tfg_performance}, where we report a very high FID and low alignment scores, except for DINOv2 alignment which is directly optimized. Lowering $\mu=1000$ doesn't allow to generate images with better quality that keeps the condition recognizable. 

We emphasize that a target predictor trained directly on noisy latents, such as ours, naturally outperforms a predictor operating on clean data, which relies on heavier approximations to guide generation. Indeed, such trained predictors are presented as an upper bound for TFG's performance in their original work. A key strength of \ours is that its target predictor emerges naturally from the diffusion model's alignment procedure, originally motivated solely by convergence speed and sample quality. Although never explicitly intended to serve as a guidance signal at inference, we demonstrate that it fulfills this role exceptionally well without requiring any additional training.

\begin{table}[tbh]
    \centering
    \caption{\textbf{TFG performance.} Evaluation of TFG with different guidance granularities.}
    \resizebox{\linewidth}{!}{%
    \begin{tabular}{llccccccccc}
        \toprule
        \textbf{Method} & \textbf{Features} & \textbf{FID} & \textbf{sFID} & \textbf{IS} & \textbf{Prec.} & \textbf{Rec.} & \textbf{DINOv2} & \textbf{JEPA} & \textbf{CLIP} & \textbf{PSNR} \\
        \midrule
        \multirow{3}{*}{\ours} & Full & \fpeval{round(1.45, 1)} & \fpeval{round(4.07, 1)} & \fpeval{round(264.69, 1)} & 0.76 & 0.69 & 0.83 & 0.69 & 0.64 & \fpeval{round(15.23, 1)} \\
        & Masked & \fpeval{round(2.30, 1)} & \fpeval{round(4.84, 1)} & \fpeval{round(212.25, 1)} & 0.75 & 0.67 & 0.85 & 0.68 & 0.63 & \fpeval{round(14.36, 1)} \\
        & Average & \fpeval{round(3.24, 1)} & \fpeval{round(5.20, 1)} & \fpeval{round(188.92, 1)} & 0.73 & 0.67 & 0.91 & 0.83 & 0.96 & 11.0 \\
        \midrule
        \multirow{3}{*}{TFG} & Full & 74.6 & 53.6 & 47.4 & 0.26 & 0.40 & 0.91 & 0.43 & 0.54 & 10.0 \\
        & Masked & 85.8 & 47.2 & 36.0 & 0.24 & 0.38 & 0.94 & 0.45 & 0.55 & 10.0 \\
        & Average & 81.1 & 37.0 & 35.8 & 0.23 & 0.41 & 0.99 & 0.51 & 0.91 & 8.5 \\
        \bottomrule
    \end{tabular}%
    }
    \label{tab:tfg_performance}
\end{table}

\subsection{Comparison with adapter-based conditioning}
\label{app:adapter}

Adapter-based methods learn additional image-prompt projection and cross-attention parameters from paired training data. In contrast, once a representation-aligned diffusion model is available, \ours requires no additional learned parameters, architectural modifications, or task-specific training. Instead, it directly exploits the model's aligned feature space at inference time and supports full-map, masked, global-average, single-patch, and multi-source conditioning within the same formulation. Unlike methods explicitly trained for structural conditioning, however, \ours does not provide direct control from inputs such as edge maps, depth maps, or poses.

\begin{figure}[tbh]
    \centering
    \begin{minipage}{\textwidth}
        \begin{minipage}{0.19\linewidth}\centering Anchor Image \end{minipage}\hfill
        \begin{minipage}{0.19\linewidth}\centering TFG \end{minipage}\hfill
        \begin{minipage}{0.19\linewidth}\centering IP-Adapter \end{minipage}\hfill
        \begin{minipage}{0.39\linewidth}
            \centering \textbf{Generated w/ REPA-G (Ours)}\\[0.2em]
            \begin{minipage}{0.48\linewidth}\centering REPA-E \end{minipage}\hfill
            \begin{minipage}{0.48\linewidth}\centering Self-Flow \end{minipage}
        \end{minipage}
        
        \vspace{0.4em}

        \begin{minipage}{0.19\linewidth}
            \centering
            \includegraphics[width=\linewidth]{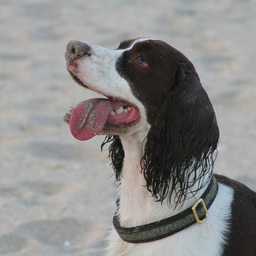}
        \end{minipage}\hfill
        \begin{minipage}{0.19\linewidth}
            \centering
            \includegraphics[width=\linewidth]{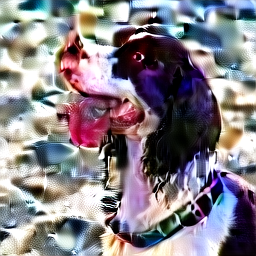}
        \end{minipage}\hfill
        \begin{minipage}{0.19\linewidth}
            \centering
            \includegraphics[width=\linewidth]{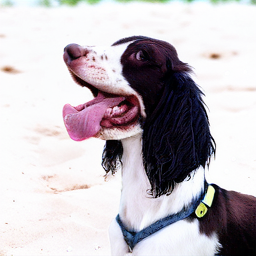}
        \end{minipage}\hfill
        \begin{minipage}{0.19\linewidth}
            \centering
            \includegraphics[width=\linewidth]{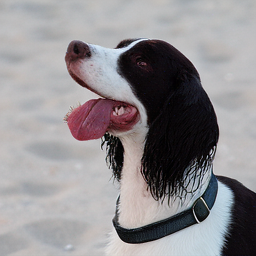}
        \end{minipage}\hfill
        \begin{minipage}{0.19\linewidth}
            \centering
            \includegraphics[width=\linewidth]{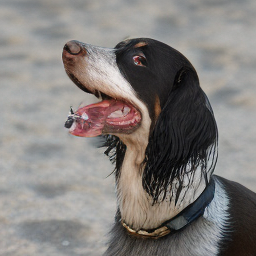}
        \end{minipage}
        
        \vspace{0.5em}

        \begin{minipage}{0.19\linewidth}
            \centering
            \includegraphics[width=\linewidth]{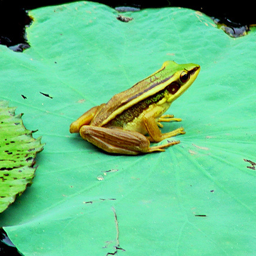}
        \end{minipage}\hfill
        \begin{minipage}{0.19\linewidth}
            \centering
            \includegraphics[width=\linewidth]{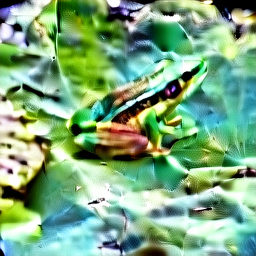}
        \end{minipage}\hfill
        \begin{minipage}{0.19\linewidth}
            \centering
            \includegraphics[width=\linewidth]{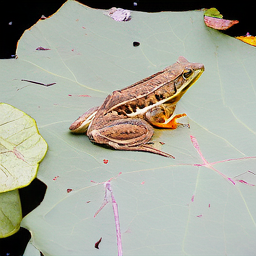}
        \end{minipage}\hfill
        \begin{minipage}{0.19\linewidth}
            \centering
            \includegraphics[width=\linewidth]{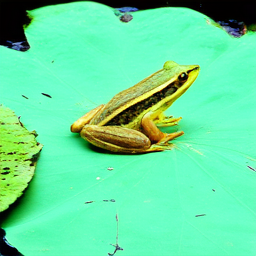}
        \end{minipage}\hfill
        \begin{minipage}{0.19\linewidth}
            \centering
            \includegraphics[width=\linewidth]{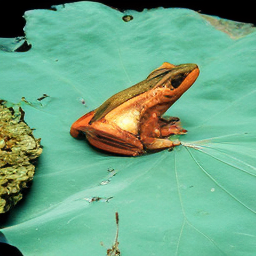}
        \end{minipage}
        
        \vspace{0.5em}

        \begin{minipage}{0.19\linewidth}
            \centering
            \includegraphics[width=\linewidth]{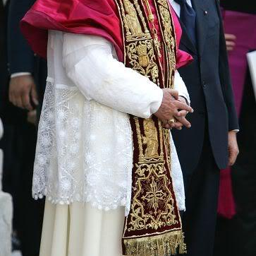}
        \end{minipage}\hfill
        \begin{minipage}{0.19\linewidth}
            \centering
            \includegraphics[width=\linewidth]{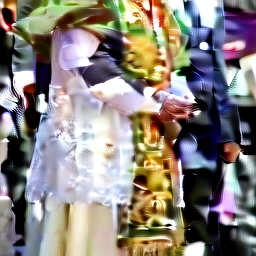}
        \end{minipage}\hfill
        \begin{minipage}{0.19\linewidth}
            \centering
            \includegraphics[width=\linewidth]{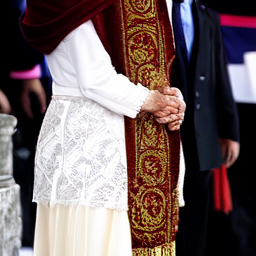}
        \end{minipage}\hfill
        \begin{minipage}{0.19\linewidth}
            \centering
            \includegraphics[width=\linewidth]{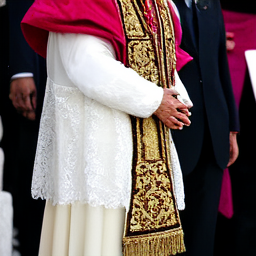}
        \end{minipage}\hfill
        \begin{minipage}{0.19\linewidth}
            \centering
            \includegraphics[width=\linewidth]{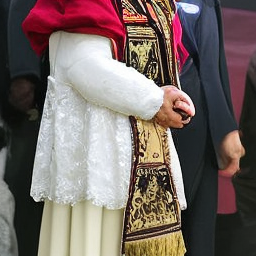}
        \end{minipage}
    \end{minipage}
    \caption{\textbf{Visual comparison with the training-free baseline TFG  \cite{ye_tfg_2024} and the trained IP-Adapter~\cite{ye2023ip-adapter}.} While the images generated with TFG respect the conditioning to the anchors, perceptual artifacts appear and degrade image quality. On the other hand, IP-adapter manages to generate images faithful to the anchor but requires additional training and parameters.} REPA-G on the Self-Flow model \cite{CheferEsser2026selfflow} also generates realistic images.
    \label{fig:tfg_visual}
\end{figure}

To better position \ours relative to learned adapter-based conditioning, we train an IP-Adapter-style module on ImageNet for 100k steps on top of REPA-E, using DINOv2 full-map features. The diffusion backbone is kept frozen, while the input projection, normalization, and image cross-attention parameters are optimized. During sampling, we apply classifier-free guidance to the image condition with a scale of 6. Qualitative results on ImageNet (see Figure~\ref{fig:tfg_visual}) show that the adapter follows the reference image, but preserves its spatial layout less precisely than \ours, consistent with the lower patch-wise DINOv2 alignment reported in Table~\ref{tab:ipadapter_fid}. Its distribution-level metrics also improve over the unconditional baseline.
Runtime results are reported in Table~\ref{tab:runtime}; due to classifier-free guidance and approximately 150M additional parameters, IP-Adapter also achieves noticeably lower throughput than \ours.

\begin{table}[htpb]
    \centering
    \caption{\textbf{Comparison with trained adapter-based conditioning.} We compare full-feature-map conditional generation using IP-Adapter and REPA-G on ImageNet and COCO, with REPA-E as the shared base model.}
    \resizebox{\textwidth}{!}{
    \begin{tabular}{l r r r r r r r r r r r r}
        \toprule
        & \multicolumn{6}{c}{\textbf{ImageNet}} & \multicolumn{6}{c}{\textbf{COCO}} \\
        \cmidrule(lr){2-7} \cmidrule(lr){8-13}
       \textbf{Condition}   & \textbf{FID}$\downarrow$ & \textbf{sFID}$\downarrow$& \textbf{IS}$\uparrow$ & \textbf{Prec.}$\uparrow$ & \textbf{Rec.}$\uparrow$ & \textbf{DINOv2}$\uparrow$ & \textbf{FID}$\downarrow$ & \textbf{sFID}$\downarrow$ & \textbf{IS}$\uparrow$ & \textbf{Prec.}$\uparrow$ & \textbf{Rec.}$\uparrow$ & \textbf{DINOv2}$\uparrow$ \\
        \midrule
        - & 15.07 & 4.46 & 55.14 & 0.65 & \textbf{0.69} & - & 36.23 & 29.98 & \textbf{55.13} & 0.51 & 0.59 & - \\
        IP-Adapter & 7.52 & 10.65 & 145.36 & 0.71 & 0.63 & 0.62 & 9.90 & 32.23 & 30.45 & 0.62 & 0.56 & 0.61 \\
        REPA-G & \textbf{1.45} & \textbf{4.07} & \textbf{264.69} & \textbf{0.76} & \textbf{0.69} & \textbf{0.83} & \textbf{4.61} & \textbf{23.27} & 35.97 & \textbf{0.67} & \textbf{0.64} & \textbf{0.82} \\

        \bottomrule
    \end{tabular}}
    \label{tab:ipadapter_fid}
\end{table}

\subsection{Resolution scalability}\label{app:scalability}
The supervision is done at the patch level, which is independent of the resolution. Therefore, our method should be able to scale to higher resolutions without any architectural modification. To validate this claim, we fine-tuned the 256×256 REPA model to output at 512×512 for 40k steps and evaluated the performance of \ours. The results below in Table~\ref{tab:resolution_scalability} confirm that our method successfully scales, maintaining superior image quality than unconditional sampling.

\begin{figure}[htbp]
    \centering
    \begin{tabular}{cccc}
        \includegraphics[width=0.18\linewidth]{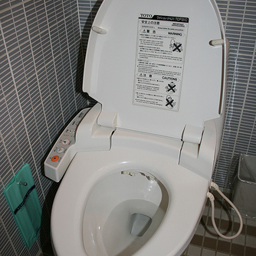} &
        \includegraphics[width=0.18\linewidth]{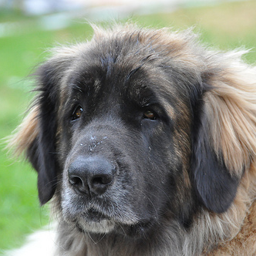} &
        \includegraphics[width=0.18\linewidth]{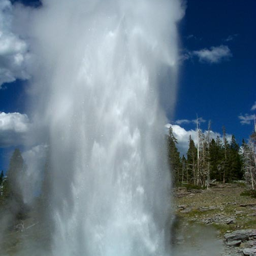} &
        \includegraphics[width=0.18\linewidth]{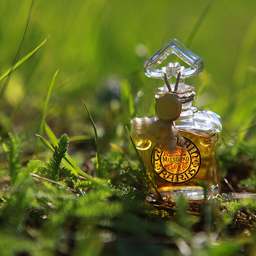} \\
        \includegraphics[width=0.18\linewidth]{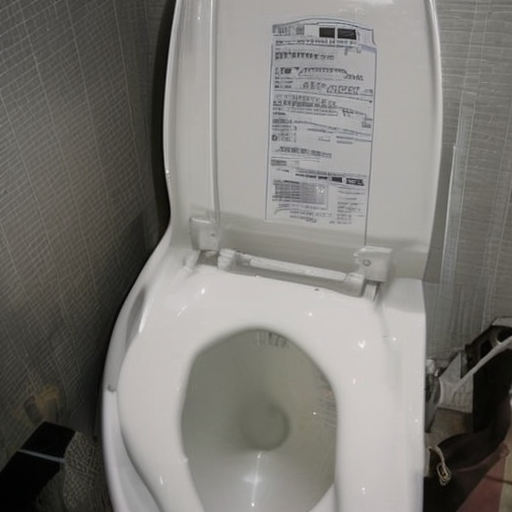} &
        \includegraphics[width=0.18\linewidth]{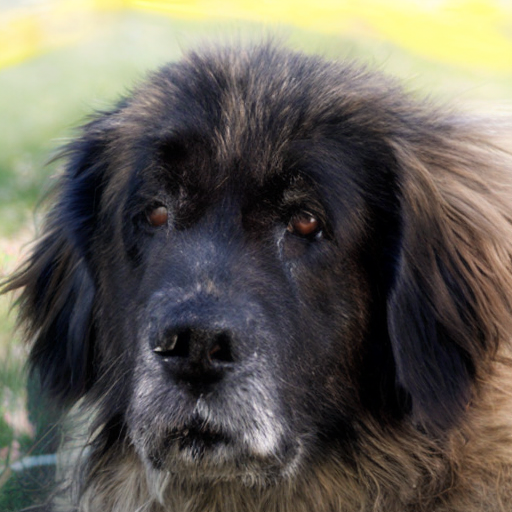} &
        \includegraphics[width=0.18\linewidth]{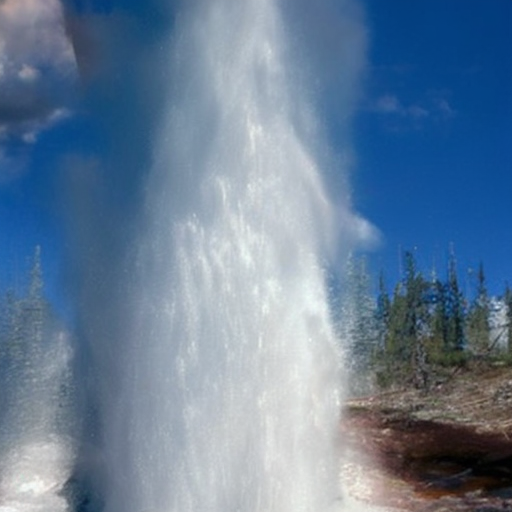} &
        \includegraphics[width=0.18\linewidth]{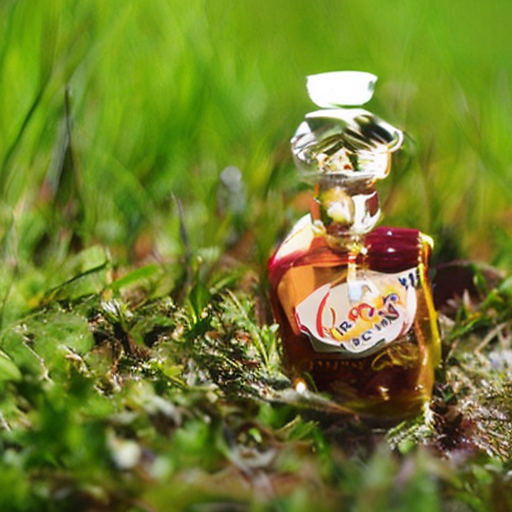} \\
    \end{tabular}
    \caption{\textbf{Qualitative results for resolution scalability.} Using the fine-tuned 512 $\times$ 512 model, \ours successfully stochastically reconstruct the conditioning image at a 512 $\times$ 512 resolution. The \textbf{top row} shows the anchor images, and the \textbf{bottom row} shows the corresponding results for \ours (512x512).}
    \label{fig:512_qualitative}
\end{figure}

\begin{table}[h]
    \centering
    \caption{\textbf{Resolution scalability.} We evaluate our method's ability to scale to higher resolutions (512$\times$512) and compare the performance across unconditional generation with different granularity settings for our feature conditioning.}
    \begin{tabular}{lcccccc}
        \toprule
        & \textbf{FID} & \textbf{sFID} & \textbf{IS} & \textbf{Prec.} & \textbf{Rec.} & \textbf{DINOv2 Align.} \\
        \midrule
        Uncond & 33.1 & 10.1 & 39.1 & 0.61 & 0.6 & - \\
        Full & 5.4 & 6.0 & 178.8 & 0.75 & 0.67 & 0.61 \\
        Masked & 6.9 & 5.9 & 149.1 & 0.73 & 0.65 & 0.68 \\
        Average & 8.9 & 5.6 & 139.9 & 0.73 & 0.66 & 0.85 \\
        \bottomrule
    \end{tabular}
    \label{tab:resolution_scalability}
\end{table}

\subsection{Runtime analysis}\label{app:runtime_analysis}
We measure throughput on a single H100 GPU batch size 32, averaging across $n=20$ batches after a 5-batch warmup. The 95\% confidence intervals (computed from the standard deviation over the 20 batches) are also reported alongside the means to confirm the statistical stability of our measurements. As reported in Table~\ref{tab:runtime}, backpropagating gradients through the first 8-transformer blocks naturally increases latency; however, the resulting inference speeds remain well within an acceptable range.

We provide a detailed analysis of the computational overhead using an H100 GPU at 256 and 512 resolution with batch size 32. While \ours introduces a 26\% throughput reduction compared to unconditional generation due to the backward pass, the resulting inference speeds remains well within an acceptable range. For reference only, we report the throughput of classifier-free guidance (CFG), a guidance broadly used in diffusion models despite a throughput reduction of 48\%.
Closer to our method, TFG requires backpropagating through the entire network, resulting in a significant throughput reduction. In contrast, our method only utilizes 2GB of additional GPU memory with batch size 32 (+25\%) to store activations for the backward pass. This is a manageable trade-off for the level of semantic control achieved.

Our multi-source composition adds no extra throughput overhead. Indeed, all guidance signals are computed from the same set of feature maps extracted during the forward pass.

We observe similar behaviour of our method at 512 resolution. The number of patches increases with the resolution, leading to a higher absolute computational cost for the backward pass.

\begin{table}[h]
    \centering
        \caption{\textbf{Inference throughput.} Feature conditioning requires backpropagating gradients through the initial blocks of the transformer. While this results in a slight reduction in throughput, the computational overhead remains minimal. The baseline is the standard class-conditioned REPA-E model \cite{leng_repa-e_2025}, whereas our approach incorporates feature guidance into this framework.}
    \resizebox{\textwidth}{!}{
    \begin{tabular}{cclcc}
        \toprule
        & \textbf{Image Resolution} & \textbf{Method} & \textbf{Throughput (img/s)} & \textbf{Peak GPU memory (GB)} \\
        \midrule
        & \multirow{5}{*}{256}
        & Uncond. & $0.50 $ & 8 \\
        && Ours (Single source) & $0.37 $ & 10 \\
        && Ours (Multi source) & $0.37 $ & 10 \\
        && TFG & $0.21 $ & 52 \\
        && CFG & $0.26 $ & 8 \\
        && IP-Adapter & $0.19 $ & 9 \\
        \midrule
        & \multirow{2}{*}{512}
        & Uncond. & $0.12 $ & 23 \\
        && Ours & $0.08 $ & 29 \\
        \bottomrule
    \end{tabular}}

    \label{tab:runtime}
\end{table}
\begin{table*}[h]
    \centering
    \scriptsize
    \caption{A diverse set of 50 ImageNet \cite{deng2009imagenet} classes and 8 hand-selected target backgrounds used to evaluate multi-source compositional generation.}
    \setlength\extrarowheight{4pt}
    \begin{tabular}{p{0.1\textwidth}|p{0.73\textwidth}}
    \toprule
    \textbf{Super-category} & \textbf{Fine-grained Class Names} \\
    \midrule
    \texttt{Animals} & \texttt{Golden retriever}, \texttt{Tabby cat}, \texttt{Zebra}, \texttt{Bullfrog}, \texttt{Tarantula}, \texttt{Goldfish}, \texttt{Monarch butterfly}, \texttt{Crane}, \texttt{Bald eagle}, \texttt{Tiger} \\
    
    \texttt{Movables} & \texttt{Bicycle}, \texttt{Motor scooter}, \texttt{Mountain bike}, \texttt{Baby stroller}, \texttt{Lawnmower}, \texttt{Shopping cart}, \texttt{Tricycle}, \texttt{Car wheel}, \texttt{Fire engine}, \texttt{Sports car} \\
    
    \texttt{Food Flora} & \texttt{Strawberry}, \texttt{Banana}, \texttt{Granny Smith}, \texttt{Plate}, \texttt{Pizza}, \texttt{Cheeseburger}, \texttt{Agaric}, \texttt{Acorn}, \texttt{Burrito}, \texttt{Bee} \\
    
    \texttt{Tools Tech} & \texttt{Acoustic guitar}, \texttt{Broom}, \texttt{Hammer}, \texttt{Screwdriver}, \texttt{Soccer ball}, \texttt{Tennis racket}, \texttt{Crash helmet}, \texttt{Wall clock}, \texttt{Laptop}, \texttt{Ping-pong ball} \\
    
    \texttt{Household} & \texttt{Teapot}, \texttt{Coffeepot}, \texttt{Toaster}, \texttt{Vase}, \texttt{Cowboy hat}, \texttt{Running shoe}, \texttt{Backpack}, \texttt{Umbrella}, \texttt{Teddy bear}, \texttt{Balloon} \\
    \midrule
    \multicolumn{2}{l}{\textit{Our hand-selected target backgrounds}} \\
    \texttt{Background} & \texttt{Grass}, \texttt{Dirt}, \texttt{Snow}, \texttt{Ice}, \texttt{Carpet}, \texttt{Water}, \texttt{Brick}, \texttt{Wood} \\
    \bottomrule
        \end{tabular}
    \label{tab:classes} 
\end{table*}

\subsection{Complementary details on  multiple-source compositional generation}\label{app:patch_metrics_app}

\textbf{More details on classes.}
In Table \ref{tab:classes}, we provide the list of 50 ImageNet \cite{deng2009imagenet} classes and 8 hand-selected target backgrounds used to evaluate multi-source compositional generation.

\paragraph{Patch metrics for multi-concept.}
As detailed in Section \ref{sec:experiments}, we evaluate the quality of our compositions using patch-wise metrics. For each patch $n$ in the generated feature map $h_{\text{gen}}$, we calculate the cosine similarity against both the corresponding anchor patch $[h_{\text{ref}}]_n$ and the broadcasted target feature $[\tilde{h}_{\text{target}}]_n$:
\begin{equation}  s_{\text{anchor}, n} = \cos([h_{\text{gen}}]_n, [h_{\text{ref}}]_n) , \quad s_{\text{target}, n} = \cos([h_{\text{gen}}]_n, [\tilde{h}_{\text{target}}]_n) \end{equation}

Next, patches are assigned to the partition ($\Omega_{\text{anchor}}$ or $\Omega_{\text{target}}$) where they exhibit the highest similarity.
\begin{equation}
n \in
\begin{cases}
\Omega_{\text{anchor}} & \text{if } s_{\text{anchor}, n} > s_{\text{target}, n} \\
\Omega_{\text{target}} & \text{otherwise}
\end{cases}
\end{equation}

The resulting scores, $\mathcal{S}_{\text{anchor}}$ and $\mathcal{S}_{\text{target}}$, are the average alignments within their respective partitions:
\begin{equation} 
\mathcal{S}{_\text{anchor}} = \frac{1}{|\Omega{_\text{anchor}}|} \sum_{n \in \Omega_{\text{anchor}}} s_{\text{anchor}, n} 
\end{equation} 

\begin{equation} \mathcal{S}{_\text{target}} = \frac{1}{|\Omega{_\text{target}}|} \sum_{n \in \Omega_{\text{target}}} s_{\text{target}, n} 
\end{equation}

Finally, we compute a combined score by taking the arithmetic mean of $\mathcal{S}_{\text{anchor}}$ and $\mathcal{S}_{\text{target}}$:

\begin{equation}
    \mathcal{S}_{\text{combined}} = \frac{1}{2} \left( \mathcal{S}_{\text{target}} + \mathcal{S}_{\text{anchor}} \right)
\end{equation}

These metrics capture the model's precision in structural preservation alongside its responsiveness to the target potential $\mathscr{V}$.

\subsection{Failure Cases of Multiple-Source Compositional Generation}
\label{app:failures}
\begin{figure}[htbp]
    \centering
    \begin{tabular}{ccc}
        \textbf{Anchor} & \textbf{Target} & \textbf{Generated} \\
        \includegraphics[width=0.22\linewidth]{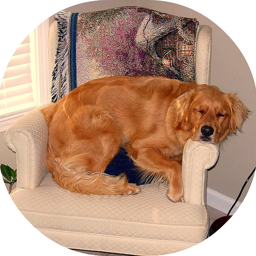} &
        \includegraphics[width=0.22\linewidth]{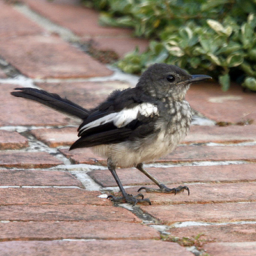} &
        \includegraphics[width=0.22\linewidth]{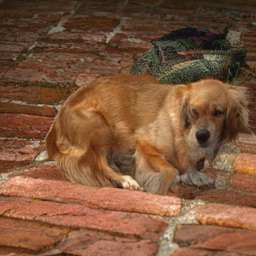} \\
        
        \includegraphics[width=0.22\linewidth]{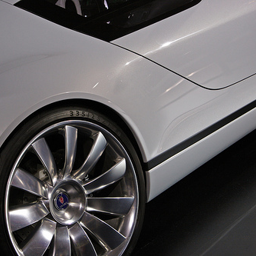} &
        \includegraphics[width=0.22\linewidth]{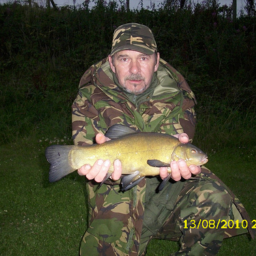} &
        \includegraphics[width=0.22\linewidth]{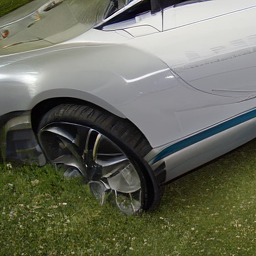} \\
        
        \includegraphics[width=0.22\linewidth]{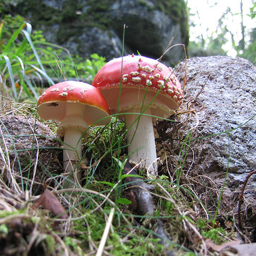} &
        \includegraphics[width=0.22\linewidth]{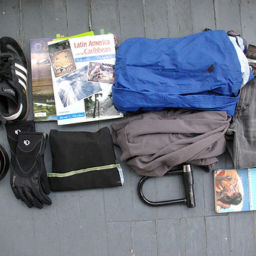} &
        \includegraphics[width=0.22\linewidth]{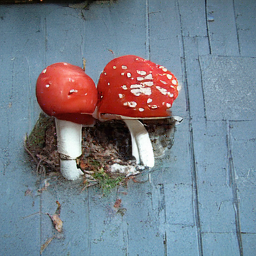} \\
    \end{tabular}
    \vspace{2mm}
    \caption{\textbf{Failure cases of multiple-source compositional generation.} Top row: Inconsistent aspect ratio. Middle row: Failed blending. Bottom row: Inconsistent view point.}
    \label{fig:multi_energy_failure}
\end{figure}

We exhibit in Figure~\ref{fig:multi_energy_failure} some failure cases of multiple-source compositional generation. Using two images with inconsitent view points as condition can lead to unrealistic blending as the resulting image will be subject to both view points. Similarly, inconsistent aspect ratios can lead to unrealistic blending. More generally, the model can fail to find a good compromise between the two conditions.   

\subsection{More Qualitative Results}
\label{app:more_qual}

\begin{figure}[ht]
  \centering
  \begin{tabular}{cc|ccc|ccc}
    \midrule
    & & &\textbf{REPA-E}& & &\textbf{SiT}& \\
  \midrule
  \textbf{Anchor} & \textbf{Mask} & \textbf{Full} & \textbf{Mask} & \textbf{Average} & \textbf{Full} & \textbf{Mask} & \textbf{Average} \\
  \includegraphics[width=0.07\textwidth]{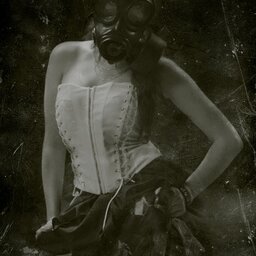} & \includegraphics[width=0.07\textwidth]{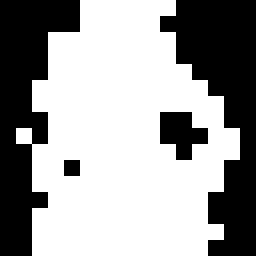}& \includegraphics[width=0.07\textwidth]{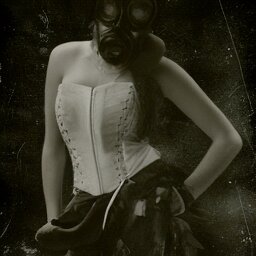} & \includegraphics[width=0.07\textwidth]{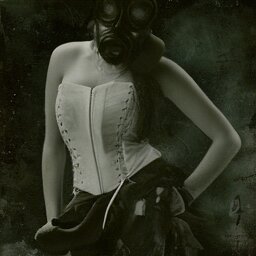} & \includegraphics[width=0.07\textwidth]{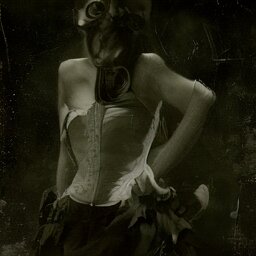}& \includegraphics[width=0.07\textwidth]{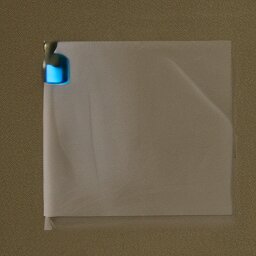} & \includegraphics[width=0.07\textwidth]{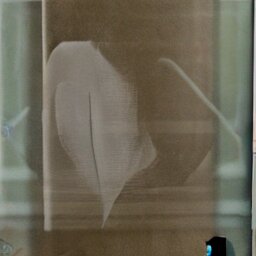} & \includegraphics[width=0.07\textwidth]{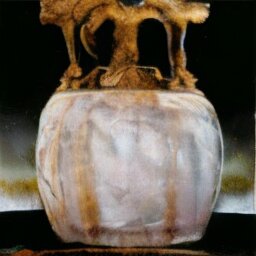} \\
  \includegraphics[width=0.07\textwidth]{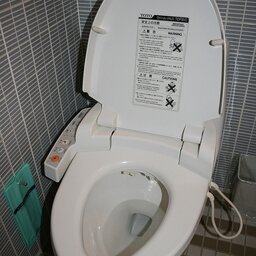} & \includegraphics[width=0.07\textwidth]{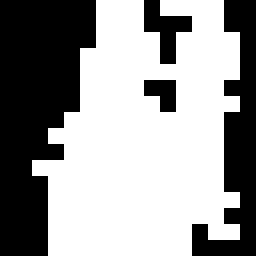}& \includegraphics[width=0.07\textwidth]{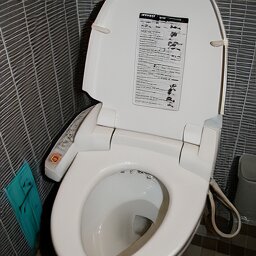} & \includegraphics[width=0.07\textwidth]{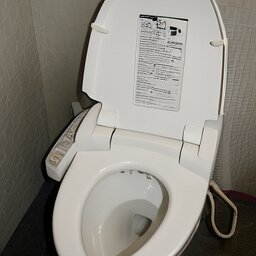} & \includegraphics[width=0.07\textwidth]{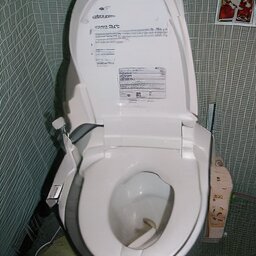}& \includegraphics[width=0.07\textwidth]{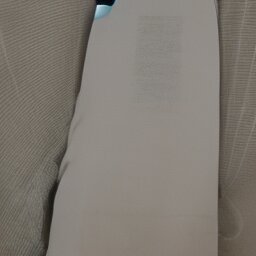} & \includegraphics[width=0.07\textwidth]{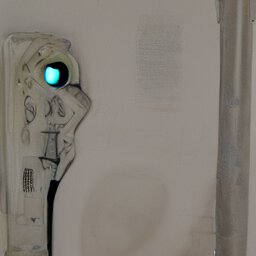} & \includegraphics[width=0.07\textwidth]{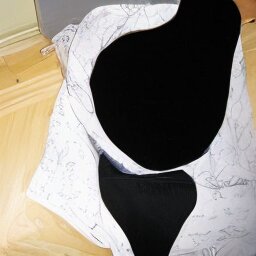} \\
  \includegraphics[width=0.07\textwidth]{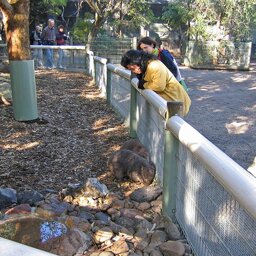} & \includegraphics[width=0.07\textwidth]{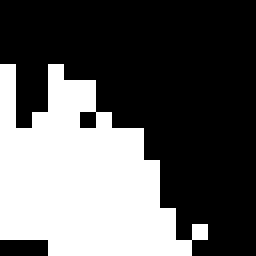}& \includegraphics[width=0.07\textwidth]{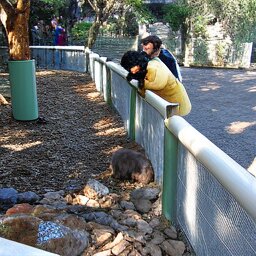} & \includegraphics[width=0.07\textwidth]{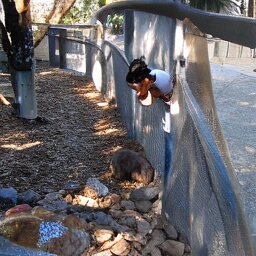} & \includegraphics[width=0.07\textwidth]{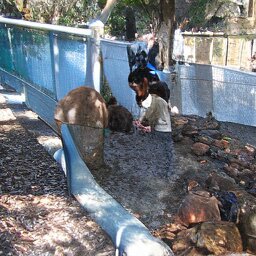}& \includegraphics[width=0.07\textwidth]{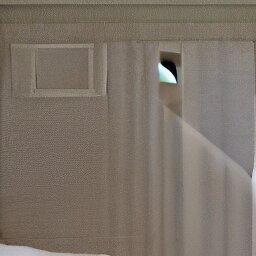} & \includegraphics[width=0.07\textwidth]{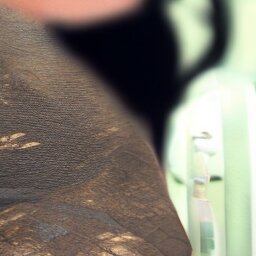} & \includegraphics[width=0.07\textwidth]{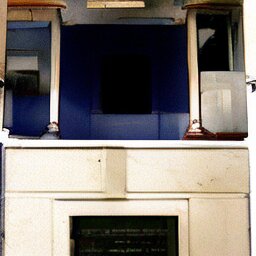} \\
  \includegraphics[width=0.07\textwidth]{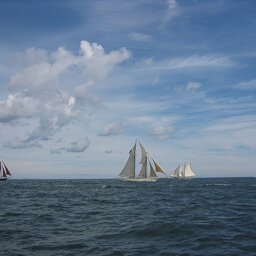} & \includegraphics[width=0.07\textwidth]{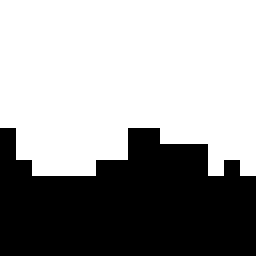}& \includegraphics[width=0.07\textwidth]{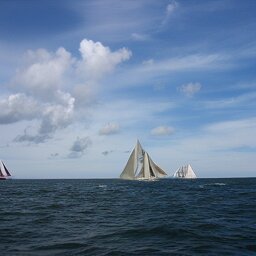} & \includegraphics[width=0.07\textwidth]{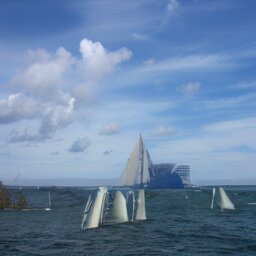} & \includegraphics[width=0.07\textwidth]{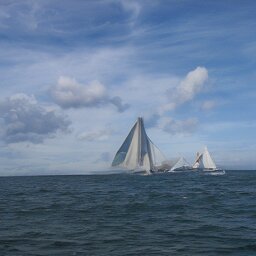}& \includegraphics[width=0.07\textwidth]{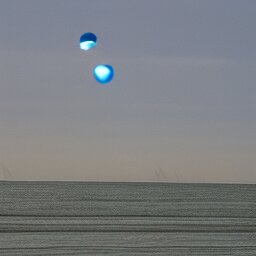} & \includegraphics[width=0.07\textwidth]{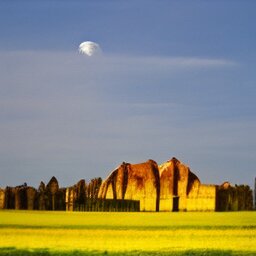} & \includegraphics[width=0.07\textwidth]{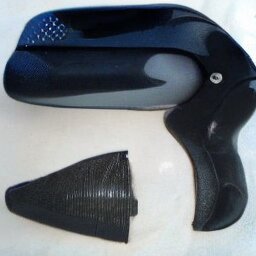} \\
  \includegraphics[width=0.07\textwidth]{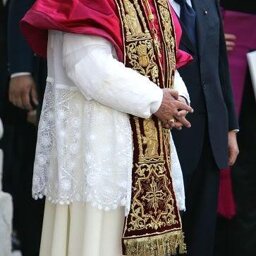} & \includegraphics[width=0.07\textwidth]{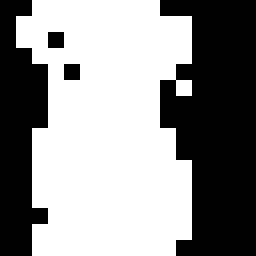}& \includegraphics[width=0.07\textwidth]{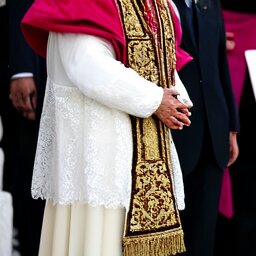} & \includegraphics[width=0.07\textwidth]{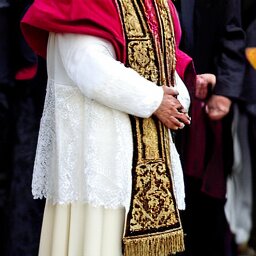} & \includegraphics[width=0.07\textwidth]{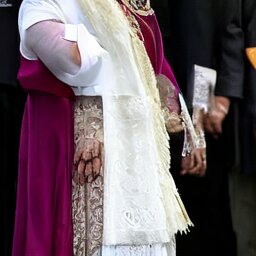}& \includegraphics[width=0.07\textwidth]{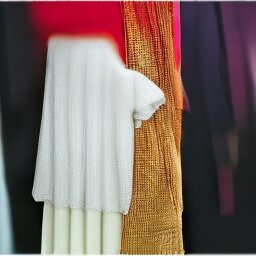} & \includegraphics[width=0.07\textwidth]{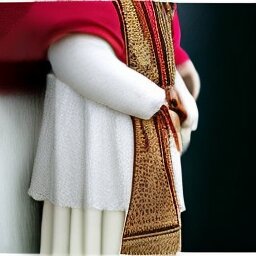} & \includegraphics[width=0.07\textwidth]{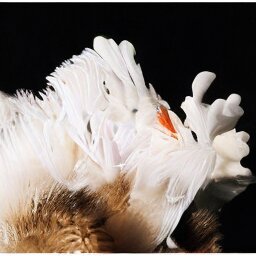} \\
  \includegraphics[width=0.07\textwidth]{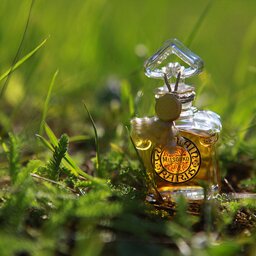} & \includegraphics[width=0.07\textwidth]{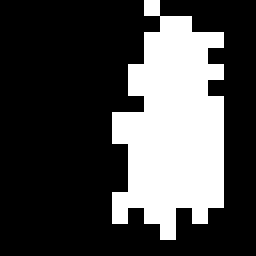}& \includegraphics[width=0.07\textwidth]{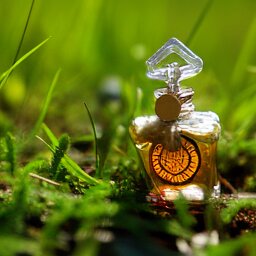} & \includegraphics[width=0.07\textwidth]{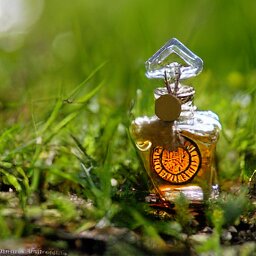} & \includegraphics[width=0.07\textwidth]{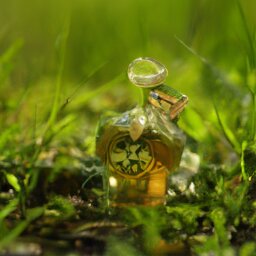}& \includegraphics[width=0.07\textwidth]{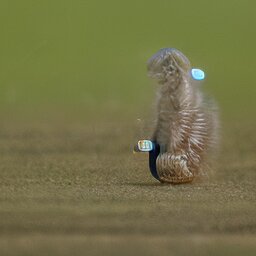} & \includegraphics[width=0.07\textwidth]{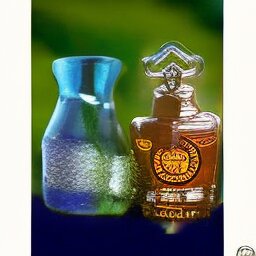} & \includegraphics[width=0.07\textwidth]{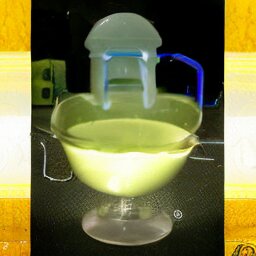} \\
  \includegraphics[width=0.07\textwidth]{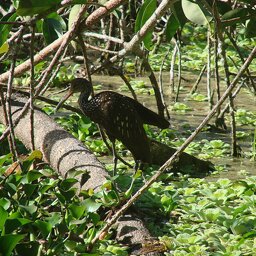} & \includegraphics[width=0.07\textwidth]{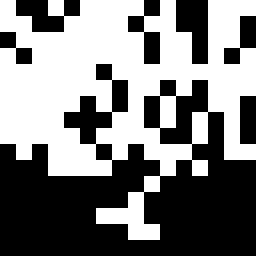}& \includegraphics[width=0.07\textwidth]{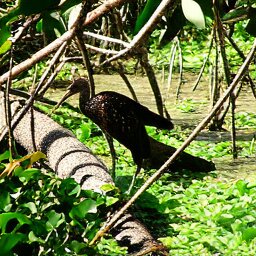} & \includegraphics[width=0.07\textwidth]{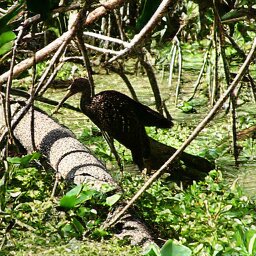} & \includegraphics[width=0.07\textwidth]{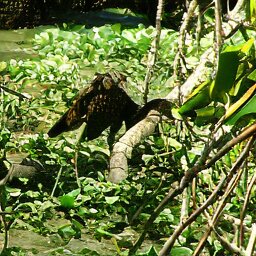}& \includegraphics[width=0.07\textwidth]{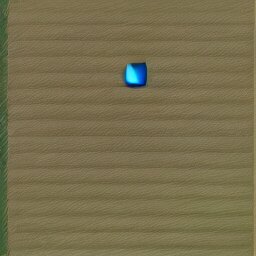} & \includegraphics[width=0.07\textwidth]{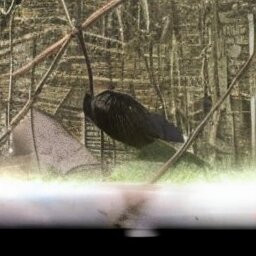} & \includegraphics[width=0.07\textwidth]{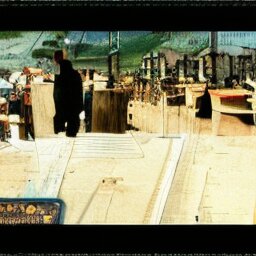} \\
  \includegraphics[width=0.07\textwidth]{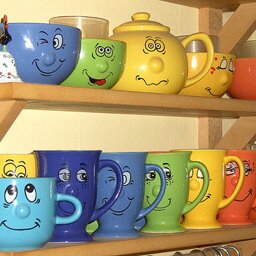} & \includegraphics[width=0.07\textwidth]{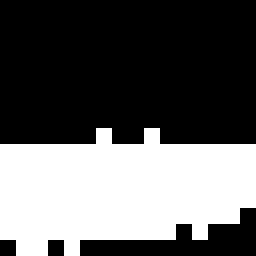}& \includegraphics[width=0.07\textwidth]{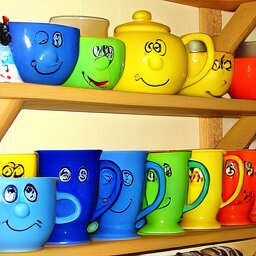} & \includegraphics[width=0.07\textwidth]{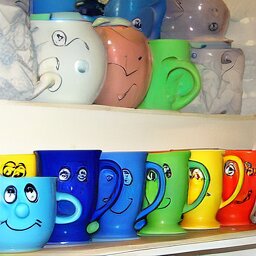} & \includegraphics[width=0.07\textwidth]{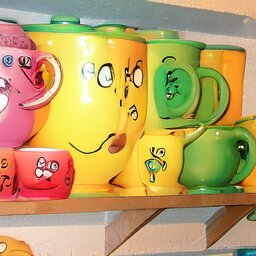}& \includegraphics[width=0.07\textwidth]{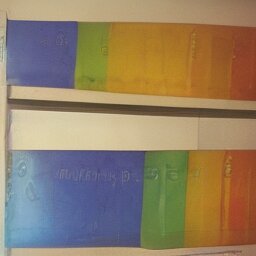} & \includegraphics[width=0.07\textwidth]{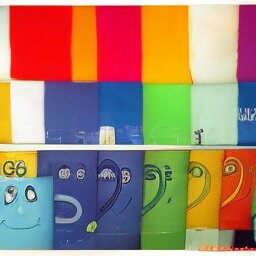} & \includegraphics[width=0.07\textwidth]{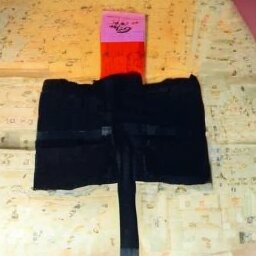} \\
  \includegraphics[width=0.07\textwidth]{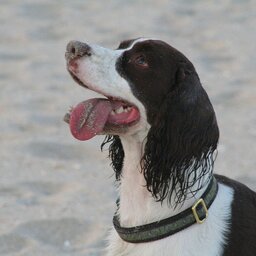} & \includegraphics[width=0.07\textwidth]{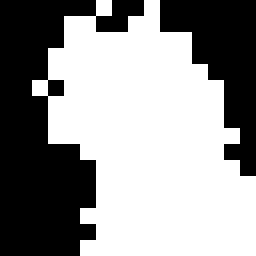}& \includegraphics[width=0.07\textwidth]{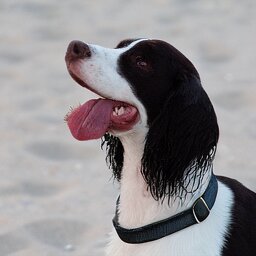} & \includegraphics[width=0.07\textwidth]{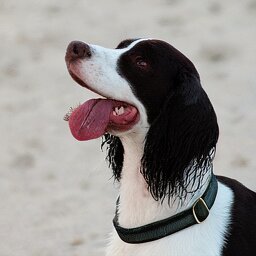} & \includegraphics[width=0.07\textwidth]{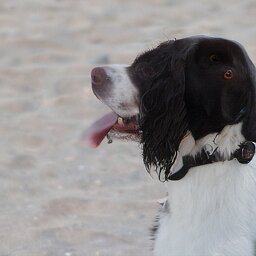}& \includegraphics[width=0.07\textwidth]{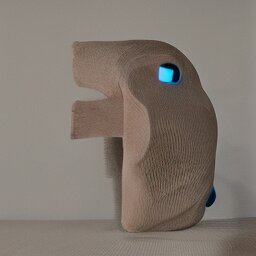} & \includegraphics[width=0.07\textwidth]{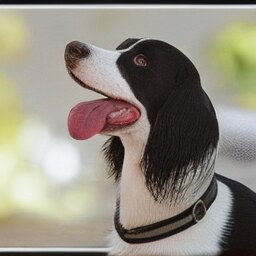} & \includegraphics[width=0.07\textwidth]{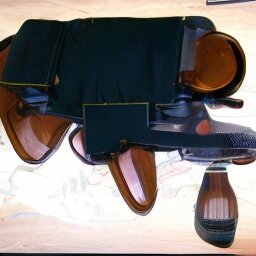} \\
  \includegraphics[width=0.07\textwidth]{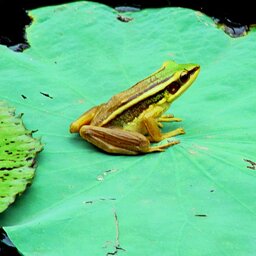} & \includegraphics[width=0.07\textwidth]{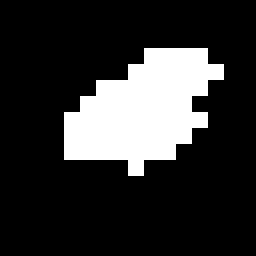}& \includegraphics[width=0.07\textwidth]{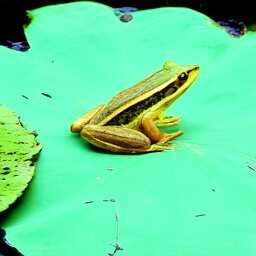} & \includegraphics[width=0.07\textwidth]{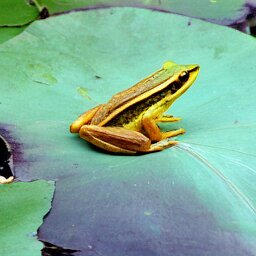} & \includegraphics[width=0.07\textwidth]{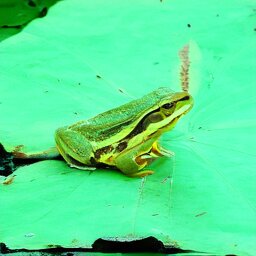}& \includegraphics[width=0.07\textwidth]{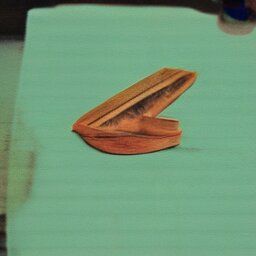} & \includegraphics[width=0.07\textwidth]{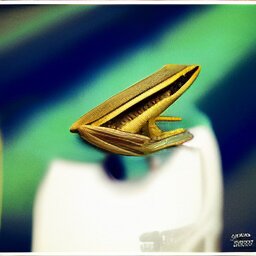} & \includegraphics[width=0.07\textwidth]{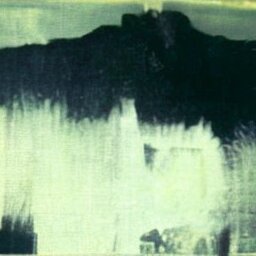} \\
  \includegraphics[width=0.07\textwidth]{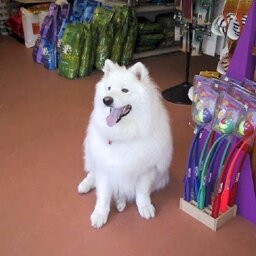} & \includegraphics[width=0.07\textwidth]{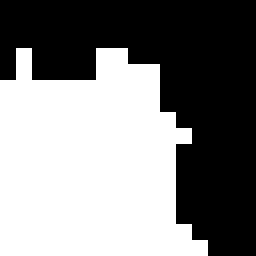}& \includegraphics[width=0.07\textwidth]{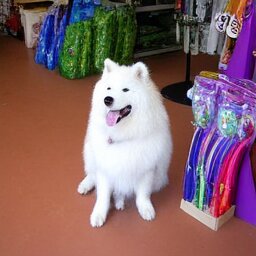} & \includegraphics[width=0.07\textwidth]{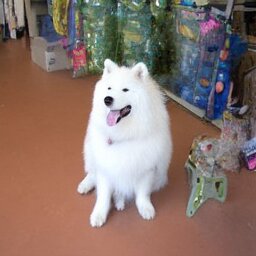} & \includegraphics[width=0.07\textwidth]{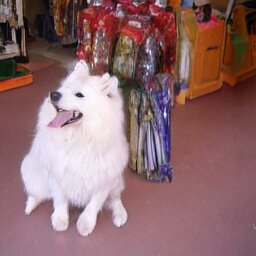}& \includegraphics[width=0.07\textwidth]{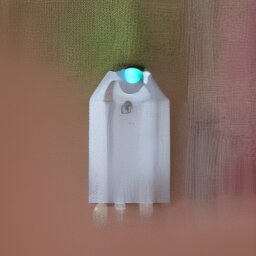} & \includegraphics[width=0.07\textwidth]{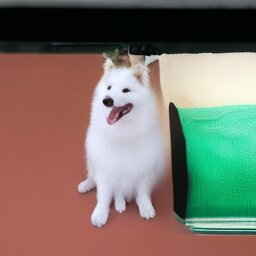} & \includegraphics[width=0.07\textwidth]{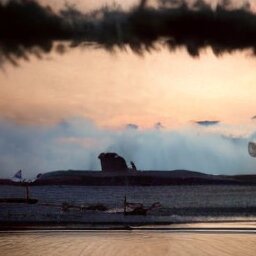} \\
  \includegraphics[width=0.07\textwidth]{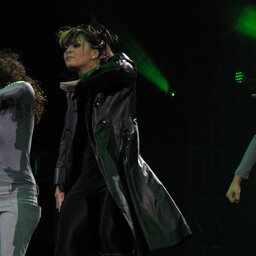} & \includegraphics[width=0.07\textwidth]{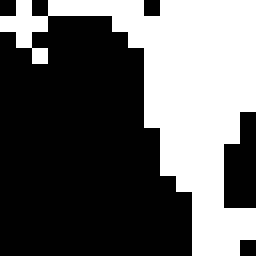}& \includegraphics[width=0.07\textwidth]{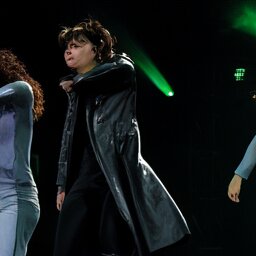} & \includegraphics[width=0.07\textwidth]{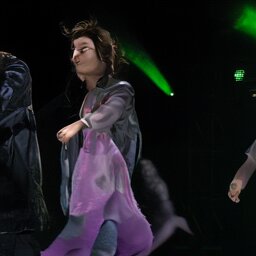} & \includegraphics[width=0.07\textwidth]{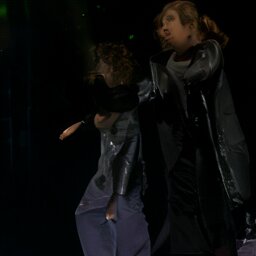}& \includegraphics[width=0.07\textwidth]{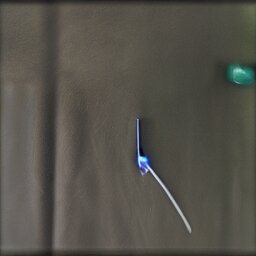} & \includegraphics[width=0.07\textwidth]{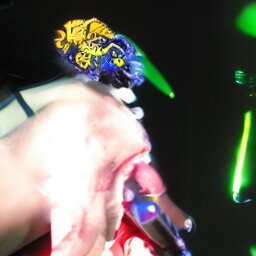} & \includegraphics[width=0.07\textwidth]{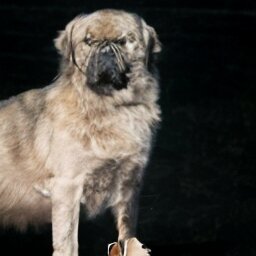} \\
  \includegraphics[width=0.07\textwidth]{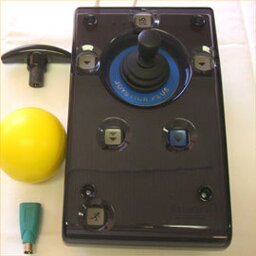} & \includegraphics[width=0.07\textwidth]{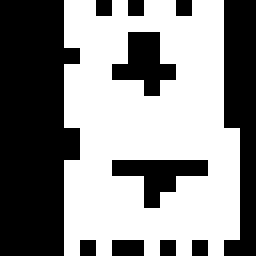}& \includegraphics[width=0.07\textwidth]{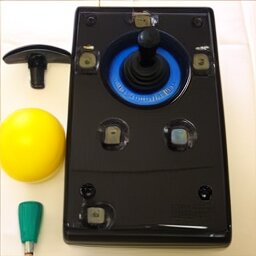} & \includegraphics[width=0.07\textwidth]{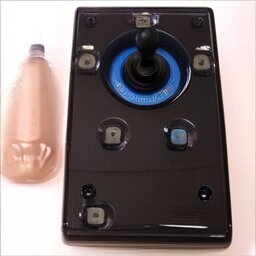} & \includegraphics[width=0.07\textwidth]{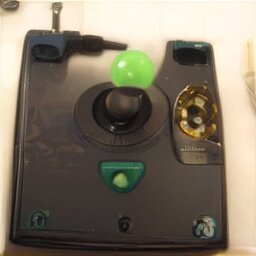}& \includegraphics[width=0.07\textwidth]{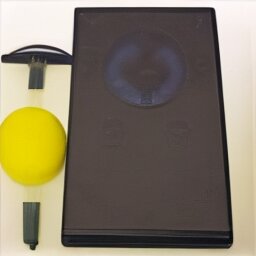} & \includegraphics[width=0.07\textwidth]{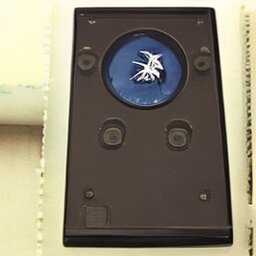} & \includegraphics[width=0.07\textwidth]{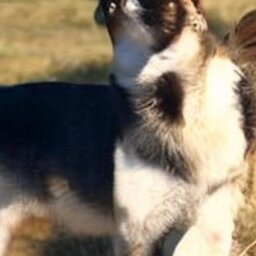} \\
  \includegraphics[width=0.07\textwidth]{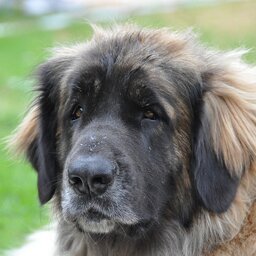} & \includegraphics[width=0.07\textwidth]{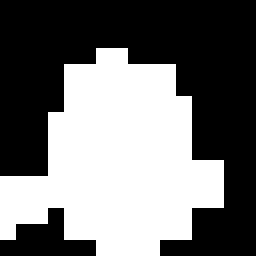}& \includegraphics[width=0.07\textwidth]{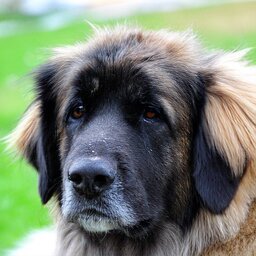} & \includegraphics[width=0.07\textwidth]{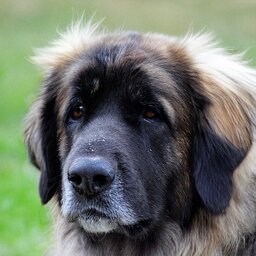} & \includegraphics[width=0.07\textwidth]{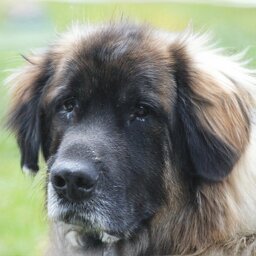}& \includegraphics[width=0.07\textwidth]{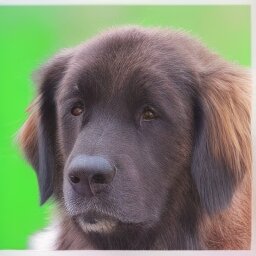} & \includegraphics[width=0.07\textwidth]{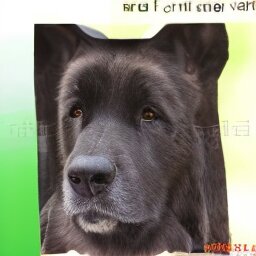} & \includegraphics[width=0.07\textwidth]{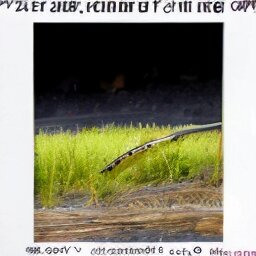} \\
  \end{tabular}
  \caption{\textbf{Qualitative results on ImageNet dataset \cite{deng2009imagenet}.} We present randomly selected samples from various conditioning modes. The first row shows the ground-truth (anchor) images, followed by their unsupervised foreground masks. Subsequent rows show generations using the full feature map, the masked feature map (using the corresponding mask), and the averaged feature map. All visualizations are from the following configuration: an unconditioned model guided by features from the diffusion model before the projection layer. The following block displays the corresponding results for standart SiT \cite{ma_sit_2024} model trained without representation alignment and REPA-E \cite{leng_repa-e_2025}. We observe that SiT models trained without representation alignment struggle to produce coherent results under feature guidance.}\label{fig:more_qual_imagenet}
  \end{figure}

\begin{figure}[ht]
  \centering
  \begin{tabular}{cc|ccc|ccc}
   \midrule
    &  & &\textbf{REPA-E}& & &\textbf{SiT}& \\
  \midrule
  \textbf{Anchor} & \textbf{Mask} & \textbf{Full} & \textbf{Mask} & \textbf{Average} & \textbf{Full} & \textbf{Mask} & \textbf{Average} \\

  \includegraphics[width=0.07\textwidth]{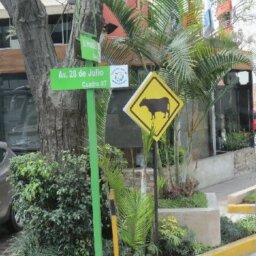} & \includegraphics[width=0.07\textwidth]{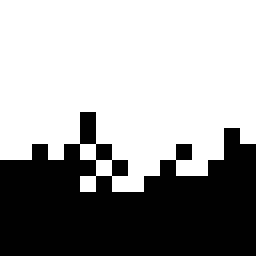} &\includegraphics[width=0.07\textwidth]{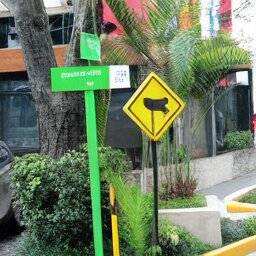} & \includegraphics[width=0.07\textwidth]{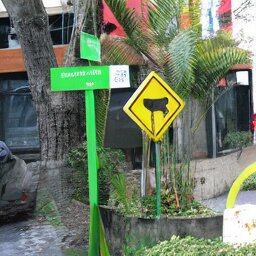} & \includegraphics[width=0.07\textwidth]{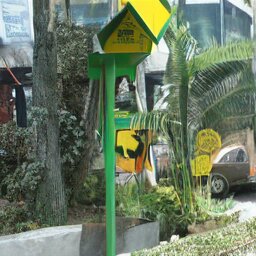} &\includegraphics[width=0.07\textwidth]{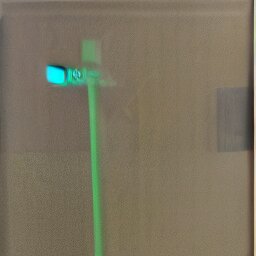} & \includegraphics[width=0.07\textwidth]{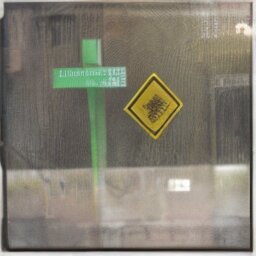} & \includegraphics[width=0.07\textwidth]{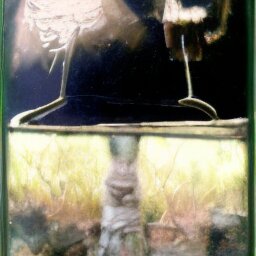} \\
  \includegraphics[width=0.07\textwidth]{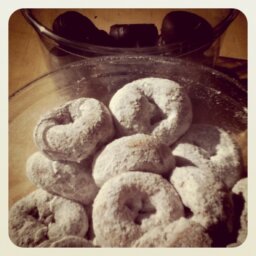} & \includegraphics[width=0.07\textwidth]{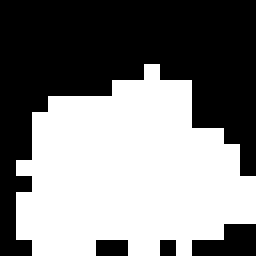} &\includegraphics[width=0.07\textwidth]{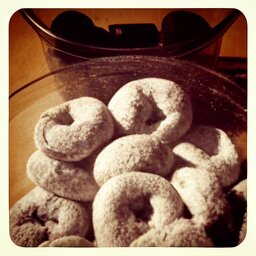} & \includegraphics[width=0.07\textwidth]{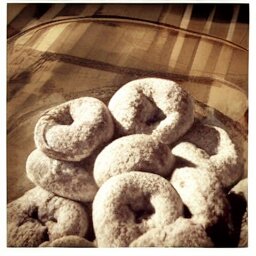} & \includegraphics[width=0.07\textwidth]{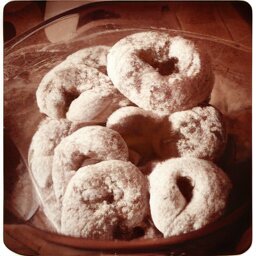} &\includegraphics[width=0.07\textwidth]{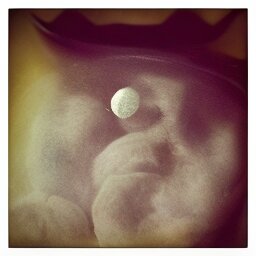} & \includegraphics[width=0.07\textwidth]{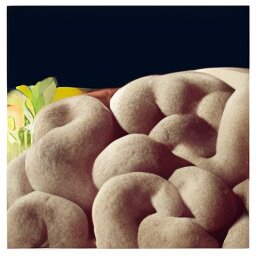} & \includegraphics[width=0.07\textwidth]{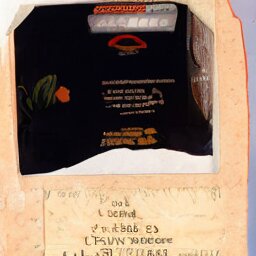} \\
  \includegraphics[width=0.07\textwidth]{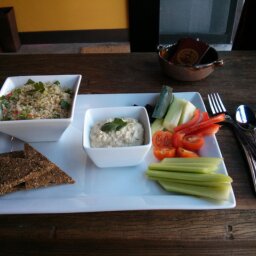} & \includegraphics[width=0.07\textwidth]{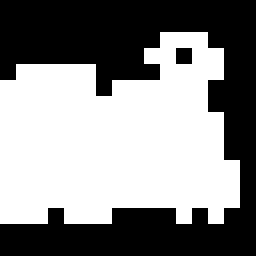} &\includegraphics[width=0.07\textwidth]{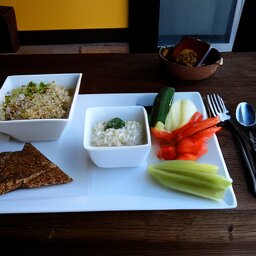} & \includegraphics[width=0.07\textwidth]{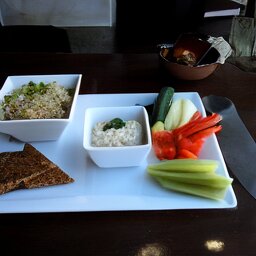} & \includegraphics[width=0.07\textwidth]{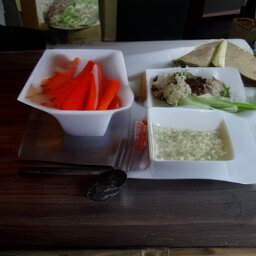} &\includegraphics[width=0.07\textwidth]{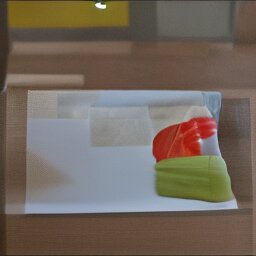} & \includegraphics[width=0.07\textwidth]{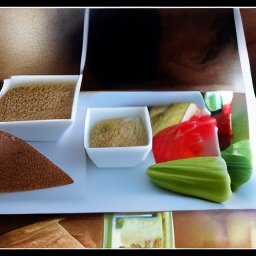} & \includegraphics[width=0.07\textwidth]{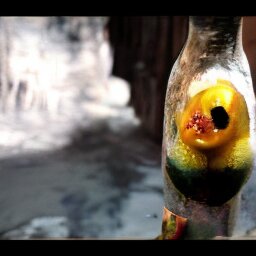} \\
  \includegraphics[width=0.07\textwidth]{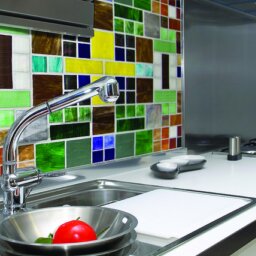} & \includegraphics[width=0.07\textwidth]{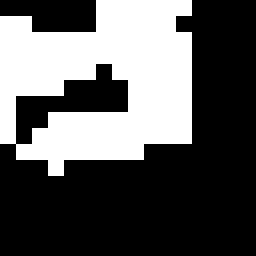} &\includegraphics[width=0.07\textwidth]{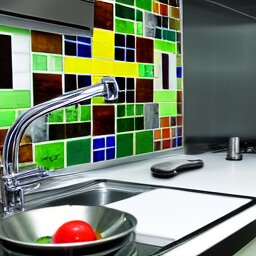} & \includegraphics[width=0.07\textwidth]{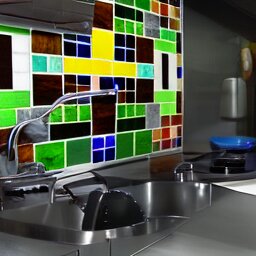} & \includegraphics[width=0.07\textwidth]{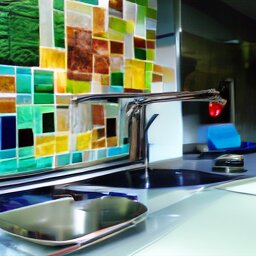} &\includegraphics[width=0.07\textwidth]{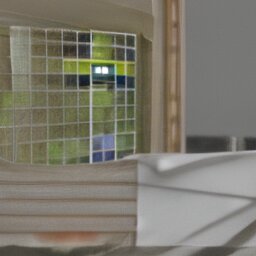} & \includegraphics[width=0.07\textwidth]{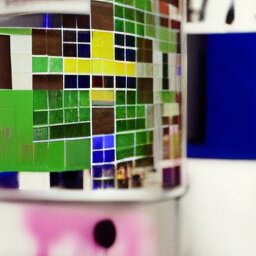} & \includegraphics[width=0.07\textwidth]{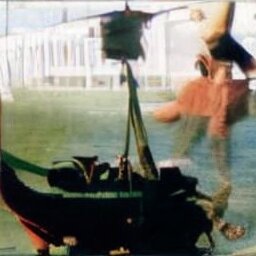} \\
  \includegraphics[width=0.07\textwidth]{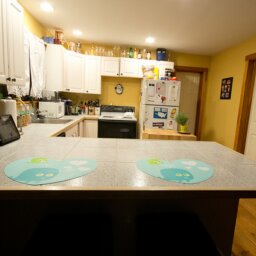} & \includegraphics[width=0.07\textwidth]{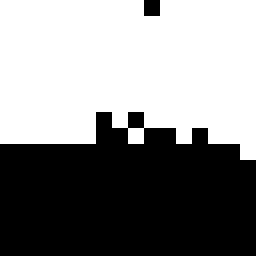} &\includegraphics[width=0.07\textwidth]{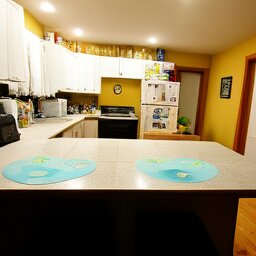} & \includegraphics[width=0.07\textwidth]{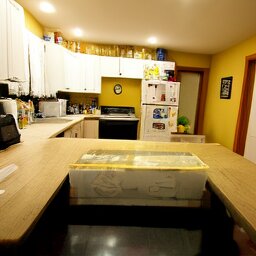} & \includegraphics[width=0.07\textwidth]{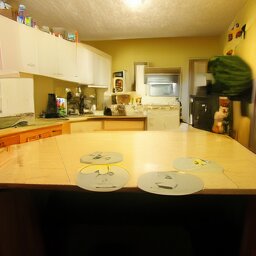} &\includegraphics[width=0.07\textwidth]{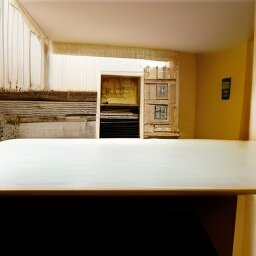} & \includegraphics[width=0.07\textwidth]{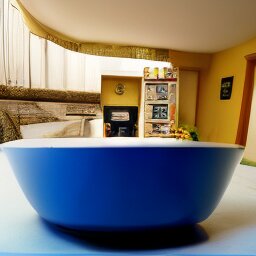} & \includegraphics[width=0.07\textwidth]{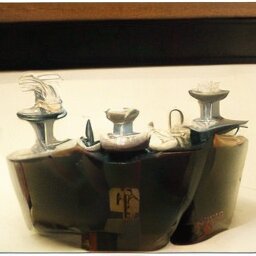} \\
  \includegraphics[width=0.07\textwidth]{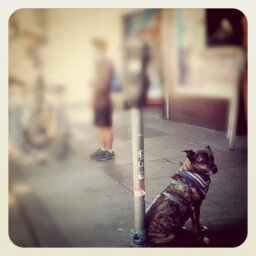} & \includegraphics[width=0.07\textwidth]{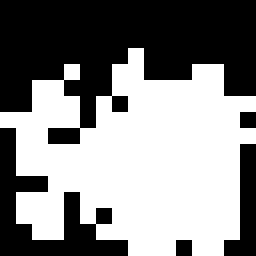} &\includegraphics[width=0.07\textwidth]{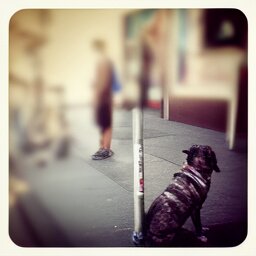} & \includegraphics[width=0.07\textwidth]{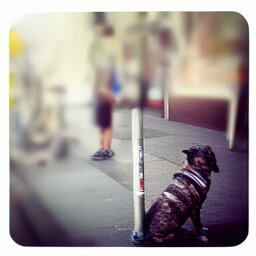} & \includegraphics[width=0.07\textwidth]{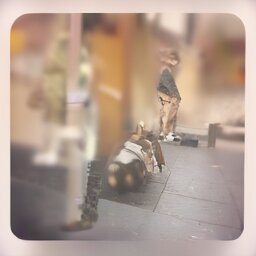} &\includegraphics[width=0.07\textwidth]{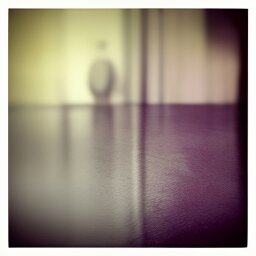} & \includegraphics[width=0.07\textwidth]{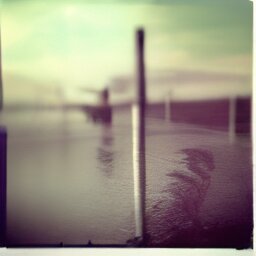} & \includegraphics[width=0.07\textwidth]{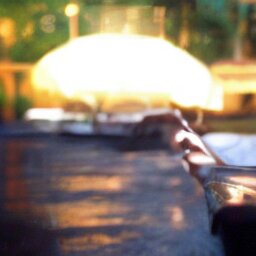} \\
  \includegraphics[width=0.07\textwidth]{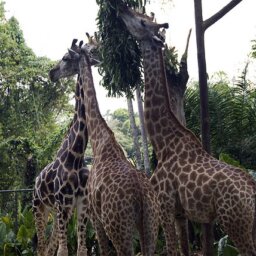} & \includegraphics[width=0.07\textwidth]{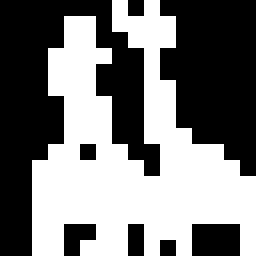} &\includegraphics[width=0.07\textwidth]{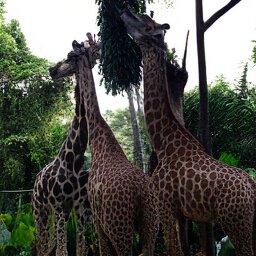} & \includegraphics[width=0.07\textwidth]{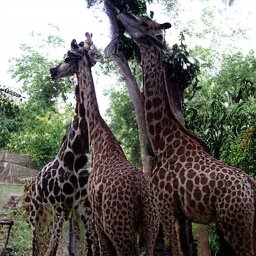} & \includegraphics[width=0.07\textwidth]{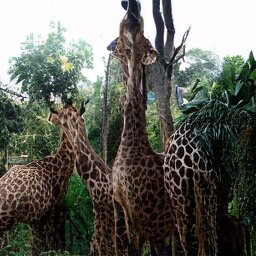} &\includegraphics[width=0.07\textwidth]{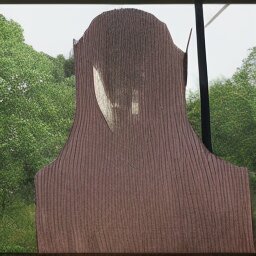} & \includegraphics[width=0.07\textwidth]{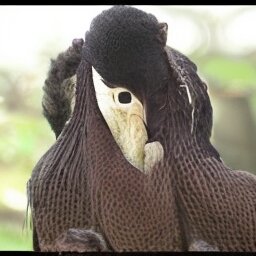} & \includegraphics[width=0.07\textwidth]{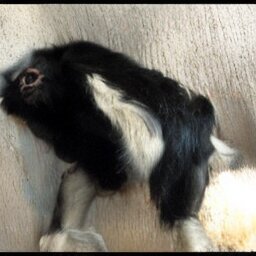} \\
  \includegraphics[width=0.07\textwidth]{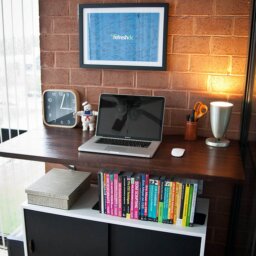} & \includegraphics[width=0.07\textwidth]{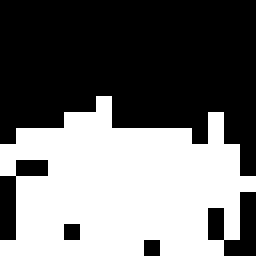} &\includegraphics[width=0.07\textwidth]{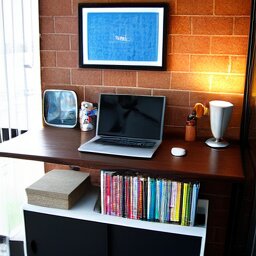} & \includegraphics[width=0.07\textwidth]{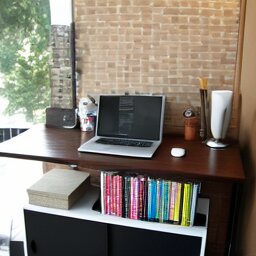} & \includegraphics[width=0.07\textwidth]{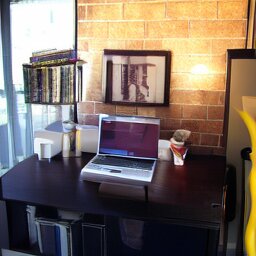} &\includegraphics[width=0.07\textwidth]{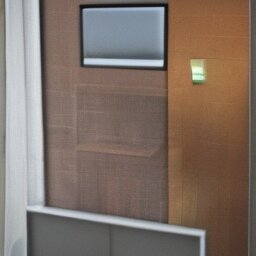} & \includegraphics[width=0.07\textwidth]{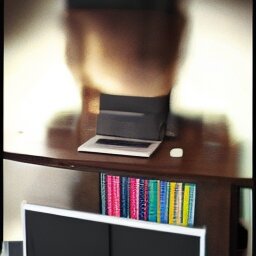} & \includegraphics[width=0.07\textwidth]{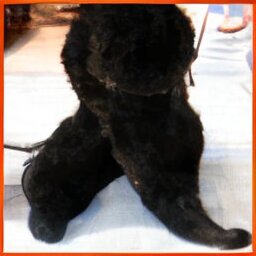} \\
  \includegraphics[width=0.07\textwidth]{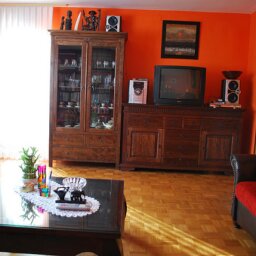} & \includegraphics[width=0.07\textwidth]{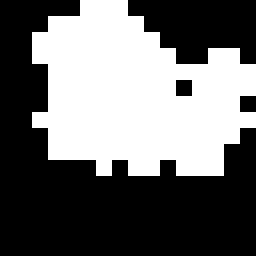} &\includegraphics[width=0.07\textwidth]{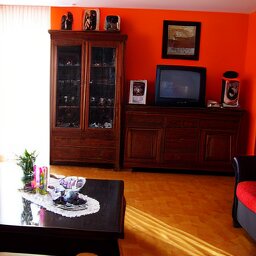} & \includegraphics[width=0.07\textwidth]{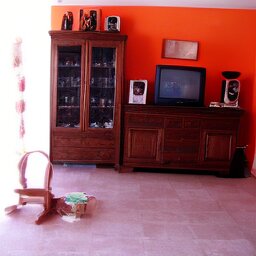} & \includegraphics[width=0.07\textwidth]{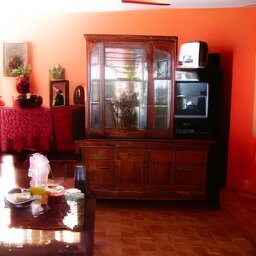} &\includegraphics[width=0.07\textwidth]{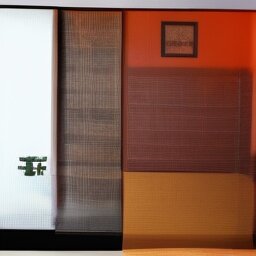} & \includegraphics[width=0.07\textwidth]{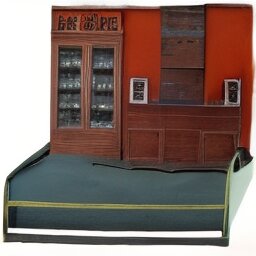} & \includegraphics[width=0.07\textwidth]{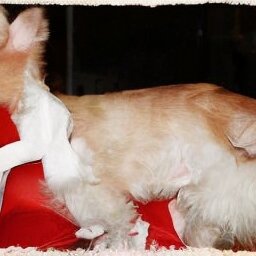} \\
  \includegraphics[width=0.07\textwidth]{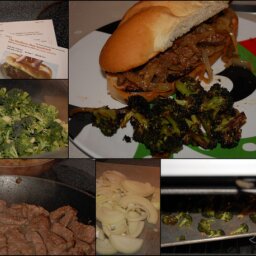} & \includegraphics[width=0.07\textwidth]{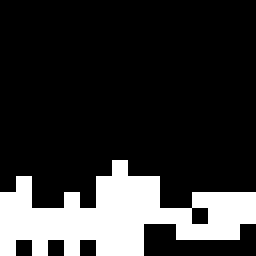} &\includegraphics[width=0.07\textwidth]{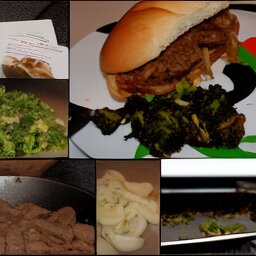} & \includegraphics[width=0.07\textwidth]{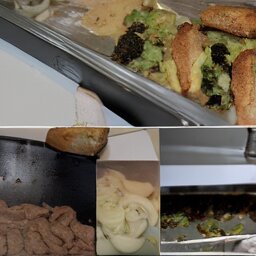} & \includegraphics[width=0.07\textwidth]{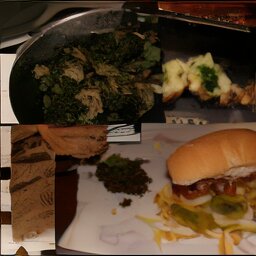} &\includegraphics[width=0.07\textwidth]{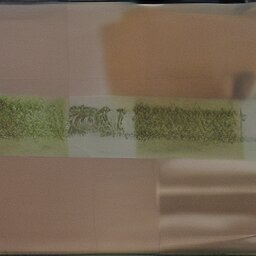} & \includegraphics[width=0.07\textwidth]{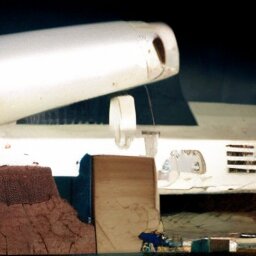} & \includegraphics[width=0.07\textwidth]{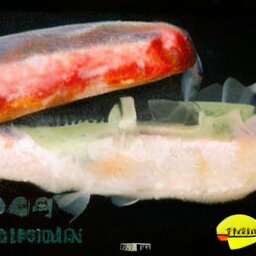} \\
  \includegraphics[width=0.07\textwidth]{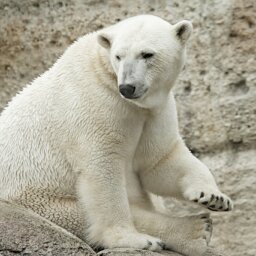} & \includegraphics[width=0.07\textwidth]{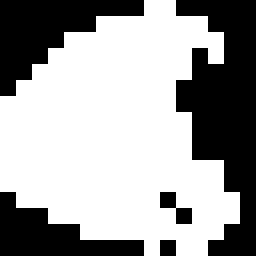} &\includegraphics[width=0.07\textwidth]{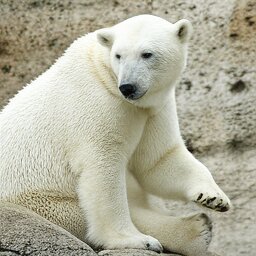} & \includegraphics[width=0.07\textwidth]{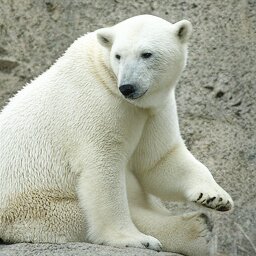} & \includegraphics[width=0.07\textwidth]{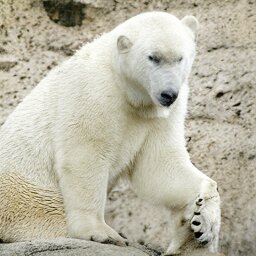} &\includegraphics[width=0.07\textwidth]{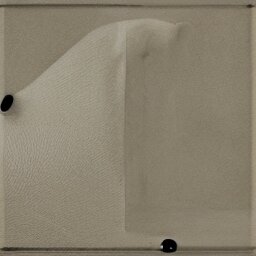} & \includegraphics[width=0.07\textwidth]{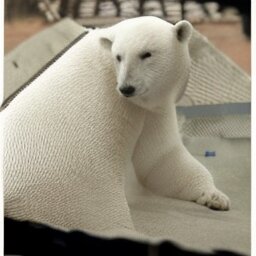} & \includegraphics[width=0.07\textwidth]{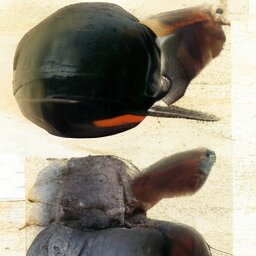} \\
  \includegraphics[width=0.07\textwidth]{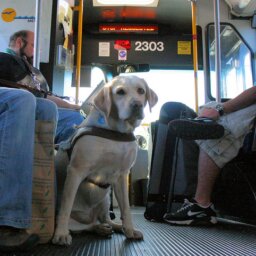} & \includegraphics[width=0.07\textwidth]{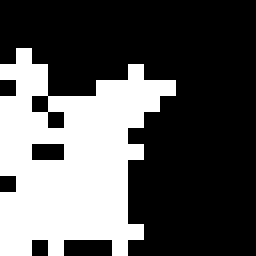} &\includegraphics[width=0.07\textwidth]{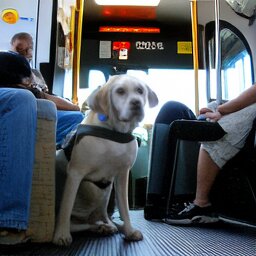} & \includegraphics[width=0.07\textwidth]{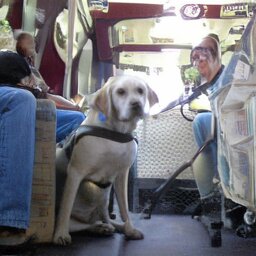} & \includegraphics[width=0.07\textwidth]{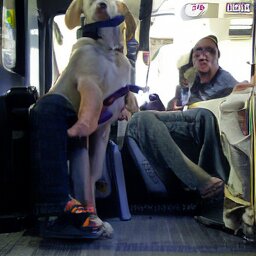} &\includegraphics[width=0.07\textwidth]{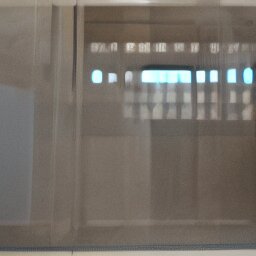} & \includegraphics[width=0.07\textwidth]{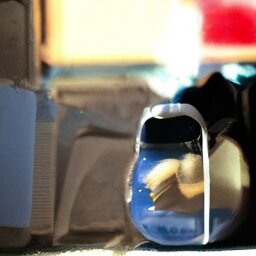} & \includegraphics[width=0.07\textwidth]{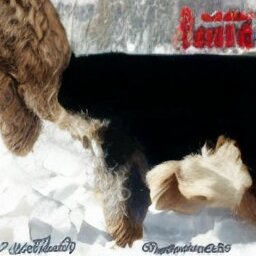} \\
  \includegraphics[width=0.07\textwidth]{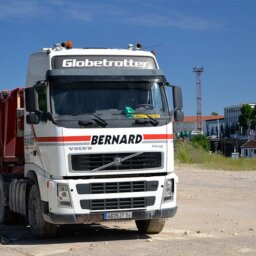} & \includegraphics[width=0.07\textwidth]{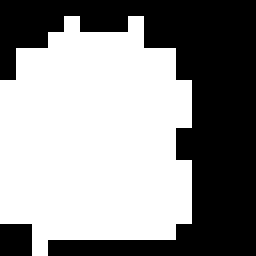} &\includegraphics[width=0.07\textwidth]{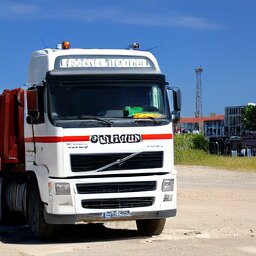} & \includegraphics[width=0.07\textwidth]{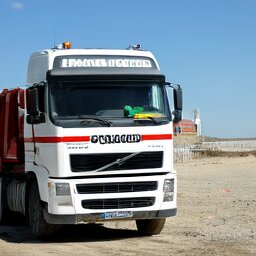} & \includegraphics[width=0.07\textwidth]{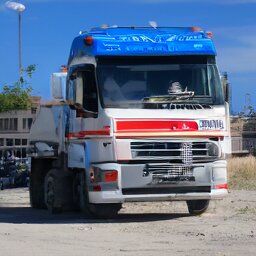} &\includegraphics[width=0.07\textwidth]{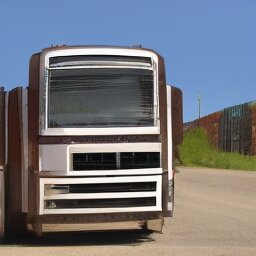} & \includegraphics[width=0.07\textwidth]{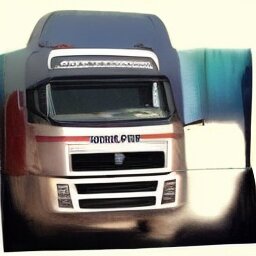} & \includegraphics[width=0.07\textwidth]{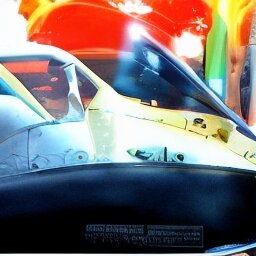} \\
  \includegraphics[width=0.07\textwidth]{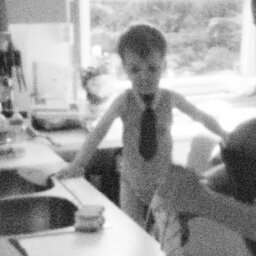} & \includegraphics[width=0.07\textwidth]{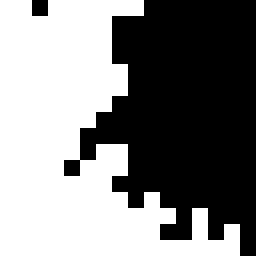} &\includegraphics[width=0.07\textwidth]{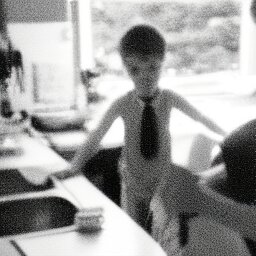} & \includegraphics[width=0.07\textwidth]{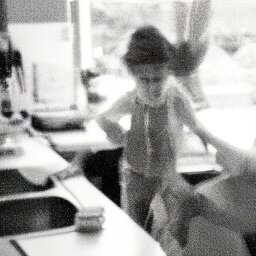} & \includegraphics[width=0.07\textwidth]{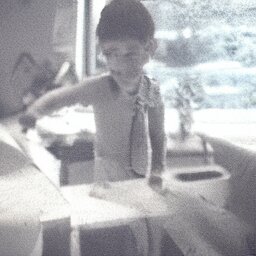} &\includegraphics[width=0.07\textwidth]{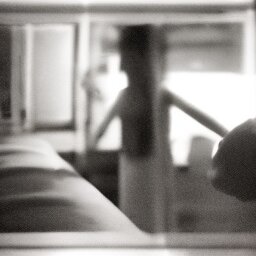} & \includegraphics[width=0.07\textwidth]{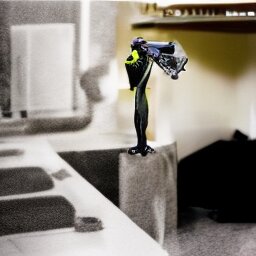} & \includegraphics[width=0.07\textwidth]{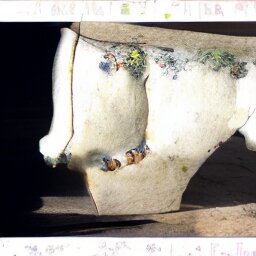} \\
  \includegraphics[width=0.07\textwidth]{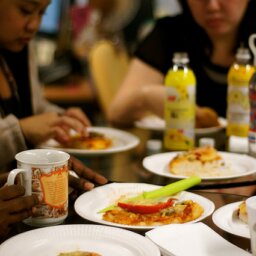} & \includegraphics[width=0.07\textwidth]{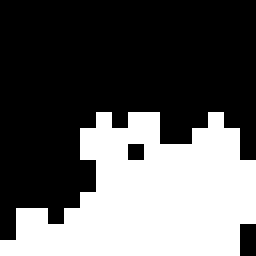} &\includegraphics[width=0.07\textwidth]{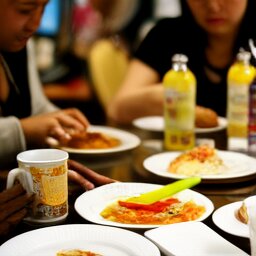} & \includegraphics[width=0.07\textwidth]{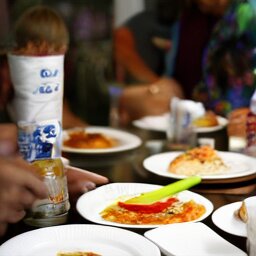} & \includegraphics[width=0.07\textwidth]{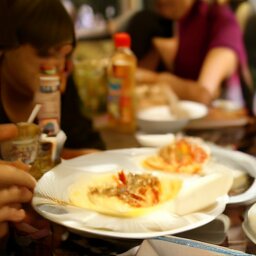} &\includegraphics[width=0.07\textwidth]{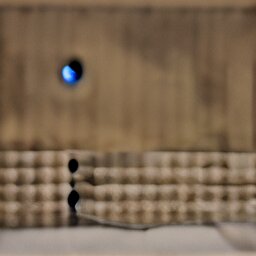} & \includegraphics[width=0.07\textwidth]{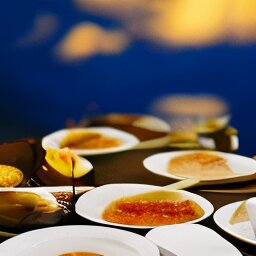} & \includegraphics[width=0.07\textwidth]{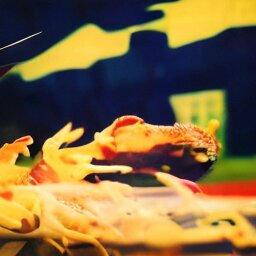} \\
  \end{tabular}
  \caption{\textbf{Qualitative results on COCO dataset \cite{coco}.} We present randomly selected samples from various conditioning modes. The first row shows the ground-truth (anchor) images, followed by their unsupervised foreground masks. Subsequent rows show generations using the full feature map, the masked feature map (using the corresponding mask), and the averaged feature map. All visualizations are from the following configuration: an unconditioned model guided by features from the diffusion model before the projection layer. The following block displays the corresponding results for standart SiT \cite{ma_sit_2024} model trained without representation alignment and REPA-E \cite{leng_repa-e_2025}. We observe that SiT models trained without representation alignment struggle to produce coherent results under feature guidance.}\label{fig:more_qual_coco}
  \end{figure}

Figure \ref{fig:more_qual_imagenet} and \ref{fig:more_qual_coco} provides additional qualitative results on ImageNet \cite{deng2009imagenet} and COCO \cite{coco}.

\section{Licenses}
\label{app:licenses}
In this work, we utilize several open-source datasets and codebases for training, evaluation, and baseline comparisons. We strictly adhere to their respective licenses and terms of use and confirm that our use of these assets aligns with their intended purpose of academic research.

\begin{itemize}
    \item \textbf{ImageNet}~\cite{deng2009imagenet}: Available under non-commercial use.
    \item \textbf{COCO}~\cite{coco}: Annotations are licensed under the Creative Commons Attribution 4.0 License (CC-BY 4.0), and images are subject to their respective Flickr terms of use.
    \item \textbf{SiT}~\cite{ma_sit_2024}, \textbf{REPA}~\cite{wang_repa_2025}, \textbf{REPA-E}~\cite{leng_repa-e_2025} and \textbf{CAD-I}~\cite{cad}: We utilize their released codebases and pre-trained models, which follow their provided open-source licenses (MIT License).
    \item \textbf{DINOv2}~\cite{oquab2024dinov2} and \textbf{Perception Encoder}~\cite{bolya2025PerceptionEncoder}: Licensed under the Apache 2.0 License.
    \item \textbf{CLIP}~\cite{radford2021learning}: Licensed under the MIT License.
    \item \textbf{JEPA}~\cite{jepa}, \textbf{MoCo v3}~\cite{chen2021empirical} and \textbf{MAE}~\cite{he2022masked}: Pre-trained models and code are distributed under the CC-BY-NC 4.0 license.
\end{itemize}


\end{document}